\documentclass[11pt]{article}

\usepackage{tgpagella}
\usepackage{mathpazo}
\usepackage{inconsolata}

\usepackage[letterpaper, margin=0.75in, top=0.6in, bottom=0.7in]{geometry}

\usepackage{microtype}
\usepackage{hyperref}
\usepackage{url}
\usepackage{booktabs}
\usepackage{natbib}
\setcitestyle{authoryear,round,citesep={;},aysep={,},yysep={;}}
\usepackage{fancyhdr}
\usepackage[table]{xcolor}

\usepackage{lineno}
\usepackage{caption}
\definecolor{darkblue}{rgb}{0, 0, 0.5}
\hypersetup{colorlinks=true, citecolor=darkblue, linkcolor=darkblue, urlcolor=darkblue}

\usepackage{graphicx}
\usepackage{subcaption}
\usepackage[many]{tcolorbox}
\usepackage{wrapfig}
\usepackage{float}

\usepackage{amsmath}
\usepackage{amssymb}
\usepackage{mathtools}
\usepackage{amsthm}

\usepackage{multirow}
\usepackage{colortbl}
\usepackage{textcomp}

\usepackage{enumitem}

\usepackage{listings}
\definecolor{solBase03}{HTML}{002B36}
\definecolor{solBase02}{HTML}{073642}
\definecolor{solBase01}{HTML}{586E75}
\definecolor{solBase00}{HTML}{657B83}
\definecolor{solBase0}{HTML}{839496}
\definecolor{solBase1}{HTML}{93A1A1}
\definecolor{solBase2}{HTML}{EEE8D5}
\definecolor{solBase3}{HTML}{FDF6E3}
\definecolor{msblue}{HTML}{073642}
\definecolor{mslight}{HTML}{586E75}
\definecolor{findingbg}{HTML}{ECF3F8}
\definecolor{findingborder}{HTML}{268BD2}
\definecolor{warnbg}{HTML}{FDF3E6}
\definecolor{warnborder}{HTML}{CB4B16}
\definecolor{greenbg}{HTML}{F0F5E8}
\definecolor{greenborder}{HTML}{859900}
\definecolor{cVanilla}{HTML}{268BD2}
\definecolor{cMemento}{HTML}{D33682}
\definecolor{cGrid}{HTML}{EEE8D5}
\definecolor{solCyan}{HTML}{2AA198}
\definecolor{solOrange}{HTML}{CB4B16}
\definecolor{solViolet}{HTML}{6C71C4}
\definecolor{solGreen}{HTML}{859900}
\definecolor{solYellow}{HTML}{B58900}
\definecolor{solRed}{HTML}{DC322F}

\tcbset{
  finding/.style={colback=cMemento!5, colframe=cMemento, boxrule=0.5pt, arc=3pt, left=6pt, right=6pt, top=4pt, bottom=4pt, fontupper=\small},
  warning/.style={colback=warnbg, colframe=warnborder, boxrule=0.5pt, arc=3pt, left=6pt, right=6pt, top=4pt, bottom=4pt, fontupper=\small},
  greenbox/.style={colback=greenbg, colframe=greenborder, boxrule=0.5pt, arc=3pt, left=6pt, right=6pt, top=4pt, bottom=4pt, fontupper=\small}
}

\usepackage{pgfplots}
\pgfplotsset{compat=1.18}
\usepgfplotslibrary{groupplots}
\usepackage{tikz}
\usetikzlibrary{arrows.meta, positioning, fit, calc, decorations.pathreplacing, patterns}

\usepackage[capitalize,noabbrev]{cleveref}

\usepackage{fontawesome5}

\usepackage[textsize=tiny]{todonotes}

\theoremstyle{plain}

\theoremstyle{definition}

\theoremstyle{remark}

\newcommand{\memento}{\textsc{Memento}}
\newcommand{\openmementos}{\textsc{OpenMementos}}

\usepackage{fontawesome5}

\definecolor{userbg}{HTML}{F2F8E8}
\definecolor{userborder}{HTML}{7CB518}
\definecolor{botbg}{HTML}{E8F4FD}
\definecolor{botborder}{HTML}{268BD2}
\definecolor{summarybg}{HTML}{FCEEF5}
\definecolor{summaryborder}{HTML}{D33682}
\definecolor{finalansbg}{HTML}{F0ECF8}
\definecolor{finalborder}{HTML}{6C71C4}
\definecolor{tagcolor}{HTML}{586E75}
\definecolor{guidebg}{HTML}{ECF3F8}
\definecolor{guideborder}{HTML}{268BD2}

\newcommand{\tagtext}[1]{{\color{tagcolor}\ttfamily\bfseries\footnotesize #1}}

\newtcolorbox{usermsg}{
  enhanced, breakable,
  colback=userbg,
  colframe=userborder,
  boxrule=0.5pt, arc=3pt,
  left=8pt, right=8pt, top=5pt, bottom=5pt,
  before skip=8pt, after skip=4pt,
  fontupper=\small,
  title={\faUser~\textbf{User}},
  coltitle=userborder!70!black,
  fonttitle=\small\bfseries,
  attach boxed title to top left={xshift=6pt, yshift=-2mm},
  boxed title style={colback=userbg, colframe=userborder, boxrule=0.4pt, arc=2pt}
}

\newtcolorbox{botmsg}{
  enhanced, breakable,
  colback=botbg,
  colframe=botborder,
  boxrule=0.5pt, arc=3pt,
  left=8pt, right=8pt, top=5pt, bottom=5pt,
  before skip=4pt, after skip=4pt,
  fontupper=\small,
  title={\faRobot~\textbf{Memento Qwen3-32B}},
  coltitle=cVanilla!80!black,
  fonttitle=\small\bfseries,
  attach boxed title to top left={xshift=6pt, yshift=-2mm},
  boxed title style={colback=botbg, colframe=botborder, boxrule=0.4pt, arc=2pt}
}

\newtcolorbox{summaryblock}{
  enhanced,
  colback=summarybg,
  colframe=summaryborder,
  boxrule=0.5pt, arc=2pt,
  left=6pt, right=6pt, top=3pt, bottom=3pt,
  before skip=2pt, after skip=5pt,
  fontupper=\footnotesize\bfseries\color{cMemento!80!black},
  borderline west={2.5pt}{0pt}{cMemento},
  title={\footnotesize\faBookmark~\textbf{memento}},
  coltitle=cMemento,
  fonttitle=\footnotesize\bfseries,
  attach boxed title to top left={xshift=4pt, yshift=-2mm},
  boxed title style={colback=summarybg, colframe=summaryborder, boxrule=0.3pt, arc=2pt, left=2pt, right=2pt, top=0.5pt, bottom=0.5pt}
}

\newtcolorbox{finalanswer}{
  enhanced,
  colback=finalansbg,
  colframe=finalborder,
  boxrule=0.5pt, arc=2pt,
  left=6pt, right=6pt, top=4pt, bottom=4pt,
  before skip=4pt, after skip=4pt,
  fontupper=\small,
  borderline west={2.5pt}{0pt}{solViolet},
}

\newtcolorbox{codeio}{
  enhanced,
  colback=solBase3!50,
  colframe=solBase1!60,
  boxrule=0.4pt, arc=2pt,
  left=6pt, right=6pt, top=3pt, bottom=3pt,
  before skip=3pt, after skip=3pt,
  fontupper=\small\ttfamily,
}

\newtcolorbox{readingguide}{
  enhanced,
  colback=guidebg!30,
  colframe=guideborder!60,
  boxrule=0.4pt, arc=3pt,
  left=8pt, right=8pt, top=6pt, bottom=6pt,
  fontupper=\small,
  before skip=10pt,
}

\begin{document}
\thispagestyle{plain}

\vspace*{-0.4in}
\begin{center}
{\Large\bf \memento{}: Teaching LLMs to Manage Their Own Context}\\[10pt]
{\normalsize Vasilis Kontonis$^{*}$ \quad Yuchen Zeng \quad Shivam Garg \quad Lingjiao Chen \quad Hao Tang \quad Ziyan Wang}\\[1pt]
{\normalsize Ahmed Awadallah \quad Eric Horvitz \quad John Langford \quad Dimitris Papailiopoulos$^{*}$}\\[2pt]
{\normalsize\bf Microsoft Research}\\[8pt]
{\footnotesize \faGithub\ \textbf{Code:} \url{https://github.com/microsoft/memento} \\
\raisebox{-0.2em}{\includegraphics[height=1.1em]{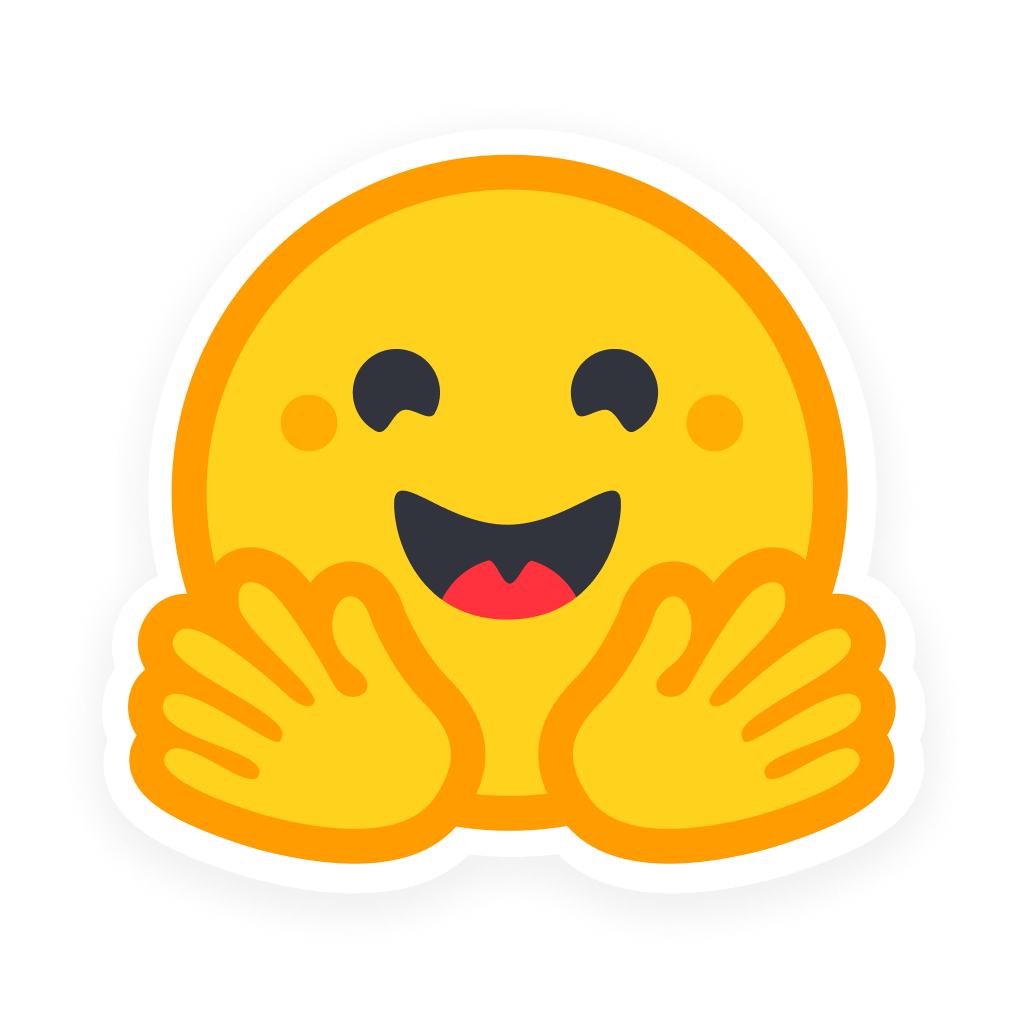}}\ \textbf{\openmementos{}:} \url{https://huggingface.co/datasets/microsoft/OpenMementos}
}
\end{center}
\renewcommand{\thefootnote}{\fnsymbol{footnote}}
\footnotetext[1]{Corresponding authors. Emails: \texttt{\{vkontonis, zengyuchen, shigarg, lingjiaochen, tanghao, t-ziyanwang, ahmed.awadallah, horvitz, jcl, dimitriosp\}@microsoft.com}}
\renewcommand{\thefootnote}{\arabic{footnote}}
\vspace{-0.5cm}
\noindent\rule{\textwidth}{0.4pt}
\vspace{-0.2cm}

\renewcommand{\abstractname}{}
\begin{abstract}
\vspace{-0.5cm}
Reasoning models think in long, unstructured streams with no mechanism for
compressing or organizing their own intermediate state. We introduce \memento{}:
a method that teaches models to segment reasoning into blocks, compress each
block into a \emph{memento}, i.e., a dense state summary, and reason forward by
attending only to mementos, reducing context, KV cache, and compute. To train
\memento{} models, we release \openmementos{}, a public dataset of 228K
reasoning traces derived from OpenThoughts-v3, segmented and annotated with
intermediate summaries. We show that a two-stage SFT recipe on \openmementos{} is effective across different model families (Qwen3, Phi-4, Olmo 3) and scales (8B--32B parameters). Trained models maintain strong accuracy on math, science, and coding benchmarks while achieving ${\sim}2.5\times$ peak KV cache
reduction. 
We extend vLLM to support our inference method,
achieving ${\sim}1.75\times$ throughput improvement while also enabling us to
perform RL and further improve accuracy. 
Finally, we identify a \emph{dual information stream}: information from each reasoning block is carried both by the memento text and by the corresponding KV states, which retain implicit information from the original block.
Removing this channel drops accuracy by 15\,pp on AIME24.
\end{abstract}
\vspace{0.2cm}

\begin{figure}[H]
    \centering
  \vspace{-0.2cm}
  \resizebox{0.95\textwidth}{!}{\begin{tikzpicture}[
  every node/.style={font=\normalsize},
  title/.style={font=\bfseries\normalsize, text=solBase02},
  rawbox/.style={draw=cVanilla, fill=findingbg, rounded corners=2pt,
                 text=solBase01, font=\footnotesize, text width=5.8cm,
                 inner sep=4pt, align=left},
  blockbox/.style={draw=#1, fill=white, rounded corners=2pt,
                   text=solBase01, font=\footnotesize, text width=5.8cm,
                   inner sep=4pt, align=left},
  blockbox/.default=cVanilla,
  fadedblock/.style={draw=#1!40, fill=white, rounded corners=2pt,
                     text=solBase01!40, font=\footnotesize, text width=2.0cm,
                     inner sep=3pt, align=left, densely dashed},
  fadedblock/.default=solBase01,
  membox/.style={draw=cMemento, fill=cMemento!8, rounded corners=2pt,
                 text=solBase01, font=\footnotesize, text width=2.6cm,
                 inner sep=3pt, align=left},
  pill/.style={draw=cVanilla, fill=white, rounded corners=3pt,
               font=\footnotesize\bfseries, text=solBase02, inner sep=2pt},
  arr/.style={-{Stealth[length=3pt]}, solBase1, thick},
  seqbox/.style={rounded corners=2pt, minimum height=0.45cm,
                 font=\footnotesize\bfseries, inner sep=3pt},
  prompt/.style={seqbox, draw=cVanilla, fill=cVanilla!15, text=solBase02,
                 minimum width=0.9cm},
  think/.style={seqbox, draw=cVanilla, fill=white, text=cVanilla,
                minimum width=0.9cm},
  memento/.style={seqbox, draw=cMemento, fill=cMemento, text=white,
                  minimum width=0.5cm},
  masked/.style={seqbox, draw=solBase1, fill=white, text=solBase1,
                 densely dashed, opacity=0.4, minimum width=0.9cm},
  answer/.style={seqbox, draw=solViolet, fill=solViolet, text=white,
                 minimum width=0.8cm},
]

\node[rawbox, anchor=north west] (raw) at (0, -0.7) {
  \textcolor{cVanilla}{\textlangle think\textrangle}\\[-1pt]
  Okay, so I need to find the sum of all integer bases $b{>}9$
  for which $17_b$ divides $97_b$.
  Let me convert to base 10\ldots\\[-1pt]
  \textcolor{cVanilla}{\textlangle/think\textrangle}~The answer is $\boxed{70}$.
};

\node[pill, anchor=south west] (s1) at ([yshift=3pt]raw.north west) {A. Trace Selection};

\node[pill, anchor=north west] (s2) at ([yshift=-0.2cm]raw.south west) {B. Segmentation};

\node[blockbox=solViolet, anchor=north west, text width=2.5cm] (b1)
  at ([yshift=-5pt]s2.south west) {
  \textcolor{solViolet}{\textbf{$T_1$}}~~Convert $17_b{=}b{+}7$, $97_b{=}9b{+}7$; set up divisibility\ldots
};
\node[membox, text width=2.5cm, right=0.4cm of b1] (m1) {
  \textcolor{cMemento}{\textbf{$M_1$}}~Need $(b{+}7) \mid (9b{+}7)$.
  Derived $b{+}7 \mid 56$.
};
\draw[arr, cMemento] (b1.east) -- (m1.west);

\node[blockbox=cVanilla, anchor=north west, text width=2.5cm] (b2)
  at ([yshift=-2pt]b1.south west) {
  \textcolor{cVanilla}{\textbf{$T_2$}}~~List divisors of 56; filter $b{+}7{>}16$\ldots
};
\node[membox, text width=2.5cm, right=0.4cm of b2] (m2) {
  \textcolor{cMemento}{\textbf{$M_2$}}~Valid: $b{=}21, 49$.
  Verified by substitution.
};
\draw[arr, cMemento] (b2.east) -- (m2.west);

\node[font=\normalsize\bfseries, text=solBase01, anchor=north] at ($(b2.east)!0.5!(m2.west)+(0,-0.75)$) {80\% compression};

\node[pill, anchor=north west] (s3) at (m1.north west |- s2.north west)
  {C. Compression};

\node[title, anchor=south] (ltitle)
  at ($(raw.north west)!0.5!(m1.north east |- raw.north east)+(0, 0.55)$)
  {Memento Data Gen};

\def\rowsep{1.0}
\def\bsep{0.08}

\node[prompt] (p1) at (7.0, -0.9) {Prompt};
\node[think, right=\bsep cm of p1] (t1a) {$T_1$};
\node[font=\footnotesize\itshape, text=solBase01, right=0.15cm of t1a] {generate $T_1$};

\node[prompt] (p2) at (7.0, -0.9-\rowsep) {Prompt};
\node[think, right=\bsep cm of p2] (t2a) {$T_1$};
\node[memento, right=\bsep cm of t2a] (t2m) {$M_1$};
\node[font=\footnotesize\itshape, text=solBase01, right=0.15cm of t2m] {compress $\to M_1$};

\node[prompt] (p3) at (7.0, -0.9-2*\rowsep) {Prompt};
\node[masked, right=\bsep cm of p3] (t3a) {$T_1$};
\node[memento, right=\bsep cm of t3a] (t3m) {$M_1$};
\node[think, right=\bsep cm of t3m] (t3b) {$T_2$};
\node[font=\footnotesize\itshape, text=solBase01, right=0.15cm of t3b] {mask $T_1$, gen. $T_2$};

\draw[densely dashed, solBase1, opacity=0.5]
  (6.4, -0.9-2.7*\rowsep) -- (14.0, -0.9-2.7*\rowsep);

\node[prompt, anchor=west] (pn) at (p3.west |- 0,-0.9-3.5*\rowsep) {Prompt};
\node[masked, right=\bsep cm of pn] (tna) {$T_1$};
\node[memento, right=\bsep cm of tna] (tnm1) {$M_1$};
\node[masked, right=\bsep cm of tnm1] (tnb) {$T_2$};
\node[memento, right=\bsep cm of tnb] (tnm2) {$M_2$};
\node[font=\footnotesize, text=solBase01, right=0.06cm of tnm2] (dots) {$\cdots$};
\node[masked, right=0.06cm of dots] (tnc) {$T_n$};
\node[memento, right=\bsep cm of tnc] (tnmn) {$M_n$};
\node[answer, right=\bsep cm of tnmn] (ans) {Answer};

\path (pn.west) -- (ans.east) coordinate[midway] (legend-center);
\node[masked, minimum width=0.6cm] (leg-m) at ([xshift=-0.8cm, yshift=-0.55cm]legend-center |- ans.south) {};
\node[font=\footnotesize, text=solBase01, right=2pt of leg-m] (leg-ml) {Masked};
\node[font=\footnotesize, text=solBase01, left=2pt of leg-m, anchor=east] (leg-tl) {Thinking};
\node[think, minimum width=0.6cm, left=2pt of leg-tl, anchor=east] (leg-t) {};
\node[memento, minimum width=0.6cm, right=0.3cm of leg-ml] (leg-me) {};
\node[font=\footnotesize, text=solBase01, right=2pt of leg-me] (leg-mel) {Memento};

\path (p1.west) -- (ans.east) coordinate[midway] (rmid);
\node[title] (rtitle) at (rmid |- ltitle) {Memento Attention at Training/Inference};

\begin{axis}[
  at={(13.6cm, -0.3cm)},
  anchor=north west,
  width=5.0cm,
  height=3.8cm,
  xmin=0, xmax=18000,
  ymin=0, ymax=2.4,
  xlabel={Tokens},
  xlabel style={font=\scriptsize\bfseries, at={(0.5,-0.08)}},
  ylabel={KV (GB)},
  ylabel style={font=\scriptsize\bfseries, at={(0.88,0.62)}},
  x tick label style={font=\scriptsize},
  y tick label style={font=\scriptsize},
  scaled x ticks=false,
  xtick={5000, 10000, 15000},
  xticklabels={5K, 10K, 15K},
  ytick={0.5, 1.0, 1.5, 2.0},
  yticklabel pos=right,
  grid=major,
  grid style={cGrid, thin},
  clip=false,
  xtick pos=left,
  ytick pos=right,
]
\addplot[cVanilla, thick, dashed] coordinates {
  (0,0.00) (257,0.05) (515,0.09) (772,0.13) (1028,0.16) (1284,0.20)
  (1539,0.23) (1794,0.27) (2048,0.30) (2301,0.34) (2554,0.37) (2807,0.40)
  (3059,0.44) (3310,0.47) (3560,0.51) (3812,0.54) (4061,0.58) (4310,0.61)
  (4558,0.65) (4805,0.68) (5052,0.71) (5299,0.75) (5546,0.78) (5792,0.82)
  (6038,0.85) (6284,0.88) (6529,0.92) (6774,0.95) (7019,0.98) (7264,1.02)
  (7508,1.05) (7753,1.08) (7996,1.12) (8240,1.15) (8482,1.18) (8724,1.22)
  (8967,1.25) (9209,1.28) (9451,1.32) (9693,1.35) (9934,1.38) (10176,1.42)
  (10418,1.45) (10659,1.48) (10899,1.52) (11139,1.55) (11379,1.58) (11620,1.62)
  (11859,1.65) (12099,1.68) (12338,1.71) (12576,1.75) (12815,1.78) (13053,1.81)
  (13291,1.85) (13528,1.88) (13766,1.91) (14003,1.94) (14240,1.98) (14477,2.01)
  (14715,2.04) (14951,2.07) (15187,2.11) (15423,2.14) (15659,2.17)
};
\addplot[cMemento, very thick] coordinates {
  (0,0.00) (261,0.05) (526,0.09) (791,0.13) (1054,0.16) (1317,0.20)
  (1580,0.24) (1843,0.27) (2104,0.31) (2365,0.34) (2625,0.38) (2875,0.07)
  (3140,0.11) (3404,0.15) (3667,0.18) (3929,0.22) (4191,0.25) (4443,0.08)
  (4708,0.11) (4973,0.15) (5236,0.18) (5499,0.22) (5761,0.26) (6022,0.29)
  (6283,0.33) (6544,0.36) (6803,0.40) (7062,0.44) (7310,0.08) (7571,0.12)
  (7835,0.15) (8099,0.19) (8360,0.23) (8621,0.26) (8884,0.30) (9144,0.33)
  (9404,0.37) (9658,0.12) (9921,0.16) (10184,0.20) (10446,0.23) (10711,0.27)
  (10971,0.31) (11233,0.34) (11493,0.38) (11753,0.41) (12011,0.45) (12270,0.48)
  (12515,0.13) (12779,0.16) (13040,0.20) (13304,0.24) (13567,0.27) (13828,0.31)
  (14089,0.34) (14349,0.38) (14610,0.42) (14869,0.45) (15127,0.49) (15386,0.52)
  (15643,0.56) (15900,0.59) (16154,0.63) (16408,0.66) (16661,0.70) (16915,0.73)
  (17167,0.77) (17383,0.22) (17646,0.25)
};
\draw[cMemento!60!black, densely dashed, semithick] (axis cs:0,0.77) -- (axis cs:18000,0.77);
\node[font=\scriptsize, text=cMemento!60!black, anchor=south east] at (axis cs:14500,0.79) {Peak KV};
\node[font=\scriptsize\bfseries, anchor=north] at (rel axis cs:0.5,-0.28) {Qwen3-8B, AIME24 P2};
\end{axis}

\coordinate (scalinganchor) at (0, -8.3);
\coordinate (chartanchor) at (4.3, -8.3);
\coordinate (kvanchor) at (13.6, -8.3);

\begin{axis}[
  at={(scalinganchor)},
  anchor=north west,
  width=5.0cm,
  height=5.2cm,
  xmode=log,
  xmin=800, xmax=130000,
  ymin=5, ymax=35,
  xlabel={Dataset Size},
  xlabel style={font=\scriptsize\bfseries, at={(0.5,-0.08)}},
  ylabel={AIME25 (\%)},
  ylabel style={font=\footnotesize\bfseries, at={(0.12,0.5)}},
  x tick label style={font=\scriptsize},
  y tick label style={font=\scriptsize},
  xtick={1000,10000,100000},
  xticklabels={1K,10K,100K},
  ytick={10,20,30},
  ymajorgrids=true,
  grid style={cGrid, thin},
  clip=false,
  xtick pos=left,
  ytick pos=left,
  title={\footnotesize\bfseries Qwen2.5-7B-Inst.},
  title style={at={(0.5,-0.42)}},
]
\addplot[thick, cVanilla, dashed, mark=square*, mark size=1.5pt] coordinates {
  (1000,8.67) (3000,16.00) (10000,20.00) (31000,23.33) (100000,32.00)
};
\addplot[very thick, cMemento, mark=triangle*, mark size=2pt] coordinates {
  (1000,9.33) (3000,13.33) (10000,19.33) (31000,21.33) (100000,27.33)
};
\end{axis}

\begin{axis}[
  at={(chartanchor)},
  anchor=north west,
  name=accplot,
  width=10.2cm,
  height=5.2cm,
  ybar=1.5pt,
  bar width=5pt,
  ymin=20, ymax=88,
  ylabel={Accuracy (\%)},
  ylabel style={font=\footnotesize\bfseries, at={(0.06,0.5)}},
  y tick label style={font=\footnotesize},
  ytick={20, 30, 40, 50, 60, 70, 80},
  xtick={2.2, 6.2, 10.2},
  xticklabels={Qwen3-8B, Phi-4-r (14B), Qwen3-32B},
  x tick label style={font=\footnotesize\bfseries, yshift=-4pt},
  xmin=0, xmax=12.2,
  enlarge x limits=0.02,
  every axis plot/.append style={fill opacity=0.85},
  nodes near coords,
  every node near coord/.append style={font=\scriptsize, rotate=90, anchor=west, xshift=1pt, yshift=0pt},
  grid=none,
  extra x ticks={4.2, 8.2},
  extra x tick labels={},
  extra x tick style={grid=major, major grid style={draw=solBase1!40, dashed}},
  clip=false,
  xtick pos=left,
  ytick pos=left,
  xtick style={draw=none},
  ytick style={draw=none},
]
\addplot[fill=cVanilla, draw=cVanilla!80!black] coordinates {
  (1.0, 66.8) (2.2, 61.4) (3.4, 73.1)
  (5.0, 71.7) (6.2, 64.1) (7.4, 64.1)
  (9.0, 75.2) (10.2, 65.9) (11.4, 78.0)
};
\addplot[fill=cMemento, draw=cMemento!80!black,
  nodes near coords={\ifnum\coordindex<3\relax\else\pgfmathprintnumber{\pgfplotspointmeta}\fi}] coordinates {
  (1.0, 57.3) (2.2, 55.8) (3.4, 66.5)
  (5.0, 67.6) (6.2, 61.6) (7.4, 61.8)
  (9.0, 72.6) (10.2, 62.1) (11.4, 74.0)
};
\pgfplotsextra{
  \fill[fill=cMemento, fill opacity=0.4, draw=cMemento!80!black, line width=0.4pt]
    ([xshift=0.75pt]{axis cs:1.0, 57.3}) rectangle ([xshift=5.75pt]{axis cs:1.0, 64.9});
  \node[font=\scriptsize, rotate=90, anchor=west] at ([xshift=3.25pt]{axis cs:1.0, 64.9}) {64.9};

  \fill[fill=cMemento, fill opacity=0.4, draw=cMemento!80!black, line width=0.4pt]
    ([xshift=0.75pt]{axis cs:2.2, 55.8}) rectangle ([xshift=5.75pt]{axis cs:2.2, 62.9});
  \node[font=\scriptsize, rotate=90, anchor=west] at ([xshift=3.25pt]{axis cs:2.2, 62.9}) {62.9};

  \fill[fill=cMemento, fill opacity=0.4, draw=cMemento!80!black, line width=0.4pt]
    ([xshift=0.75pt]{axis cs:3.4, 66.5}) rectangle ([xshift=5.75pt]{axis cs:3.4, 68.8});
  \node[font=\scriptsize, rotate=90, anchor=west] at ([xshift=3.25pt]{axis cs:3.4, 68.8}) {68.8};
}
\node[font=\scriptsize, text=solBase01, anchor=north] at (axis cs:1.0,21) {AIME26};
\node[font=\scriptsize, text=solBase01, anchor=north] at (axis cs:2.2,21) {GPQ};
\node[font=\scriptsize, text=solBase01, anchor=north] at (axis cs:3.4,21) {LCB};
\node[font=\scriptsize, text=solBase01, anchor=north] at (axis cs:5.0,21) {AIME26};
\node[font=\scriptsize, text=solBase01, anchor=north] at (axis cs:6.2,21) {GPQ};
\node[font=\scriptsize, text=solBase01, anchor=north] at (axis cs:7.4,21) {LCB};
\node[font=\scriptsize, text=solBase01, anchor=north] at (axis cs:9.0,21) {AIME26};
\node[font=\scriptsize, text=solBase01, anchor=north] at (axis cs:10.2,21) {GPQ};
\node[font=\scriptsize, text=solBase01, anchor=north] at (axis cs:11.4,21) {LCB};
\end{axis}

\begin{axis}[
  at={(kvanchor)},
  anchor=north west,
  name=kvplot,
  width=5.0cm,
  height=5.2cm,
  ybar=1.5pt,
  bar width=5pt,
  ymin=0, ymax=3.2,
  ylabel={Peak KV (GB)},
  ylabel style={font=\footnotesize\bfseries, at={(0.88,0.5)}},
  y tick label style={font=\footnotesize},
  ytick={0, 1, 2, 3},
  yticklabel pos=right,
  xtick={1, 2, 3},
  xticklabels={Qwen3-8B, Phi-4-r (14B), Qwen3-32B},
  x tick label style={font=\scriptsize\bfseries, rotate=25, anchor=north east},
  xmin=0.3, xmax=3.7,
  every axis plot/.append style={fill opacity=0.85},
  nodes near coords,
  every node near coord/.append style={font=\scriptsize, rotate=90, anchor=west, xshift=1pt, yshift=0pt,
    /pgf/number format/.cd, fixed, fixed zerofill, precision=1},
  grid=none,
  clip=false,
  xtick pos=left,
  ytick pos=right,
  xtick style={draw=none},
  ytick style={draw=none},
]
\addplot[fill=cVanilla, draw=cVanilla!80!black] coordinates {(1, 1.80) (2, 2.09) (3, 2.60)};
\addplot[fill=cMemento, draw=cMemento!80!black]
  coordinates {(1, 0.73) (2, 0.93) (3, 1.25)};
\end{axis}

\node[anchor=north, font=\footnotesize] at ($(accplot.south)!0.5!(kvplot.south)+(-2.7,-0.6cm)$) {
  \tikz{
    \fill[cVanilla, draw=cVanilla!80!black, fill opacity=0.85] (0,0) rectangle (0.25,0.25);
    \node[right, font=\footnotesize] at (0.3, 0.12) {Base};
    \fill[cMemento, draw=cMemento!80!black, fill opacity=0.85] (1.2,0) rectangle (1.45,0.25);
    \node[right, font=\footnotesize] at (1.5, 0.12) {Memento};
    \fill[cMemento, draw=cMemento!80!black, fill opacity=0.4] (3.2,0) rectangle (3.5,0.25);
    \node[right, font=\footnotesize] at (3.6, 0.12) {+RL};
  }
};

\end{tikzpicture}
}
\end{figure}

\clearpage

\addtocounter{figure}{-1}
\begin{figure}[t]
    \caption{\textbf{\memento{} overview.} \textbf{Top left:} SFT data
    generation pipeline. Starting from a reasoning trace, we split text into
    sentences, use an LLM to score each sentence boundary as a potential stopping point, optimize boundary selection
    algorithmically, and finally use an LLM to summarize each block into a memento. \textbf{Top
    right:} Sparse attention during inference. The model produces alternating
    thinking blocks ($T_i$) and mementos ($M_i$); once a memento is generated,
    KV cache entries for its preceding thinking block are physically removed.
    The sawtooth KV trace shows the resulting memory pattern. \textbf{Bottom
    left:} Data scaling on Qwen2.5-7B-Instruct (AIME25); \memento{} scales similarly
    to vanilla SFT on OpenThoughts \citep{openthoughts2025} across dataset sizes (\Cref{fig:sample_complexity}). \textbf{Bottom
    center:} Accuracy across three models on AIME26, GPQA-D, and LCB; lighter
    bars show RL gains for Qwen3-8B (\Cref{sec:rl}). \textbf{Bottom right:} Peak
    KV cache (GB), averaged across all benchmark categories, showing
    ${\sim}2$--$2.5\times$ reduction. 
    }
    \label{fig:memento-mechanism}
\end{figure}

\section{Introduction}
\label{sec:intro}

Large language models routinely reason at test time, spending  thousands of
tokens working through a problem before arriving at an
answer~\citep{openai_o1,deepseekr1,qwq2024}. This has led to dramatic gains on
hard reasoning benchmarks, but has also created a new problem: {\it
reasoning models have no built-in mechanism to organize their chain-of-thought}. A 32K-token CoT is a flat, unstructured stream, and there is no
mechanism for the model to mark an intermediate result as worth keeping, or to
compress a long derivation into a compact conclusion that it can reference
later. Every past token sits in the attention window at
equal cost, and the model has learned no way to drop it.

We propose \memento{}, an approach that trains models to segment their chain of
thought into semantically coherent blocks and, after each block, generate a
compressed summary that we call a
\textit{memento}\footnote{Named after the 2000 film by Christopher Nolan, in
which the protagonist compensates for anterograde amnesia by maintaining
external memory artifacts---an analogy for a model that must reason from
compressed summaries of its own past thinking.}. Rather than a summary in the
usual expository sense, each memento is a minimal record of a reasoning block,
preserving its conclusions, intermediate values, and key directional decisions
in as few tokens as possible. Once a memento is produced, the preceding
thinking block is masked within a single, uninterrupted generation call via a
custom  vLLM based engine~\citep{kwon2023vllm}: at every subsequent step, the model
attends only to past mementos and the current block. Each block is compressed
to a ${\sim}5$--$20\times$ smaller size on average, so the effective context
the model attends to is a fraction of the full trace.
The inference procedure is illustrated in \Cref{fig:memento-mechanism} (right).

Crucially, because masking happens in-place rather than by restarting
generation, the KV cache entries of each memento are computed while the full
block is still in context and retained after the block is masked. Despite the
fact that the original thinking tokens are gone, they remain implicitly present
in the representations of the memento KV states 
This creates a \emph{dual information stream}: the explicit memento text plus an
implicit representational channel through the cached KV states. We verify this
experimentally: recomputing memento KVs without block context reduces accuracy
by 15\,pp on AIME'24 (\Cref{sec:kv-ablation}), and our probing experiments
(\Cref{sec:kv-probing}) demonstrate that block information not present in the
memento text is still recoverable from the memento KV states, with upper layers
carrying the most task-relevant signal.

A prime concern is that compression could destroy reasoning capacity. Our
experiments across three model families, i.e., Qwen3 (8B/32B), Phi-4-reasoning
(14B), and Olmo-3-7B-Think, show that this is not the case. On AIME'26,
Qwen3-32B with \memento{} loses just 2.6\,pp (72.6\% vs.\ 75.2\%) while
cutting peak KV cache by ${\sim}2\times$. Averaged across the five benchmark groups in \Cref{tab:main-sft}, the
accuracy gap is 3.5\,pp at 32B and 6.3\,pp at 8B, and we observe that the gap shrinks
with scale within the same model family, suggesting that larger models manage
compressed context more effectively. Further, we show that RL fine-tuning can close the remaining gap, enabled
by native block masking support in our vLLM fork.

To train \memento{} models, we construct \openmementos{}, a public dataset of
228K segmented and summarized reasoning traces derived from
OpenThoughts~\citep{openthoughts2025}. Building this dataset required solving a
non-trivial annotation problem: reasoning traces lack natural segment
boundaries, and naive summarization loses the precise intermediate state a
model needs to continue. Our pipeline combines LLM-scored boundary detection,
algorithmic segmentation, and iterative judge-refined
summarization to produce training data where each memento is both faithful and
minimal.
Our key contributions can be summarized as follows:
\begin{enumerate}[leftmargin=*, topsep=-1pt,itemsep=-1ex,partopsep=1ex,parsep=1ex]
    \item \textbf{\openmementos{}}: a 228K-trace public dataset of segmented,
    summarized reasoning chains, with the annotation pipeline and code.
    \item \textbf{Demonstration at scale} across three model families (Qwen3
    8B/32B, Phi-4-reasoning 14B, Olmo-3-7B-Think), showing that models internalize
    summarization as a learned capability while preserving reasoning accuracy
    at 2--3$\times$ peak KV cache reduction. 
    \item \textbf{Native block masking in vLLM}, a custom fork that supports
    in-place KV cache masking within a single generation call: a key
    infrastructure bottleneck for both inference and training with
    \memento{}. This enables \textbf{RL fine-tuning with block masking},
    which allows us to close the accuracy gap.
    \item \textbf{The dual information stream}: we identify and verify that
    memento KV states encode information from masked blocks which is a mechanism
    absent in restart-based approaches. Removing this channel drops accuracy
    by 15\,pp on AIME'24.
\end{enumerate}

\section{Related Work}
\label{sec:related}

Context management for long-running models is typically handled through external
infrastructure: separate summarizers, memory modules, or orchestration
logic~\citep{resum2025,amem2025}. We instead focus on teaching models to manage
their own context during reasoning, as an internal capability rather than an
external system. Among works that do train models for context management,
MemAgent~\citep{memagent2025} reads text in segments and updates a fixed-size
memory via an overwrite strategy trained with RL, and MEM1~\citep{mem1_2025}
takes a similar approach for multi-turn agent interactions, maintaining a
compact internal state across tool calls and environment observations. Both primarily focus on managing
external information, retrieved documents, tool outputs, and environment
observations, rather than complex reasoning chains typically observed in models
solving hard math or coding problems.

The most closely related works to ours train models to compress their own
reasoning output in domains such as math: InftyThink~\citep{inftyThink2025},
InftyThink+~\citep{inftyThinkPlus2026},
Accordion-Thinking~\citep{accordionThinking2026}, and The Markovian
Thinker~\citep{markovianThinker2025}. These works break reasoning into
(potentially variable-length) chunks and train the model to continue from a
compact textual carryover rather than the full prior chunk. InftyThink relies on
SFT alone, while InftyThink+, Accordion-Thinking, and The Markovian Thinker
additionally use RL to improve this chunk-to-chunk carryover. All of these works
operate at the text level: after each reasoning chunk, the future context is
rebuilt from compact text alone, discarding the original reasoning tokens and
their KV cache representations.  \memento{} differs in that it retains summary
KV entries via in-engine attention masking rather than text-level context
rebuilding, creating a dual stream of information: the explicit memento text,
and the implicit representations encoded in the memento's KV cache. Our
experiments demonstrate that useful information is stored in these KV
entries---dropping them and recomputing memento KVs without block context
reduces AIME24 accuracy by 15 percentage points (\Cref{sec:kv-ablation}).

Another related work (PENCIL~\citep{pencil2025}) explores learned context
management for models trained from scratch on synthetic tasks. PENCIL teaches
models to erase intermediate reasoning via reduction rules, enabling small (25M
parameter) models to solve 3-SAT problems and Einstein's Puzzle (a
multi-constraint logic deduction) with 2K context length. PENCIL demonstrates
that the potential of context management extends beyond memory and throughput
efficiency; it can enable models to solve significantly harder problems than
standard CoT permits.  Whether such gains can be achieved at larger scale and in
settings such as math and coding problems, for instance through \memento{}-style
compression, is an interesting direction for future work.

A closely related line of work compresses reasoning chunks into learned
\emph{gist tokens}, i.e., special-purpose tokens whose KV cache entries encode a
compressed representation of the preceding chunk, after which the original
tokens are evicted~\citep{lightThinker2025,monea2025breadcrumbs}. A limitation
of gist-token approaches is that interpretability is lost: the compressed state
is encoded entirely in hidden representations. \memento{}'s summaries are
natural-language text, preserving interpretability while still achieving
compression.

A complementary line of work aims to reduce the memory footprint of reasoning
through shorter traces or direct KV cache compression. These include methods
that train models to skip low-importance
tokens~\citep{tokenSkip2025,stepEntropy2025}, train on compressed reasoning
traces~\citep{c3ot2024,tokenSqueeze2025}, or steer the model to produce shorter
traces via RL~\citep{thinkPrune2025,gfpo2025}. Other works replace explicit
reasoning tokens with latent
representations~\citep{codi2025,colar2025,coconut2024}.  At the KV cache level,
inference-time methods such as ThinKV~\citep{ramachandran2025thinkv},
R-KV~\citep{cai2025r}, LazyEviction~\citep{lazyEviction2025}, and Reasoning Path
Compression~\citep{song2025reasoning} prune or quantize cache entries based on
attention patterns, while architectures such as sliding-window
attention~\citep{beltagy2020longformer} limit the attention span by
design~\citep{olmo3}.  These architecture-level approaches are orthogonal to \memento{} and many
can likely compose with it: for example, we show that Olmo-3-7B-Think that has
sliding-window attention can be effectively combined with \memento{}.

Finally, we note a naming overlap with recent work on memory-augmented
agents: \citet{zhou2025mementoagents} and \citet{wang2026memento2} also use the
name \emph{Memento} for systems that maintain external episodic memory for
inference-time agent adaptation.

\section{\openmementos{} Dataset}
\label{sec:dataset}
\label{sec:format}
\label{sec:pipeline}

Training models to simultaneously reason and manage context requires high-quality annotated data: reasoning traces segmented into semantically coherent blocks, paired with dense summaries. 
A core challenge is that typical reasoning traces are not a sequence of independent thoughts; they are a continuous stream without ``natural'' boundaries. Here we describe our data generation pipeline (\Cref{fig:memento-mechanism}, top left), which takes raw CoT traces and produces structured traces annotated with mementos.

\paragraph{Design rationale.}
Each stage in our pipeline reflects a deliberate design choice. Early on, we tried having a frontier LLM directly segment CoTs into semantically coherent blocks. This failed. Even strong models struggle with a combinatorial optimization problem that considers all possible partitions, requiring simultaneous reasoning about block coherence, size balance, and semantic boundaries. 

To simplify, we factored the problem: boundary \emph{scoring} asks a local question (``is this a good place to slice the CoT?''), which LLMs handle well, while the global optimization of boundary selection given the LLM scores is handled algorithmically. We then use an LLM judge to grade the quality of summarization. Defining ``good summary'' programmatically is difficult, but LLMs can give reasonable scores against explicit rubrics. We then chose iterative refinement of mementos over single-shot summarization as initial mementos often miss key formulas or intermediate values; a zero-shot approach achieves only 28\% pass rate (scoring $\geq 8/10$ on our rubric), while the judge-feedback loop brings this to 92\%. 

\paragraph{Stage 0: Seed selection.}
We source 228K reasoning traces from OpenThoughts-v3~\citep{openthoughts2025}, a widely-adopted dataset of CoT traces generated by QwQ-32B, and process them through our annotation pipeline to produce the final \openmementos{} dataset. While we could regenerate traces using stronger teachers, we leverage OpenThoughts because: (1) substantial effort has already been invested in generating these traces at scale; (2) the OpenThinker-3 paper provides extensive baselines, making it an ideal testbed; and (3) our hypothesis that traces from a relatively strong reasoner (QwQ-32B) should transfer across model families is confirmed empirically (works for Qwen3, Phi-4, Olmo 3).

\paragraph{Stage 1: Sentence splitting.}
We partition reasoning traces into atomic ``sentences'': complete, modular thoughts that can stand alone. Code blocks and multi-line math are detected and protected as atomic units. Plain text is split at sentence boundaries (avoiding splits inside parentheses, inline math, or abbreviations). Finally, we merge logically connected fragments: sentences ending with colons are attached to the next; continuation words (Therefore, Thus, So) signal a need to merge with the preceding sentence; short fragments ($<$5 tokens) and consecutive math expressions are consolidated. This structure-aware splitting reduces candidate boundaries by ${\sim}2\times$ vs.\ naive sentence splitting (397 $\to$ 187 per trace on average).

\paragraph{Stage 2: Boundary scoring.}
An LLM judge (GPT-5.x in our case) evaluates each inter-sentence boundary as a potential breakpoint, scoring from 0 (mid-thought, would disrupt flow) to 3 (major transition, natural chapter boundary). The prompt instructs: ``Never score 2--3 mid-calculation. Score 0 if previous sentence ends with `:' or `='.'' Because traces contain hundreds of boundaries, we score them in batches: the judge sees a window of consecutive sentences and scores each boundary within that window.  See  \Cref{fig:scoring-example} for an example of boundary scoring.

\begin{figure}[ht!]
\centering
\begin{tcolorbox}[colback=findingbg, colframe=findingborder, boxrule=0.4pt, arc=2pt, left=5pt, right=5pt, top=3pt, bottom=3pt, fontupper=\footnotesize, title={\footnotesize\textbf{Boundary scoring example} --- complex-plane geometry trace}, coltitle=black, colbacktitle=findingbg]
\begin{tabular}{@{}p{0.82\linewidth}r@{}}
\textbf{[177]} \textit{``Hence, this approach again leads to the only solution being zero.''} & {0.0} \\[2pt]
\textbf{[178]} \textit{``Therefore, unless I made a mistake \ldots the answer is the trivial solution.''} & 1.5 \\[2pt]
\textbf{[179]} \textit{``Perhaps the problem's answer is indeed zero, so the distance is zero.''} & \textbf{2.0} \\[2pt]
\textbf{[180]} \textit{``Alternatively, maybe I made an error in interpretation.''} & \textbf{3.0} \\[2pt]
\textbf{[181]} \textit{``Let me check with an example.''} & 0.5 \\[2pt]
$\cdots$ & \\[-2pt]
\textbf{[185]} \textit{``Suppose I take $r{=}2$. Then $u{=}5$, $v{=}\pm\!\sqrt{7}$\,\ldots''} & {0.0} \\[2pt]
\textbf{[186]} \textit{``But $\sqrt{56} \approx 7.48$, $8\sqrt{2} \approx 11.31$, which are not equal.''} & {0.0} \\[2pt]
\end{tabular}
\end{tcolorbox}
\caption{Boundary scoring assigns each inter-sentence boundary a score from 0 to 3.  Sentence~179 wraps up a conclusion (score~2.0); sentence~180 pivots to a new strategy (score~3.0---the strongest possible boundary); sentences 185--186 are mid-derivation (scores~0.0---never split here). The segmentation optimizer selects cuts at high-scoring transitions.}
\label{fig:scoring-example}
\end{figure}

\paragraph{Stage 3: Segmentation.}
Given $n$ sentences with boundary scores $s_1, \ldots, s_{n-1}$, we partition the trace into $K$ contiguous blocks by maximizing
$\frac{1}{K}\sum_{b \in \text{boundaries}} s_b
  \;-\;\lambda\;\cdot\;
  \sigma(\ell_1, \ldots, \ell_K)\,/\,\mu(\ell_1, \ldots, \ell_K)$,
subject to every block containing at least 200 tokens. Here, $b$ is the boundary positions,  $\ell_k$ is the token count of the $k$-th block, $\mu$ and $\sigma$ are the mean and standard deviation of the block sizes, and $\lambda = 0.5$. We optimize over valid partitions and values of $K$.
We observe that the first term rewards cutting at strong semantic boundaries: a partition that places cuts at score-3 transitions (major topic changes) scores higher than one cutting low-score transitions (mid-derivation). 
The second term penalizes uneven block sizes. The coefficient of variation $\sigma/\mu$ is scale-invariant, so the penalty applies equally whether the trace is 5K or 50K tokens. Without this term, the optimizer would greedily cut at the top-scoring boundaries regardless of balance, often producing one very long block and several tiny ones.

\paragraph{Stage 4: Iterative memento generation.}
Each block is compressed into a memento: a terse state representation that preserves all logically relevant information (definitions, formulas, intermediate values, chosen strategies, rejected approaches) needed for subsequent blocks to succeed. Unlike traditional summarization, the goal is \emph{``lossless compression'' of reasoning state}: mementos must capture everything a future reasoning step might need, targeting ${\sim}15$--$25$\% of original tokens while being purely extractive (no new derivations or error corrections).

\paragraph{Compressor.} The compressor call (using GPT-5.x) receives all blocks and produces one memento per block using terse notation (semicolon-separated clauses, ``name: value'' pairs, compact math). The prompt instructs: ``You are a STATE-COMPRESSOR. Minimize tokens subject to fully capturing all logically relevant information.''

\paragraph{Judge.} A separate LLM call (again using GPT-5.x) evaluates each memento on a 0--10 scale across six dimensions: (1) formulas extracted verbatim (0--3), (2) numerical values preserved (0--2), (3) methods explicitly named (0--2), (4) validation included (0--1), (5) no hallucinations (0--1), and (6) result-first structure (0--1). If the score falls below the acceptance threshold $\tau = 8$ (out of 10), the judge provides actionable feedback (e.g., ``Missing formula: $K^2 - 3K + 3$'') used to refine the memento; mementos scoring $\geq \tau$ are accepted without refinement. We use max $T = 2$ iterations, i.e., Compressor $\rightarrow$ Judge 
$\rightarrow$ Compressor   $\rightarrow$  Judge.  Adding more iterations does not significantly improve memento quality and results in longer mementos. See \Cref{fig:refinement-example} for an example of the iterative improvement of mementos.

\begin{tcolorbox}[finding]
\textbf{Iterative refinement is essential.} Single-pass memento generation achieves only 28\% pass rate ($\geq 8/10$ judge score). Two iterations of judge feedback bring this to 92\%. Initial mementos often miss critical formulas or intermediate values that downstream blocks need to reason correctly. 
\end{tcolorbox}

\begin{figure}[ht!]
    \centering
    \includegraphics[width=\textwidth]{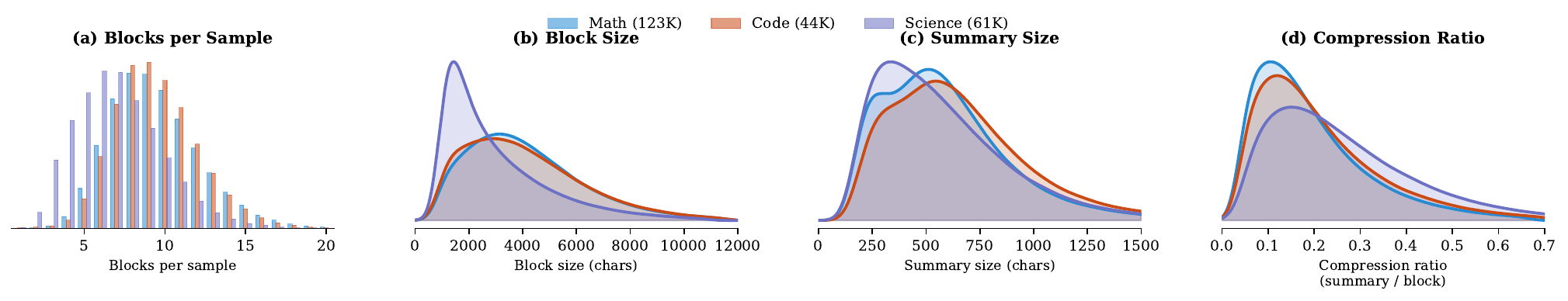}
    \caption{\footnotesize \textbf{\openmementos{} dataset distributions by domain} (228K samples). \textbf{(a)}~Math and code have $\sim$9 blocks/sample; science has $\sim$7. \textbf{(b)}~Block sizes range from 2.3K (science) to 3.8K (math) chars. \textbf{(c)}~Summary sizes cluster around 509--603 chars across all domains, indicating a stable compression target. \textbf{(d)}~Math achieves the tightest compression ratio (median 0.16) due to its larger blocks.}
    \label{fig:training-data-boxplots}
\end{figure}

\textbf{Dataset statistics.}
\Cref{fig:training-data-boxplots} characterizes the final \openmementos{} dataset (228K samples: 54\% math, 19\% code, 27\% science). Math and code traces produce more blocks per sample (median~9) than science (median~7), and math has the largest blocks (median~3.8K chars). Summary sizes are remarkably stable across domains (median~509--603 chars), yielding median compression ratios of 0.16 (math), 0.18 (code), and 0.23 (science)---corresponding to ${\sim}4$--$6\times$ block-level compression. Across the full dataset, the average block contains ${\sim}1{,}150$ tokens and the average memento ${\sim}194$ tokens, for a trace-level compression of ${\sim}6\times$ (from ${\sim}10{,}900$ block tokens to ${\sim}1{,}850$ memento tokens per trace).

\section{Training the \memento{} Models}
\label{sec:experiments}

We use a two-stage SFT procedure on \openmementos{} that separates format
learning from context management.
The intuition follows standard curriculum
learning: we first let the model acquire the block-memento format under normal
conditions, then introduce the harder constraint of operating without access to
masked content, see \Cref{sec:multistage} for an ablation.\\
\textbf{Stage 1: Full Attention}: Standard causal attention over all tokens. 
    Loss is computed on all tokens, including thinking blocks, mementos, special
    tokens, and the final answer. The model learns the block-memento format
    without any context management pressure.\\
\textbf{Stage 2: Memento Attention}: After each completed memento, \emph{the
    preceding thinking block is masked from all subsequent attention}. 
    This teaches the model to produce self-contained
    mementos that carry all information needed for downstream reasoning.

The attention mask implementation maintains a \textbf{block cache} that tracks whether each token belongs to a thinking block, summary, or other content. When \texttt{<|summary\_end|>} is generated, the preceding block is marked as completed and masked from future attention. For training, this mask is constructed upfront as a dense matrix; for inference, the block cache is stateful across autoregressive steps.
Four special tokens (\texttt{<|block\_start|>}, \texttt{<|block\_end|>}, \texttt{<|summary\_start|>}, \texttt{<|summary\_end|>}) are added and initialized as the mean embedding of semantically related existing tokens (e.g., \texttt{<|block\_start|>} from \textit{block, start, begin, section, step}) plus small Gaussian noise.

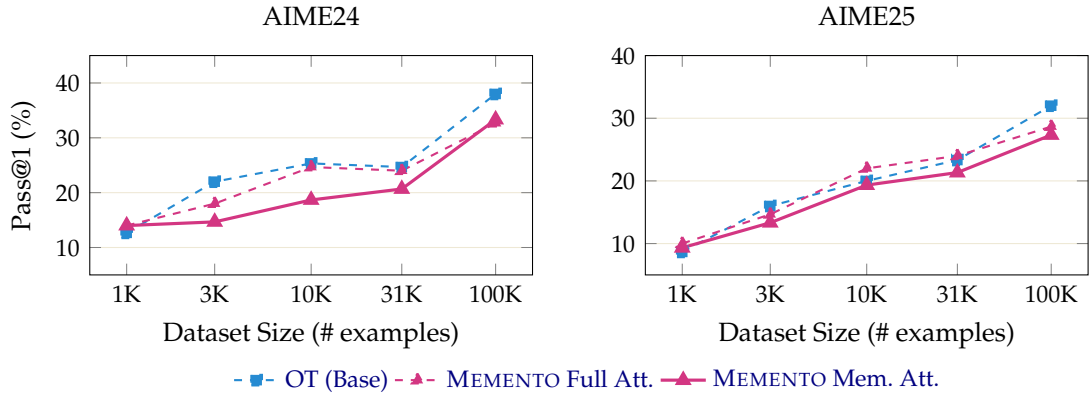
\begin{figure}[H]
    \centering
    \begin{tikzpicture}
      \begin{groupplot}[
        group style={group size=2 by 1, horizontal sep=1.5cm},
        width=0.42\textwidth, height=4.5cm,
        xlabel={Dataset Size (\# examples)},
        xlabel style={font=\small},
        ylabel style={font=\small},
        xmode=log,
        xtick={1000,3000,10000,31000,100000},
        xticklabels={1K,3K,10K,31K,100K},
        tick label style={font=\footnotesize},
        ymajorgrids=true,
        grid style={cGrid, thin},
        legend style={font=\footnotesize, draw=none},
      ]
        \nextgroupplot[
          title={\small AIME24}, title style={at={(0.5,1.02)}},
          ylabel={Pass@1 (\%)},
          ymin=5, ymax=45,
          legend to name=legscaling,
          legend columns=3,
        ]
        \addplot[thick, cVanilla, dashed, mark=square*, mark size=2pt] coordinates {
          (1000,12.67) (3000,22.00) (10000,25.33) (31000,24.67) (100000,38.00)
        }; \addlegendentry{OT (Base)}
        \addplot[thick, cMemento, dashed, mark=triangle*, mark size=2.5pt] coordinates {
          (1000,14.00) (3000,18.00) (10000,24.67) (31000,24.00) (100000,32.67)
        }; \addlegendentry{\memento{} Full Att.}
        \addplot[very thick, cMemento, mark=triangle*, mark size=2.5pt] coordinates {
          (1000,14.00) (3000,14.67) (10000,18.67) (31000,20.67) (100000,33.33)
        }; \addlegendentry{\memento{} Mem.\ Att.}

        \nextgroupplot[
          title={\small AIME25}, title style={at={(0.5,1.02)}},
          ymin=5, ymax=40,
        ]
        \addplot[thick, cVanilla, dashed, mark=square*, mark size=2pt] coordinates {
          (1000,8.67) (3000,16.00) (10000,20.00) (31000,23.33) (100000,32.00)
        };
        \addplot[thick, cMemento, dashed, mark=triangle*, mark size=2.5pt] coordinates {
          (1000,10.00) (3000,14.67) (10000,22.00) (31000,24.00) (100000,28.67)
        };
        \addplot[very thick, cMemento, mark=triangle*, mark size=2.5pt] coordinates {
          (1000,9.33) (3000,13.33) (10000,19.33) (31000,21.33) (100000,27.33)
        };

      \end{groupplot}
      \node at ($(group c1r1.south)!0.5!(group c2r1.south) + (0,-1.4)$) {\ref{legscaling}};
    \end{tikzpicture}
    \caption{\textbf{Training data scaling.} Pass@1 accuracy on AIME24 and AIME25 for Qwen2.5-7B-Instruct fine-tuned on 1K--100K examples. All methods improve monotonically with data size.}
    \label{fig:sample_complexity}
\end{figure}

\paragraph{Data Scaling.} When training ``from scratch'' a non-reasoning model
(Qwen2.5-7B-Instruct), data scaling follows a similar monotonic trend as
standard reasoning SFT~\citep{openthoughts2025}.
We study how performance scales with the amount of \openmementos{} training data
by fine-tuning Qwen2.5-7B-Instruct on varying amounts of data (1K, 3K, 10K, 31K,
100K examples), comparing vanilla OpenThoughts (OT), \openmementos{} with full
attention (OM/Full), and \openmementos{} with memento attention (OM/Mem).
\label{sec:data-scaling}
As shown in \Cref{fig:sample_complexity}, all three methods improve
monotonically from 1K to 100K.  OT achieves the highest accuracy across all data
budgets, while OM/Full and OM/Mem trail by a modest margin.

\paragraph{Fine-Tuning Reasoning Models.} 
When starting with already strong reasoning models, we found that training for
more epochs on fewer samples is more effective than training on more samples
with fewer epochs. We train on 31K samples from the 228K \openmementos{} pool
with 32K sequence length, as further gains are more effectively attainable
through reinforcement learning (\Cref{sec:rl}) rather than additional supervised
data. We release the full 228K dataset to support future research in both
directions.

\paragraph{Hyperparameters.} We use the same hyperparameters for all models and stages.
Key hyperparameters: learning rate $8 \times 10^{-5}$, cosine schedule
with 5\% warmup, 5 epochs per stage, AdamW ($\beta_1{=}0.9$, $\beta_2{=}0.999$),
no weight decay, bfloat16 precision, batch size 512, and 32 B200 GPUs.
See \Cref{appendix:training} for details.

\begin{table}[ht!]
\centering
\caption{\footnotesize
\memento{} achieves 2--3$\times$ peak KV reduction on models with
uniform attention layers while maintaining
strong reasoning performance; RL on top of our Qwen3-8B further improves accuracy.
Olmo-3-7B shows more modest savings ($\sim$0.85--0.93$\times$) due to its hybrid
sliding-window architecture (\Cref{sec:olmo3-discussion}).
The $\Delta$ columns show changes of \memento{} and Mem.+RL relative
to Control: accuracy deltas are additive (pp); KV deltas are
multiplicative (Method\,/\,Control, so 0.39$\times$ means 61\% KV reduction).
\emph{Metrics.}
Accuracy~(\%): pass@1 accuracy.
Peak KV~(GB): peak KV cache size; determines the minimum memory
required to serve a request.
AUC KV~(GB$\cdot$ktok): area under the KV-occupancy-vs-token curve,
capturing total memory-time cost that penalizes both large footprints
and long generations (see \Cref{fig:kv-trace} for an illustration).
\emph{Rows.}
For each model we report Base (unmodified), Control (Base fine-tuned
on the same OpenThoughts source traces used to create \openmementos{},
but without block/memento annotations, for the same number of training
examples), and \memento{} (SFT on \openmementos{} with block masking);
for Qwen3-8B we additionally report \memento{}+RL (\Cref{appendix:rl-details}).
Olmo-3 \memento{} competition-math evaluations use 8 generations per problem
(vs.\ 64 for all other models) and are run with HuggingFace
Transformers rather than vLLM.
See \Cref{appendix:evaluation} for full benchmark and evaluation
details.}
\label{tab:main-sft}
\label{tab:kv-metrics}
\newcommand{\se}[1]{\textcolor{gray}{\scriptsize$_{#1}$}}
\newcommand{\cB}{\cellcolor{cVanilla!10}}
\newcommand{\cM}{\cellcolor{cMemento!10}}
\newcommand{\cR}{\cellcolor{cMemento!5}}
\newcommand{\cS}{\cellcolor{black!5}}
\newcommand{\dt}[1]{\textcolor{gray}{\scriptsize\hspace{1pt}#1}}
\newcommand{\dx}[1]{\textcolor{gray}{\scriptsize\hspace{1pt}\phantom{$-$}#1}}
\setlength{\tabcolsep}{2.5pt}
\scriptsize
\begin{tabular}{ll c rc rc rc rc rc}
\toprule
& & & \multicolumn{2}{c}{\textbf{AIME'26}} & \multicolumn{2}{c}{\textbf{Comp.\ Math}} & \multicolumn{2}{c}{\textbf{MATH-500}} & \multicolumn{2}{c}{\textbf{GPQA-D}} & \multicolumn{2}{c}{\textbf{LCB~v6}} \\
\cmidrule(lr){4-5}\cmidrule(lr){6-7}\cmidrule(lr){8-9}\cmidrule(lr){10-11}\cmidrule(lr){12-13}
& & & Val & \hspace{2pt}$\Delta$ & Val & \hspace{2pt}$\Delta$ & Val & \hspace{2pt}$\Delta$ & Val & \hspace{2pt}$\Delta$ & Val & \hspace{2pt}$\Delta$ \\
\midrule
\multirow{12}{*}{\rotatebox{60}{Qwen3-8B}}
 & \cB & \cB Acc     & \cB 66.8\se{1.1} & \cB & \cB 54.3\se{0.3} & \cB & \cB 90.5\se{0.9} & \cB & \cB 61.4\se{2.4} & \cB & \cB 73.1\se{1.0} & \cB \\
 & \cB Base & \cB Peak KV & \cB 2.41 & \cB & \cB 2.71 & \cB & \cB 0.84 & \cB & \cB 1.23 & \cB & \cB 1.76 & \cB \\
 & \cB & \cB AUC KV  & \cB 25.3 & \cB & \cB 30.9 & \cB & \cB 4.3 & \cB & \cB 6.6 & \cB & \cB 15.6 & \cB \\
\cmidrule(l){2-13}
 & \cS & \cS Acc     & \cS 64.7\se{1.1} & \cS & \cS 49.2\se{0.3} & \cS & \cS 89.7\se{1.0} & \cS & \cS 57.8\se{2.5} & \cS & \cS 70.0\se{1.0} & \cS \\
 & \cS Control & \cS Peak KV & \cS 2.59 & \cS & \cS 2.82 & \cS & \cS 0.88 & \cS & \cS 1.60 & \cS & \cS 1.89 & \cS \\
 & \cS & \cS AUC KV  & \cS 28.3 & \cS & \cS 33.1 & \cS & \cS 4.7 & \cS & \cS 11.8 & \cS & \cS 19.2 & \cS \\
\cmidrule(l){2-13}
 & \cM & \cM Acc     & \cM 57.3\se{1.1} & \cM \dt{$-$7.4} & \cM 45.1\se{0.3} & \cM \dt{$-$4.1} & \cM 90.1\se{0.9} & \cM \dt{$+$0.4} & \cM 55.8\se{2.5} & \cM \dt{$-$2.0} & \cM 66.5\se{1.0} & \cM \dt{$-$3.5} \\
 & \cM \memento{} & \cM Peak KV & \cM 1.02 & \cM \dx{0.39$\times$} & \cM 1.08 & \cM \dx{0.38$\times$} & \cM 0.41 & \cM \dx{0.47$\times$} & \cM 0.56 & \cM \dx{0.35$\times$} & \cM 0.60 & \cM \dx{0.32$\times$} \\
 & \cM & \cM AUC KV  & \cM 9.7 & \cM \dx{0.34$\times$} & \cM 10.7 & \cM \dx{0.32$\times$} & \cM 1.9 & \cM \dx{0.40$\times$} & \cM 4.0 & \cM \dx{0.34$\times$} & \cM 5.6 & \cM \dx{0.29$\times$} \\
\cmidrule(l){2-13}
 & \cR & \cR Acc     & \cR 64.9\se{1.1} & \cR \dt{$+$0.2} & \cR 49.4\se{0.3} & \cR \dt{$+$0.2} & \cR 91.0\se{0.9} & \cR \dt{$+$1.3} & \cR 62.9\se{2.4} & \cR \dt{$+$5.1} & \cR 68.8\se{1.0} & \cR \dt{$-$1.2} \\
 & \cR Mem.\ + RL & \cR Peak KV & \cR 1.45 & \cR \dx{0.56$\times$} & \cR 1.48 & \cR \dx{0.52$\times$} & \cR 0.68 & \cR \dx{0.77$\times$} & \cR 1.24 & \cR \dx{0.77$\times$} & \cR 1.12 & \cR \dx{0.59$\times$} \\
 & \cR & \cR AUC KV  & \cR 14.9 & \cR \dx{0.53$\times$} & \cR 16.4 & \cR \dx{0.50$\times$} & \cR 3.2 & \cR \dx{0.68$\times$} & \cR 9.2 & \cR \dx{0.78$\times$} & \cR 10.3 & \cR \dx{0.54$\times$} \\
\midrule
\multirow{9}{*}{\rotatebox{60}{Phi-4-r (14B)}}
 & \cB & \cB Acc     & \cB 71.7\se{1.0} & \cB & \cB 55.1\se{0.3} & \cB & \cB 87.3\se{1.1} & \cB & \cB 64.1\se{2.4} & \cB & \cB 64.1\se{1.0} & \cB \\
 & \cB Base & \cB Peak KV & \cB 2.65 & \cB & \cB 3.06 & \cB & \cB 1.43 & \cB & \cB 0.80 & \cB & \cB 2.45 & \cB \\
 & \cB & \cB AUC KV  & \cB 28.8 & \cB & \cB 35.9 & \cB & \cB 17.8 & \cB & \cB 3.7 & \cB & \cB 29.2 & \cB \\
\cmidrule(l){2-13}
 & \cS & \cS Acc     & \cS 69.8\se{1.0} & \cS & \cS 51.4\se{0.3} & \cS & \cS 90.6\se{0.9} & \cS & \cS 64.1\se{2.4} & \cS & \cS 65.0\se{1.0} & \cS \\
 & \cS Control & \cS Peak KV & \cS 3.04 & \cS & \cS 3.48 & \cS & \cS 1.04 & \cS & \cS 2.11 & \cS & \cS 2.64 & \cS \\
 & \cS & \cS AUC KV  & \cS 28.6 & \cS & \cS 36.7 & \cS & \cS 4.9 & \cS & \cS 14.2 & \cS & \cS 26.8 & \cS \\
\cmidrule(l){2-13}
 & \cM & \cM Acc     & \cM 67.6\se{1.1} & \cM \dt{$-$2.2} & \cM 48.7\se{0.3} & \cM \dt{$-$2.7} & \cM 89.7\se{1.0} & \cM \dt{$-$0.9} & \cM 61.6\se{2.4} & \cM \dt{$-$2.5} & \cM 61.8\se{1.1} & \cM \dt{$-$3.2} \\
 & \cM \memento{} & \cM Peak KV & \cM 1.17 & \cM \dx{0.38$\times$} & \cM 1.25 & \cM \dx{0.36$\times$} & \cM 0.51 & \cM \dx{0.49$\times$} & \cM 0.80 & \cM \dx{0.38$\times$} & \cM 0.92 & \cM \dx{0.35$\times$} \\
 & \cM & \cM AUC KV  & \cM 11.3 & \cM \dx{0.40$\times$} & \cM 13.1 & \cM \dx{0.36$\times$} & \cM 2.6 & \cM \dx{0.53$\times$} & \cM 6.2 & \cM \dx{0.44$\times$} & \cM 9.5 & \cM \dx{0.35$\times$} \\
\midrule
\multirow{9}{*}{\rotatebox{60}{Qwen3-32B}}
 & \cB & \cB Acc     & \cB 75.2\se{1.0} & \cB & \cB 62.7\se{0.3} & \cB & \cB 91.9\se{0.9} & \cB & \cB 65.9\se{2.4} & \cB & \cB 78.0\se{0.9} & \cB \\
 & \cB Base & \cB Peak KV & \cB 3.24 & \cB & \cB 3.67 & \cB & \cB 1.26 & \cB & \cB 1.89 & \cB & \cB 2.88 & \cB \\
 & \cB & \cB AUC KV  & \cB 26.7 & \cB & \cB 34.7 & \cB & \cB 5.5 & \cB & \cB 9.7 & \cB & \cB 22.9 & \cB \\
\cmidrule(l){2-13}
 & \cS & \cS Acc     & \cS 74.1\se{1.0} & \cS & \cS 58.5\se{0.3} & \cS & \cS 91.8\se{0.9} & \cS & \cS 64.6\se{2.4} & \cS & \cS 75.3\se{0.9} & \cS \\
 & \cS Control & \cS Peak KV & \cS 3.83 & \cS & \cS 4.51 & \cS & \cS 1.36 & \cS & \cS 2.45 & \cS & \cS 3.05 & \cS \\
 & \cS & \cS AUC KV  & \cS 35.2 & \cS & \cS 48.5 & \cS & \cS 6.2 & \cS & \cS 15.7 & \cS & \cS 27.7 & \cS \\
\cmidrule(l){2-13}
 & \cM & \cM Acc     & \cM 72.6\se{1.0} & \cM \dt{$-$1.5} & \cM 56.2\se{0.3} & \cM \dt{$-$2.3} & \cM 91.1\se{0.9} & \cM \dt{$-$0.7} & \cM 62.1\se{2.4} & \cM \dt{$-$2.5} & \cM 74.0\se{1.0} & \cM \dt{$-$1.3} \\
 & \cM \memento{} & \cM Peak KV & \cM 1.67 & \cM \dx{0.44$\times$} & \cM 1.74 & \cM \dx{0.39$\times$} & \cM 0.64 & \cM \dx{0.47$\times$} & \cM 1.07 & \cM \dx{0.44$\times$} & \cM 1.12 & \cM \dx{0.37$\times$} \\
 & \cM & \cM AUC KV  & \cM 14.0 & \cM \dx{0.40$\times$} & \cM 15.7 & \cM \dx{0.32$\times$} & \cM 2.8 & \cM \dx{0.45$\times$} & \cM 7.6 & \cM \dx{0.48$\times$} & \cM 9.3 & \cM \dx{0.34$\times$} \\
\midrule
\multirow{9}{*}{\rotatebox{60}{Olmo 3 (7B)}}
 & \cB & \cB Acc     & \cB 67.9\se{1.1} & \cB & \cB 52.7\se{0.3} & \cB & \cB 91.3\se{0.9} & \cB & \cB 50.8\se{2.5} & \cB & \cB 64.5\se{1.0} & \cB \\
 & \cB Base & \cB Peak KV & \cB 3.95 & \cB & \cB 4.21 & \cB & \cB 2.11 & \cB & \cB 3.21 & \cB & \cB 3.33 & \cB \\
 & \cB & \cB AUC KV  & \cB 50.8 & \cB & \cB 60.1 & \cB & \cB 10.7 & \cB & \cB 30.8 & \cB & \cB 40.0 & \cB \\
\cmidrule(l){2-13}
 & \cS & \cS Acc     & \cS 59.8\se{1.1} & \cS & \cS 48.3\se{0.3} & \cS & \cS 90.4\se{0.9} & \cS & \cS 45.7\se{2.5} & \cS & \cS 58.8\se{1.1} & \cS \\
 & \cS Control & \cS Peak KV & \cS 3.51 & \cS & \cS 3.78 & \cS & \cS 2.00 & \cS & \cS 2.94 & \cS & \cS 3.22 & \cS \\
 & \cS & \cS AUC KV  & \cS 37.1 & \cS & \cS 46.0 & \cS & \cS 9.1 & \cS & \cS 23.8 & \cS & \cS 37.6 & \cS \\
\cmidrule(l){2-13}
 & \cM & \cM Acc     & \cM 55.4\se{3.2} & \cM \dt{$-$4.4} & \cM 48.1\se{0.9} & \cM \dt{$-$0.2} & \cM 91.1\se{0.9} & \cM \dt{$+$0.7} & \cM 49.5\se{2.5} & \cM \dt{$+$3.8} & \cM 56.0\se{1.1} & \cM \dt{$-$2.8} \\
 & \cM \memento{} & \cM Peak KV & \cM 3.21 & \cM \dx{0.91$\times$} & \cM 3.43 & \cM \dx{0.91$\times$} & \cM 1.70 & \cM \dx{0.85$\times$} & \cM 2.72 & \cM \dx{0.93$\times$} & \cM 2.21 & \cM \dx{0.69$\times$} \\
 & \cM & \cM AUC KV  & \cM 37.8 & \cM \dx{1.02$\times$} & \cM 43.6 & \cM \dx{0.95$\times$} & \cM 8.5 & \cM \dx{0.93$\times$} & \cM 25.2 & \cM \dx{1.06$\times$} & \cM 20.6 & \cM \dx{0.55$\times$} \\
\bottomrule
\end{tabular}
\end{table}

\paragraph{Results and Evaluation.}
\label{sec:results}
We evaluate \memento{} across four model families and scales: Qwen3-8B,
Phi-4-reasoning (14B), Qwen3-32B, and Olmo-3-7B. 
Example \memento{} traces from Qwen3-32B are provided in \Cref{app:example}.
All results report pass@1 accuracy. We evaluate on 14 benchmarks spanning competition
mathematics (11 contests sourced from MathArena, grouped as ``Comp.\ Math'' in \Cref{tab:main-sft}), standard math (MATH-500), science (GPQA Diamond), and code
(LiveCodeBench~v6); \Cref{tab:main-sft} summarizes accuracy alongside KV cache footprints.

\paragraph{Control Runs.} To decouple the effect of performing SFT on already strong reasoning models
(that leads to some performance loss) we do control (shown in Gray in
\Cref{tab:main-sft}) runs where we train the base models on the original
unmodified OpenThoughts subsets.  As one may expect the highest performance loss,
both for \memento{} and the control runs, happens for the most challenging
Competition math benchmarks while for easier benchmarks such as MATH-500
\memento{} is able to match baselines almost perfectly. 

\paragraph{Scale Helps.}
Within the Qwen3 family the accuracy gap shrinks with scale, from $-$6.3\,pp at
8B to $-$3.5\,pp at 32B (averaged across the five benchmark groups in \Cref{tab:main-sft}). This suggests that larger models manage compressed
context more effectively and that further gains may be achievable at greater
scale.

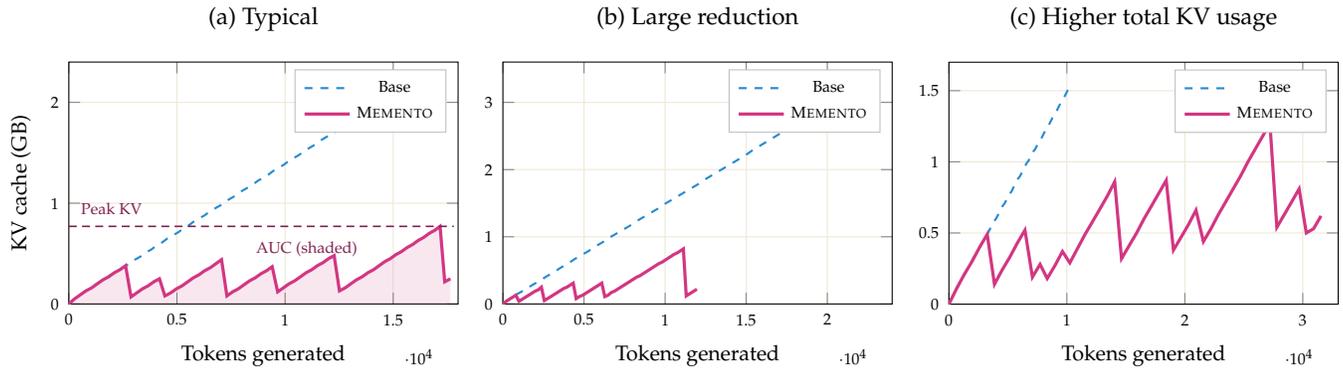
\begin{figure}[ht!]
\centering
\begin{tikzpicture}
  \begin{axis}[
    width=0.38\textwidth, height=4.8cm,
    xlabel={Tokens generated},
    ylabel={KV cache (GB)},
    xlabel style={font=\scriptsize},
    ylabel style={font=\scriptsize},
    xmin=0, xmax=18000,
    ymin=0, ymax=2.4,
    grid=major,
    grid style={cGrid, thin},
    tick label style={font=\tiny},
    title={\footnotesize (a) Typical},
    title style={at={(0.5,1.02)}},
    legend style={
      at={(0.97,0.97)}, anchor=north east,
      font=\tiny,
      draw=gray!50,
    },
  ]
    \addplot[fill=cMemento, fill opacity=0.12, draw=none, forget plot] coordinates {
      (0,0.00) (261,0.05) (526,0.09) (791,0.13) (1054,0.16) (1317,0.20)
      (1580,0.24) (1843,0.27) (2104,0.31) (2365,0.34) (2625,0.38) (2875,0.07)
      (3140,0.11) (3404,0.15) (3667,0.18) (3929,0.22) (4191,0.25) (4443,0.08)
      (4708,0.11) (4973,0.15) (5236,0.18) (5499,0.22) (5761,0.26) (6022,0.29)
      (6283,0.33) (6544,0.36) (6803,0.40) (7062,0.44) (7310,0.08) (7571,0.12)
      (7835,0.15) (8099,0.19) (8360,0.23) (8621,0.26) (8884,0.30) (9144,0.33)
      (9404,0.37) (9658,0.12) (9921,0.16) (10184,0.20) (10446,0.23) (10711,0.27)
      (10971,0.31) (11233,0.34) (11493,0.38) (11753,0.41) (12011,0.45) (12270,0.48)
      (12515,0.13) (12779,0.16) (13040,0.20) (13304,0.24) (13567,0.27) (13828,0.31)
      (14089,0.34) (14349,0.38) (14610,0.42) (14869,0.45) (15127,0.49) (15386,0.52)
      (15643,0.56) (15900,0.59) (16154,0.63) (16408,0.66) (16661,0.70) (16915,0.73)
      (17167,0.77) (17383,0.22) (17646,0.25)
    } \closedcycle;
    \addplot[cVanilla, thick, dashed] coordinates {
      (0,0.00) (257,0.05) (515,0.09) (772,0.13) (1028,0.16) (1284,0.20)
      (1539,0.23) (1794,0.27) (2048,0.30) (2301,0.34) (2554,0.37) (2807,0.40)
      (3059,0.44) (3310,0.47) (3560,0.51) (3812,0.54) (4061,0.58) (4310,0.61)
      (4558,0.65) (4805,0.68) (5052,0.71) (5299,0.75) (5546,0.78) (5792,0.82)
      (6038,0.85) (6284,0.88) (6529,0.92) (6774,0.95) (7019,0.98) (7264,1.02)
      (7508,1.05) (7753,1.08) (7996,1.12) (8240,1.15) (8482,1.18) (8724,1.22)
      (8967,1.25) (9209,1.28) (9451,1.32) (9693,1.35) (9934,1.38) (10176,1.42)
      (10418,1.45) (10659,1.48) (10899,1.52) (11139,1.55) (11379,1.58) (11620,1.62)
      (11859,1.65) (12099,1.68) (12338,1.71) (12576,1.75) (12815,1.78) (13053,1.81)
      (13291,1.85) (13528,1.88) (13766,1.91) (14003,1.94) (14240,1.98) (14477,2.01)
      (14715,2.04) (14951,2.07) (15187,2.11) (15423,2.14) (15659,2.17)
    };
    \addlegendentry{Base}
    \addplot[cMemento, very thick] coordinates {
      (0,0.00) (261,0.05) (526,0.09) (791,0.13) (1054,0.16) (1317,0.20)
      (1580,0.24) (1843,0.27) (2104,0.31) (2365,0.34) (2625,0.38) (2875,0.07)
      (3140,0.11) (3404,0.15) (3667,0.18) (3929,0.22) (4191,0.25) (4443,0.08)
      (4708,0.11) (4973,0.15) (5236,0.18) (5499,0.22) (5761,0.26) (6022,0.29)
      (6283,0.33) (6544,0.36) (6803,0.40) (7062,0.44) (7310,0.08) (7571,0.12)
      (7835,0.15) (8099,0.19) (8360,0.23) (8621,0.26) (8884,0.30) (9144,0.33)
      (9404,0.37) (9658,0.12) (9921,0.16) (10184,0.20) (10446,0.23) (10711,0.27)
      (10971,0.31) (11233,0.34) (11493,0.38) (11753,0.41) (12011,0.45) (12270,0.48)
      (12515,0.13) (12779,0.16) (13040,0.20) (13304,0.24) (13567,0.27) (13828,0.31)
      (14089,0.34) (14349,0.38) (14610,0.42) (14869,0.45) (15127,0.49) (15386,0.52)
      (15643,0.56) (15900,0.59) (16154,0.63) (16408,0.66) (16661,0.70) (16915,0.73)
      (17167,0.77) (17383,0.22) (17646,0.25)
    };
    \addlegendentry{\memento{}}
    \draw[cMemento!60!black, densely dashed, semithick] (axis cs:0,0.77) -- (axis cs:17800,0.77);
    \node[font=\tiny, text=cMemento!60!black, anchor=south west] at (axis cs:100,0.79) {Peak KV};
    \node[font=\tiny, text=cMemento!60!black] at (axis cs:11000,0.55) {AUC (shaded)};
  \end{axis}
\end{tikzpicture}\hfill
\begin{tikzpicture}
  \begin{axis}[
    width=0.38\textwidth, height=4.8cm,
    xlabel={Tokens generated},
    ylabel={},
    xlabel style={font=\scriptsize},
    xmin=0, xmax=24000,
    ymin=0, ymax=3.6,
    grid=major,
    grid style={cGrid, thin},
    tick label style={font=\tiny},
    title={\footnotesize (b) Large reduction},
    title style={at={(0.5,1.02)}},
    legend style={
      at={(0.97,0.97)}, anchor=north east,
      font=\tiny,
      draw=gray!50,
    },
  ]
    \addplot[cVanilla, thick, dashed] coordinates {
      (0,0.00) (450,0.08) (900,0.15) (1349,0.21) (1797,0.28) (2241,0.34)
      (2685,0.41) (3124,0.47) (3563,0.54) (4000,0.60) (4435,0.67) (4866,0.73)
      (5296,0.79) (5726,0.86) (6156,0.92) (6585,0.99) (7013,1.05) (7441,1.11)
      (7868,1.17) (8292,1.24) (8717,1.30) (9141,1.36) (9564,1.42) (9986,1.49)
      (10407,1.55) (10828,1.61) (11248,1.67) (11666,1.73) (12085,1.79) (12502,1.86)
      (12919,1.92) (13335,1.98) (13751,2.04) (14166,2.10) (14580,2.16) (14993,2.22)
      (15404,2.29) (15816,2.35) (16227,2.41) (16638,2.47) (17048,2.53) (17456,2.59)
      (17863,2.65) (18271,2.71) (18679,2.77) (19086,2.83) (19491,2.89) (19896,2.95)
      (20300,3.01) (20704,3.07) (21107,3.12) (21509,3.18) (21911,3.25) (22311,3.30)
      (22711,3.36) (23053,3.41)
    };
    \addlegendentry{Base}
    \addplot[cMemento, very thick] coordinates {
      (0,0.00) (199,0.04) (397,0.07) (596,0.10) (795,0.13) (989,0.04)
      (1188,0.07) (1386,0.10) (1585,0.13) (1782,0.16) (1979,0.19) (2175,0.21)
      (2372,0.25) (2554,0.05) (2753,0.08) (2952,0.11) (3151,0.14) (3350,0.17)
      (3547,0.20) (3744,0.23) (3941,0.25) (4138,0.28) (4334,0.31) (4517,0.08)
      (4715,0.11) (4913,0.13) (5112,0.16) (5309,0.19) (5505,0.22) (5702,0.25)
      (5898,0.28) (6095,0.31) (6281,0.11) (6480,0.13) (6678,0.17) (6875,0.19)
      (7072,0.22) (7268,0.25) (7466,0.28) (7662,0.31) (7858,0.34) (8053,0.37)
      (8249,0.40) (8444,0.42) (8639,0.45) (8833,0.48) (9027,0.51) (9221,0.54)
      (9414,0.57) (9607,0.60) (9799,0.62) (9989,0.65) (10180,0.68) (10369,0.71)
      (10558,0.74) (10748,0.76) (10937,0.79) (11126,0.82) (11303,0.12) (11499,0.15)
      (11697,0.18) (11894,0.21) (11959,0.22)
    };
    \addlegendentry{\memento{}}
  \end{axis}
\end{tikzpicture}\hfill
\begin{tikzpicture}
  \begin{axis}[
    width=0.38\textwidth, height=4.8cm,
    xlabel={Tokens generated},
    ylabel={},
    xlabel style={font=\scriptsize},
    xmin=0, xmax=33000,
    ymin=0, ymax=1.7,
    grid=major,
    grid style={cGrid, thin},
    tick label style={font=\tiny},
    title={\footnotesize (c) Higher total KV usage},
    title style={at={(0.5,1.02)}},
    legend style={
      at={(0.97,0.97)}, anchor=north east,
      font=\tiny,
      draw=gray!50,
    },
  ]
    \addplot[cVanilla, thick, dashed] coordinates {
      (0,0.00) (192,0.04) (386,0.07) (578,0.10) (772,0.13) (964,0.16)
      (1156,0.19) (1347,0.21) (1538,0.24) (1729,0.27) (1920,0.30) (2110,0.33)
      (2299,0.35) (2488,0.38) (2677,0.41) (2866,0.44) (3055,0.46) (3243,0.49)
      (3430,0.52) (3618,0.55) (3806,0.58) (3993,0.60) (4180,0.63) (4366,0.66)
      (4552,0.69) (4737,0.71) (4921,0.74) (5106,0.77) (5291,0.79) (5475,0.82)
      (5661,0.85) (5846,0.88) (6031,0.90) (6215,0.93) (6400,0.96) (6584,0.99)
      (6768,1.01) (6952,1.04) (7135,1.07) (7318,1.09) (7502,1.12) (7685,1.15)
      (7868,1.17) (8050,1.20) (8233,1.23) (8415,1.25) (8598,1.28) (8780,1.31)
      (8963,1.33) (9145,1.36) (9327,1.39) (9508,1.42) (9690,1.44) (9870,1.47)
      (10052,1.50) (10233,1.52) (10415,1.55)
    };
    \addlegendentry{Base}
    \addplot[cMemento, very thick] coordinates {
      (0,0.00) (657,0.11) (1308,0.21) (1960,0.30) (2603,0.40) (3246,0.49)
      (3862,0.14) (4512,0.24) (5160,0.33) (5805,0.43) (6445,0.52) (7078,0.19)
      (7728,0.28) (8333,0.18) (8980,0.27) (9627,0.37) (10251,0.29) (10894,0.39)
      (11538,0.49) (12176,0.58) (12808,0.67) (13431,0.76) (14054,0.86) (14634,0.32)
      (15271,0.41) (15913,0.50) (16543,0.60) (17174,0.69) (17798,0.78) (18422,0.87)
      (19036,0.38) (19680,0.47) (20315,0.56) (20946,0.66) (21575,0.44) (22215,0.53)
      (22847,0.63) (23476,0.72) (24097,0.81) (24725,0.90) (25351,1.00) (25973,1.09)
      (26596,1.18) (27213,1.27) (27803,0.54) (28436,0.63) (29066,0.72) (29689,0.81)
      (30298,0.50) (30915,0.53) (31546,0.62)
    };
    \addlegendentry{\memento{}}
  \end{axis}
\end{tikzpicture}
\caption{\footnotesize\textbf{KV cache traces on individual problems (Qwen3-8B, both answers correct).}
\textbf{(a)}~AIME24 P2: typical sawtooth pattern with 6 compactions; peak 0.77 vs 2.17\,GB ($2.8\times$ reduction).
\textbf{(b)}~AIME24 P26: \memento{} solves the problem in 12k tokens (vs 23k);
frequent compactions keep peak at 0.82 vs 3.41\,GB ($4.2\times$).
\textbf{(c)}~AIME24 P5: \memento{} generates $3\times$ more tokens (31k vs 10k) with many compactions.
Peak is still lower (1.27 vs 1.55\,GB), but the total KV area-under-curve is $2.1\times$ \emph{higher} than the base---a failure mode where block masking induces excessive generation.}
\label{fig:kv-trace}
\label{fig:kv-trace-examples}
\end{figure}

\paragraph{Peak KV cache and AUC savings.}  We observe that peak KV is reduced by
2--3$\times$ and KV AUC (area under the KV-cache-size curve over generation
steps) capturing total memory-time cost,  by 2--3.5$\times$ on competition math,
with even larger reductions on benchmarks where the base model generates long
responses. \Cref{fig:kv-trace} illustrates the range of per-problem KV cache
behaviors produced by block masking.

\paragraph{Memento on Olmo-3-7B-Think.}
\label{sec:olmo3-discussion}
We applied our
\openmementos{} dataset and training recipe to Olmo-3-7B-Think, which uses a
hybrid attention architecture: 24 of its 32 layers employ sliding-window
attention (window size 4096), while only 8 layers use full causal attention.
Additionally, Olmo 3 uses multi-head attention (MHA, 32 KV heads) rather than
grouped query attention (GQA, 8 KV heads in Qwen3).
\memento{} transferred with no architecture-specific modifications: accuracy is
well preserved, with Comp.\ Math dropping only 0.2\,pp relative to Control and
MATH-500 \emph{improving} by 0.7\,pp (\Cref{tab:main-sft}).
However, KV cache savings are substantially more modest than for the other model
families (${\sim}0.85$--$0.93\times$ peak vs.\ $0.35$--$0.47\times$).
This is because sliding-window layers already cap their KV cache at 4096 tokens
regardless of block masking---so 75\% of layers gain nothing from eviction.
Only the 8 full-attention layers benefit, limiting the overall reduction.
On some benchmarks, the AUC metric is even slightly \emph{worse} for
\memento{} (e.g., AIME'26: $1.02\times$), because
summaries lengthen the response, increasing total memory-time cost despite a
lower peak. LCB shows the largest savings ($0.69\times$ peak, $0.55\times$ AUC),
likely because code problems have shorter responses where fewer tokens exceed
the sliding window.

\begin{tcolorbox}[finding]
\textbf{Models learn to summarize their own reasoning.} 
After SFT on \openmementos{}, models internalize the block-and-summarize process
as a new capability: they produce self-contained mementos that reduce peak KV
cache by $2$--$3\times$ on models with uniform attention while maintaining strong accuracy across 
benchmarks. On MATH-500 the gap is under 1\,pp; on the hardest competition math
benchmarks Qwen3-32B stays within 2.6\,pp on AIME'26. The gap shrinks with scale
($-$6.3\,pp at 8B $\to$ $-$3.5\,pp at 32B), suggesting \memento{} becomes
increasingly effective at larger model sizes. \memento{} also transfers to
Olmo-3-7B's hybrid sliding-window architecture with minimal accuracy loss,
though KV savings are inherently limited by the sliding window's bounded cache.
\end{tcolorbox}

\paragraph{Compression behavior.}
\label{sec:compression}
\label{sec:compression-dynamics}
\label{sec:block-summary-distributions}
\label{sec:inference-compression}

\begin{figure}[H]
    \centering
    \includegraphics[width=\textwidth]{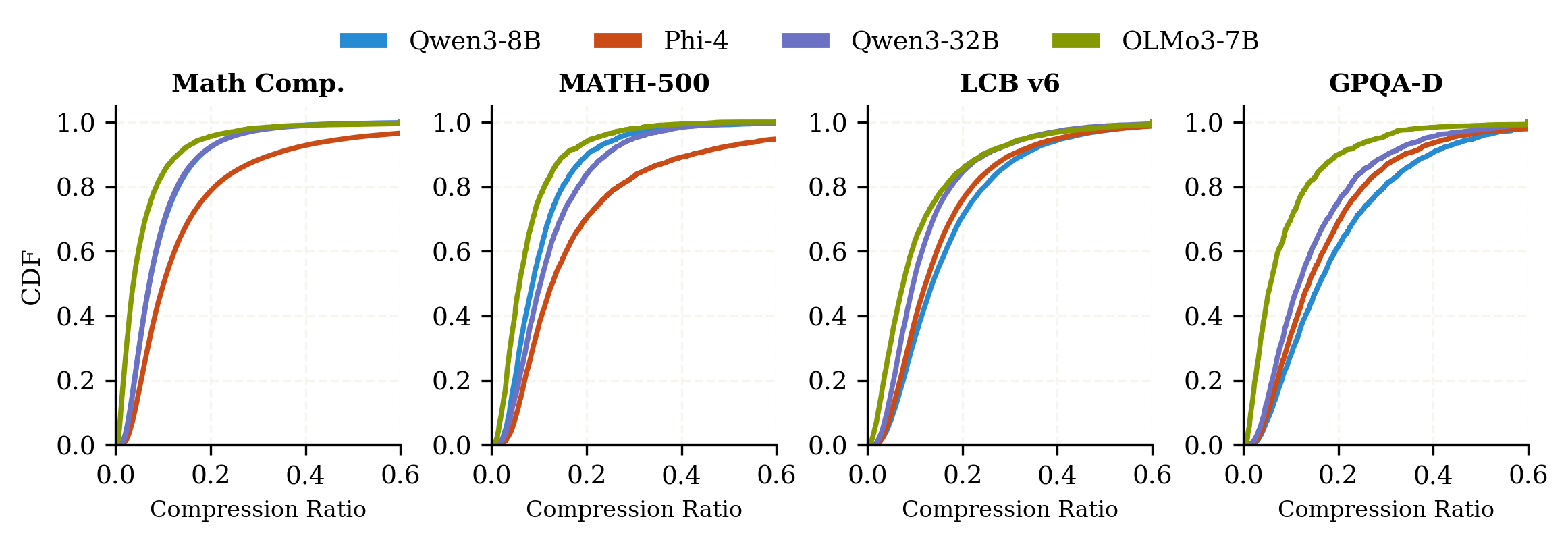}
    \caption{\textbf{CDF of compression ratio} (summary chars / block chars). OLMo3-7B and Qwen3-8B achieve the tightest compression on competition math. Phi-4 has the widest compression spread, especially on MATH-500.}
    \label{fig:compression-cdf}
\end{figure}

How does compression vary across model families and benchmarks? Summary sizes are remarkably stable (median 260--615 chars), matching the training distribution (\Cref{fig:training-data-boxplots}), while block sizes vary widely across models and tasks. This confirms the model learns a consistent compression skill that generalizes to harder problems. \Cref{fig:compression-cdf} shows the full CDF of compression ratios across all four \memento{} models, revealing that the bulk of blocks achieve 5--20$\times$ compression with a thin tail of low-compression outliers.

\begin{tcolorbox}[finding]
\textbf{Summary length is stable; compression scales with difficulty.} Across four model families and four benchmarks, mementos converge to $\sim$260--615 characters regardless of block length, matching training targets. Compression is strongest on competition math (9--27$\times$) and weakest on shorter-block benchmarks (6--9$\times$), confirming the model learns a stable summary skill, not a fixed ratio.
\end{tcolorbox}

\section{Improving Accuracy via RL}
\label{sec:improving-accuracy}

\paragraph{Capability under compression.}\label{sec:ensemble}
We first investigate whether we can match the baseline performance with \memento{}
or there is some inherent limitation due to compression.
We focus on math, which is the most challenging domain for compression (\Cref{tab:main-sft}).
Generating $n{=}64$ independent completions per problem for all three
model families on AIME 2024/25/26, we find that coverage (pass@64) is nearly
identical: the gap averages only 2.6\,pp and the Jaccard similarity between Base
and \memento{} solved sets averages 96.4\%, reaching 100\% in two of nine
settings (\Cref{tab:coverage}).

\begin{wraptable}{r}{0.55\textwidth}
\vspace{-12pt}
\centering
\caption{\textbf{Problem coverage} (pass@64) and solved-set overlap for Base
vs.\ \memento{} on AIME ($n{=}64$ per problem, 30 problems per benchmark).}
\label{tab:coverage}
\scriptsize
\begin{tabular}{llcccc}
\toprule
\textbf{Model} & \textbf{Bench.} & \textbf{Base} & \textbf{\memento{}} & \textbf{Ret.} & \textbf{Jacc.} \\
\midrule
\multirow{3}{*}{Qwen3-8B}
 & AIME'24 & 93.3 & 90.0 & 96.4 & 96.4 \\
 & AIME'25 & 93.3 & 86.7 & 92.9 & 92.9 \\
 & AIME'26 & 86.7 & 90.0 & 100.0 & 96.3 \\
\midrule
\multirow{3}{*}{\shortstack[l]{Phi-4-r\\(14B)}}
 & AIME'24 & 93.3 & 93.3 & 100.0 & 100.0 \\
 & AIME'25 & 93.3 & 90.0 & 96.4 & 96.4 \\
 & AIME'26 & 93.3 & 90.0 & 96.4 & 96.4 \\
\midrule
\multirow{3}{*}{Qwen3-32B}
 & AIME'24 & 93.3 & 93.3 & 100.0 & 100.0 \\
 & AIME'25 & 90.0 & 83.3 & 92.6 & 92.6 \\
 & AIME'26 & 93.3 & 90.0 & 96.4 & 96.4 \\
\bottomrule
\end{tabular}
\vspace{-8pt}
\end{wraptable}

\begin{figure}[H]
    \centering
    \includegraphics[width=\linewidth]{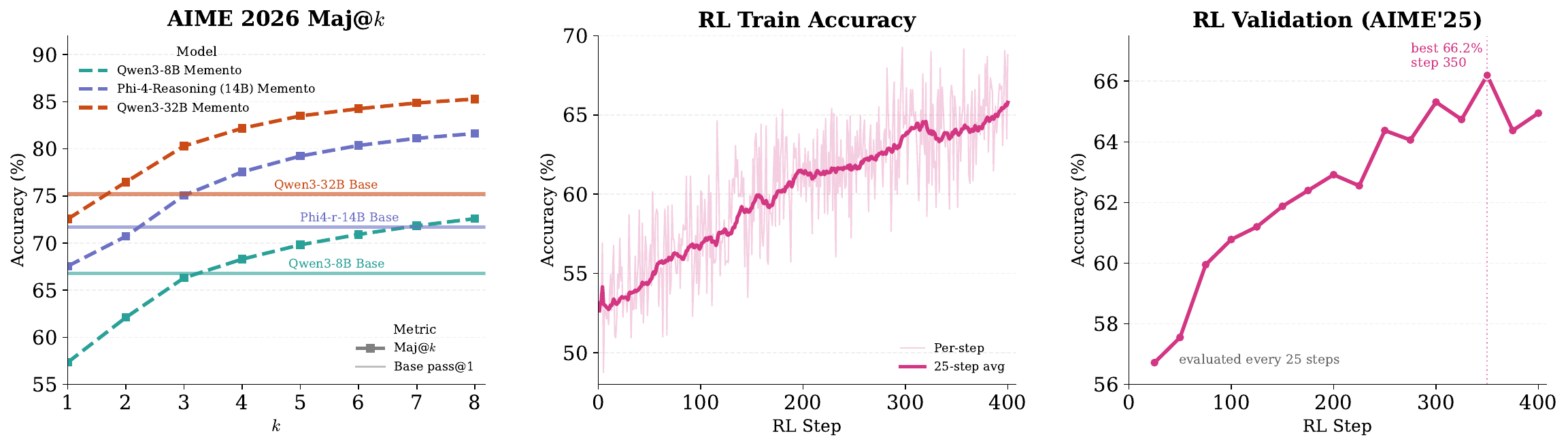}
    \caption{\footnotesize\textbf{Majority-vote headroom and Qwen3-8B CISPO~\citep{minimax_m1} RL trajectory.}
    Left: AIME 2026 maj@$k$ for the three \memento{} SFT models (\Cref{tab:main-sft}); horizontal
    lines show each Base model's pass@1. All models match Base accuracy by $k{=}2$--$3$. Middle: per-step RL training accuracy
    (faint raw trace with 25-step moving average).
    Right: AIME'25 validation accuracy evaluated every 25 steps,
    peaking at 66.2\% at step~350 (used in \Cref{tab:main-sft}).
    Majority voting uses uniform tie-breaking among tied majority answers.}
    \label{fig:pass_maj_at_k}
\end{figure}

\paragraph{Majority voting recovers the gap.}
The coverage analysis above shows that \memento{} models \emph{can} solve nearly the same problems as their base counterparts---they just do so less consistently.
Majority voting (maj@$k$) makes this concrete: as shown in the left panel of \Cref{fig:pass_maj_at_k}, all three \memento{} SFT models match or exceed the Base pass@1 accuracy with just $k{=}2$--$3$ samples.
For Qwen3-32B on AIME'26, maj@2 already surpasses the Base pass@1 line.
This tells us two things: (1)~the accuracy gap after SFT is a \emph{consistency} problem, not a \emph{capability} problem---the correct answers are in the distribution, they are just not the mode; and (2)~RL is a natural fix, since it can sharpen the distribution toward correct traces without needing to teach new skills.

\paragraph{Recovering accuracy via RL.}\label{sec:rl}
Given that the correct answers are already present in the \memento{} distribution, RL should improve pass@1 by reallocating probability mass toward correct compressed traces rather than by teaching entirely new skills. We fine-tune the Qwen3-8B
\memento{} SFT checkpoint with CISPO~\citep{minimax_m1}, Clipped
Importance-Sampled Policy Optimization, a GRPO~\citep{grpo} variant that clips
and detaches the importance-sampling weight.  Similarly to \citet{minimax_m1},
we found CISPO to be more stable than standard GRPO during training. We also add
a KL penalty ($\beta{=}0.001$) to prevent the response-length collapse we
observed in initial runs without regularization, and adopt rule-based math
rewards. 

Rollouts use memento attention block masking
via our custom vLLM engine (\Cref{sec:inference}). Training uses
sparse block-masked attention (similar to Stage 2 of SFT) to match the inference-time masking pattern. 
Full hyperparameters and training details are provided in \Cref{appendix:rl-details}.

\paragraph{CISPO algorithm.}
We use CISPO~\citep{minimax_m1} (Clipped Importance-Sampled Policy Optimization), a GRPO~\citep{grpo} variant that replaces the PPO~\citep{ppo} clipped surrogate objective with a stop-gradient clipped importance-sampling weight:
\begin{equation}
  L = -\operatorname{sg}\!\bigl(\operatorname{clip}(r_t(\theta),\, 1{-}\epsilon_\text{low},\, 1{+}\epsilon_\text{high})\bigr) \cdot A_t \cdot \log \pi_\theta(a_t \mid s_t),
\end{equation}
where $r_t(\theta) = \pi_\theta / \pi_{\theta_\text{old}}$ is the importance ratio and $\operatorname{sg}$ denotes stop-gradient. Unlike PPO clipping, which zeros out gradients when the ratio exceeds the trust region, CISPO ensures every token contributes a gradient signal---the clipped ratio acts as a fixed per-token weight. We add a KL penalty term $\beta \cdot D_\text{KL}(\pi_\theta \| \pi_\text{ref})$ with $\beta = 0.001$ to prevent excessive drift from the SFT checkpoint.

\paragraph{Block length capping.}
Because we use accuracy as the sole reward signal, we observed the model learning to generate fewer and longer reasoning blocks, undermining the KV cache savings that block masking provides. To maintain low peak KV cache occupancy during RL rollouts, we cap individual blocks at 7K tokens: when a block exceeds this limit during generation, the vLLM engine forces a \texttt{<|block\_end|>} token and the model continues from a new block.

The middle and right panels of \Cref{fig:pass_maj_at_k} show the training and
validation trajectories. Train accuracy rises from 52.7\% to 65.8\% (25-step
moving average) over 400 steps, while AIME'25 validation peaks at 66.2\% at
step~350. After RL, \memento{}+RL raises AIME'26 from 57.3
to 64.9 and Comp.\ Math from 45.1 to 49.4, while also improving GPQA-D from
55.8 to 62.9 above the 61.4 vanilla baseline.  The compression remains
substantial: peak KV rises from 1.08 to 1.48\,GB after RL, still well below the
2.71\,GB vanilla footprint. RL therefore converts the majority-voting headroom
into stronger single-sample accuracy while preserving much of \memento{}'s
memory advantage.

\begin{tcolorbox}[finding]
\paragraph{Matching the baselines with RL.}
The pass@1 drop after SFT on \openmementos{} reflects reduced consistency, not
lost knowledge.  Without any additional training, majority voting at $k{=}3$
already recovers base-model accuracy (Figure~\ref{fig:pass_maj_at_k}, left).
With RL fine-tuning we can significantly improve pass@1 accuracy and match or
improve over the control run for Qwen3-8B (\Cref{tab:main-sft}).
\end{tcolorbox}

\section{Inference and the Implicit KV Channel}
\label{sec:inference}
\label{sec:throughput}

\subsection{Serving Memento Models with vLLM}
\label{sec:vllm-serving}

\memento{}'s block masking requires \emph{non-standard, data-dependent sparse
attention}: which tokens are masked depends on the generated sequence itself,
not on a fixed pattern known at compile time.  To the best of our knowledge, no
production inference framework, including vLLM~\citep{kwon2023vllm},
SGLang~\citep{sglang2024}, or TensorRT-LLM, provides a built-in mechanism for
request-level custom sparse attention masks that evolve during generation. We
therefore build native block masking support directly into vLLM's V1 engine,
extending it so that it \emph{physically} removes masked tokens from the KV
cache. Our approach operates purely at the Python level of vLLM, can be
installed as a simple patch on top of an existing vLLM installation, and works
with the vanilla FlashAttention and FlashInfer kernels, requiring no custom
sparse attention kernel.  For more details on the implementation, see
\Cref{appendix:vllm-implementation}.

\begin{figure}[ht!]
\centering
\begin{tikzpicture}
  \begin{groupplot}[
    group style={group size=2 by 1, horizontal sep=2.0cm},
    width=0.48\textwidth, height=5.5cm,
    xlabel={Concurrency},
    xlabel style={font=\small},
    xmin=0, xmax=260,
    xtick={16,64,128,240},
    tick label style={font=\footnotesize},
    ymajorgrids=true,
    grid style={cGrid, thin},
    legend style={font=\footnotesize, draw=gray!50},
  ]
    \nextgroupplot[
      ylabel={Aggregate tok/s}, ylabel style={font=\small},
      ymin=0, ymax=5000,
      title={\small Token Throughput}, title style={at={(0.5,1.02)}},
      legend pos=north west,
      legend to name=legtpfull,
      legend columns=2,
    ]
    \addplot[thick, cVanilla, mark=square*, mark size=2pt] coordinates {
      (16, 1295) (32, 1729) (64, 2151) (96, 2386) (128, 2374) (240, 2447)
    }; \addlegendentry{Vanilla}
    \addplot[thick, cMemento, mark=triangle*, mark size=2.5pt] coordinates {
      (16, 1503) (32, 2153) (64, 2992) (96, 3419) (128, 3630) (240, 4290)
    }; \addlegendentry{\memento{}}
    \node[font=\tiny, text=cMemento!80!black, anchor=south] at (axis cs:240, 4350) {$1.75\times$};
    \node[font=\tiny, text=gray, anchor=south west] at (axis cs:100, 2420) {\textit{KV full}};

    \nextgroupplot[
      ylabel={Wall-clock time (s)}, ylabel style={font=\small},
      ymin=0, ymax=2300,
      title={\small Batch Completion Time}, title style={at={(0.5,1.02)}},
      legend pos=north east,
    ]
    \addplot[thick, cVanilla, mark=square*, mark size=2pt] coordinates {
      (16, 2084) (32, 1514) (64, 1274) (96, 1138) (128, 1157) (240, 1096)
    }; \addlegendentry{Vanilla}
    \addplot[thick, cMemento, mark=triangle*, mark size=2.5pt] coordinates {
      (16, 1954) (32, 1332) (64, 982) (96, 872) (128, 828) (240, 693)
    }; \addlegendentry{\memento{}}
    \node[font=\tiny, text=cMemento!80!black, anchor=north] at (axis cs:240, 640) {$1.58\times$ faster};
  \end{groupplot}
  \node at ($(group c1r1.south)!0.5!(group c2r1.south) + (0,-1.0)$) {\ref{legtpfull}};
\end{tikzpicture}
\caption{\textbf{Serving throughput (Qwen3-8B, 1$\times$ B200 GPU).} AIME24 $\times$ 8 repetitions (240 requests, 32K max tokens). Left: \memento{} sustains $1.75\times$ higher token throughput at full concurrency. Right: $1.58\times$ faster batch completion. Vanilla plateaus as KV cache fills GPU memory.}
\label{fig:throughput-full}
\label{fig:throughput}
\end{figure}
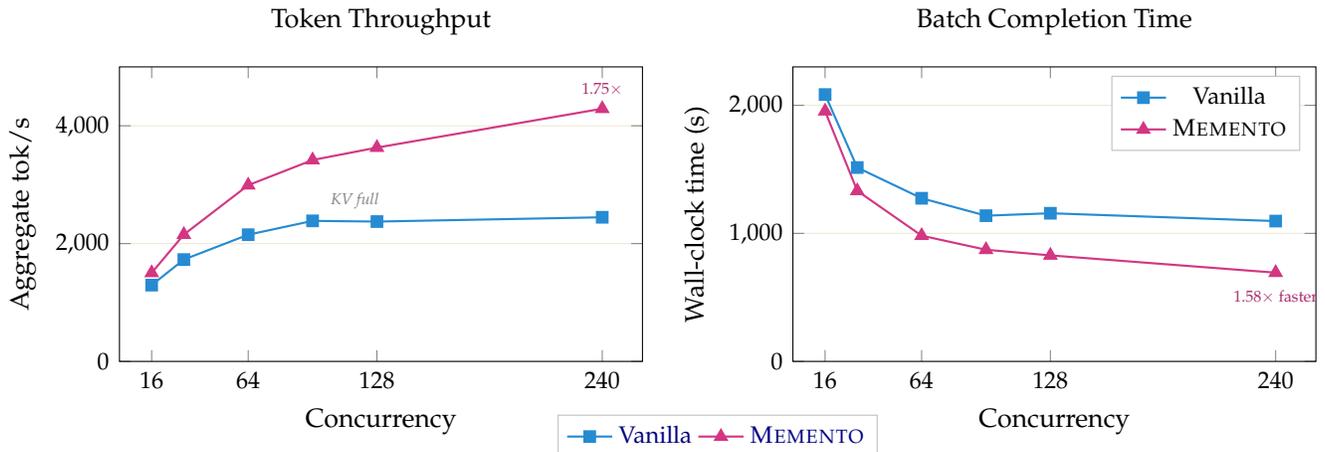

\paragraph{Throughput experiments.}
 We benchmark serving throughput on Qwen3-8B with AIME24 $\times$ 8
repetitions (240 concurrent requests) with 32K max tokens on a single B200 GPU.
At high concurrency, vanilla vLLM becomes KV-cache-bound: throughput plateaus as
the KV cache fills GPU memory. \memento{}'s block masking frees KV cache entries
as blocks complete, allowing the engine to sustain higher batch sizes and
throughput throughout the run.
\memento{} sustains 4{,}290 tok/s vs.\ 2{,}447 for vanilla
($1.75\times$) and completes the batch in 693s vs.\ 1{,}096s ($1.58\times$
faster).
This infrastructure was also crucial for enabling RL fine-tuning with
\memento{}: generating 32K-token training rollouts requires an inference engine
that natively supports block masking during generation, since each rollout must
produce and compact blocks on the fly.  Without the vLLM integration, generating
these long traces at the scale required for RL would be infeasible.

\begin{tcolorbox}[finding]
\textbf{Memento + vLLM.}
\memento{}'s vLLM integration physically removes masked KV entries, sustaining
$1.75\times$ higher throughput at full concurrency on a single B200 GPU.  Our vLLM
implementation enabled us to perform reasoning RL by supporting on-the-fly block masking
during 32K-token rollout generation.
\end{tcolorbox}

\subsection{The Dual Information Stream}
\label{appendix:kv-stream}

\subsubsection{KV Cache Ablation: Do Memento KV States Carry Block Information?}
\label{sec:kv-ablation}

Under memento attention, block content is masked for \emph{future} tokens, but
the memento's KV values were computed \emph{during generation} while the model
could still attend to the full block. Do these KV states carry useful
information beyond the memento text itself?
We denote thinking block~$i$ as $T_i$ and its corresponding memento as $M_i$.

\paragraph{Experiment.} We compare two inference modes on the \emph{same}
Qwen3-8B memento attention checkpoint: 

\begin{itemize}[leftmargin=*]
\item \textbf{Memento attention} (normal): While generating $M_i$, the model
attends to all tokens in $T_i$ as well as the prompt and all preceding
mementos. Once $M_i$ is complete, $T_i$ is masked from all subsequent
attention---but $M_i$'s KV cache entries, which were computed with block
context, are retained. Future tokens therefore attend to memento KV states that
implicitly encode block content.
\item \textbf{Memento attention + restart}:
Generation of each memento proceeds in two steps.  \emph{Step~1 (generation):}
$M_i$'s text is generated identically to normal memento attention: the
model attends to $T_i$ and produces the same summary tokens.  \emph{Step~2
(KV recomputation):} After $M_i$ is complete, we discard the KV cache and
run a fresh prefill pass over the effective context: prompt $+$ $M_1$ $+$
$M_2$ $+$ $\cdots$ $+$ $M_i$ (with standard causal masking within
and across mementos). Critically, all past blocks are now masked and each
memento's KV entries are recomputed attending only to the prompt
and preceding mementos, not to the block it originally summarized.
The generated memento text is identical in both conditions; only the KV representations
differ. This isolates the question: does the information encoded in the KV
states (from having attended to the block during generation) matter beyond
what the memento text conveys?
\end{itemize}

\begin{wraptable}{r}{0.38\textwidth}
\vspace{-12pt}
\centering
\caption{\textbf{KV ablation} on full AIME24 (30 problems, Qwen3-8B memento attention checkpoint, 32K generation). Block masking accuracy is from the 64-repetition evaluation in \Cref{tab:main-sft}; the restart experiment uses 8 repetitions.}
\label{tab:kv-ablation}
\small
\begin{tabular}{lc}
\toprule
\textbf{Inference mode} & \textbf{Pass@1} \\
\midrule
Memento att.\ (normal) & 66.1\% \\
Memento att.\ + restart  & 50.8\% \\
\midrule
$\Delta$ & $-$15.3\,pp \\
\bottomrule
\end{tabular}
\vspace{-8pt}
\end{wraptable}

The 15\,pp drop confirms that memento KV states carry significant information
from the masked blocks. Mementos function as compressed pointers into cached
reasoning state, not just standalone text replacements. This distinguishes
\memento{} from prior iterative summarization methods~\cite{inftyThink2025,
r1compress2025} which discard original tokens entirely after summarization:
unlike those methods, \memento{} retains the KV cache, and this retention is
critical.

\begin{tcolorbox}[finding]
\textbf{KV states carry reasoning capacity.} Recomputing memento KVs
without block access drops AIME24 accuracy from 66.1\% to 50.8\%. Mementos are
not standalone text replacements---their cached KV representations form a
high-bandwidth implicit channel that restart-based methods discard.
\end{tcolorbox}

\begin{figure}[H]
\centering
\begin{tikzpicture}[
  seqbox/.style={rounded corners=2pt, minimum height=0.45cm,
                 font=\footnotesize\bfseries, inner sep=3pt},
  prompt/.style={seqbox, draw=cVanilla, fill=cVanilla!15, text=solBase02,
                 minimum width=0.8cm},
  think/.style={seqbox, draw=cVanilla, fill=white, text=cVanilla,
                minimum width=0.7cm},
  mem/.style={seqbox, draw=cMemento, fill=cMemento, text=white,
              minimum width=0.5cm},
  masked/.style={seqbox, draw=solBase1, fill=white, text=solBase1,
                 densely dashed, opacity=0.4, minimum width=0.7cm},
]
\def\bsep{0.08}

\node[font=\scriptsize\bfseries, text=solBase02, anchor=east] at (-0.15, 0) {Step 1};

\node[prompt] (p1) at (0.8, 0) {Prompt};
\node[masked, right=\bsep cm of p1] (t1a) {$T_1$};
\node[mem, right=\bsep cm of t1a] (m1a) {$M_1$};
\node[font=\footnotesize, text=solBase01, right=0.06cm of m1a] (d1) {$\cdots$};
\node[think, right=0.06cm of d1] (tia) {$T_i$};
\node[mem, right=\bsep cm of tia] (mia) {$M_i$};

\node[font=\scriptsize\itshape, text=solBase01, anchor=west] at ([xshift=0.2cm]mia.east)
     {$M_i$ KV includes block info};

\draw[-{Stealth[length=4pt]}, thick, solBase1]
      ([yshift=-0.35cm]$(tia.south)!0.5!(mia.south)$) -- ++(0, -0.5cm)
      node[midway, right=2pt, font=\scriptsize, text=solBase01] {discard KV, re-prefill};

\node[font=\scriptsize\bfseries, text=solBase02, anchor=east] at (-0.15, -1.5) {Step 2};

\node[prompt] (p2) at (0.8, -1.5) {Prompt};
\node[masked, right=\bsep cm of p2] (t1b) {$T_1$};
\node[mem, right=\bsep cm of t1b] (m1b) {$M_1$};
\node[font=\footnotesize, text=solBase01, right=0.06cm of m1b] (d2) {$\cdots$};
\node[masked, right=0.06cm of d2] (tib) {$T_i$};
\node[mem, right=\bsep cm of tib] (mib) {$M_i$};

\node[font=\scriptsize\itshape, text=cMemento!80!black, anchor=west] at ([xshift=0.2cm]mib.east)
     {$M_i$ KV lacks block info};

\end{tikzpicture}
\caption{\textbf{Restart ablation.} \emph{Step~1}: $M_i$ is generated with full attention to $T_i$ (same as normal memento attention). \emph{Step~2}: KV cache is discarded and recomputed via prefill over prompt $+$ $M_{1..i}$ only---$T_i$ is masked, so $M_i$'s KV states no longer encode block information. The 15\,pp accuracy drop (\Cref{tab:kv-ablation}) shows the KV channel carries significant reasoning capacity.}
\label{fig:kv-ablation}
\end{figure}
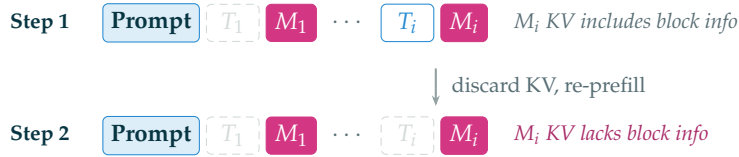

\subsubsection{Probing the Implicit KV Channel}
\label{sec:kv-probing}

The KV ablation in \Cref{sec:kv-ablation} shows that memento KV states matter for downstream accuracy.
But \emph{what} information do they carry? We design a probing experiment that injects a known signal into a masked block and measures how much of it can be recovered from downstream memento KV states that never directly attended to that block.

\paragraph{Experimental design.}
We inject a random 5-digit ``passcode'' (00000--99999) into the content of a target block $T_2$ in a real AIME'25 reasoning trace, then run a forward pass with \memento{} block masking (\texttt{keep\_last\_n\_blocks=0}). We extract KV states (keys and values concatenated) from memento token positions at specific layers and train a probe (MLP, 512$\times$256 hidden units, 128 bottleneck) to predict the 5 individual digits from these features. 
We report the average accuracy across the 5 predictions (one for each passcode digit) with the random prediction baseline being $10 \%$.
Crucially, the validation split is \emph{label-unique}: no digit combination appears in both train and validation, preventing memorization.

We evaluate three probing conditions:
\begin{itemize}[leftmargin=*,itemsep=2pt]
  \item \textbf{Direct:} Probe the KV states of memento $M_2$, which \emph{can} attend to the target block $T_2$. This measures the upper bound of information encoded in a single memento's KV states.
  \item \textbf{Masked:} Probe the KV states of memento $M_3$, which \emph{cannot} attend to $T_2$ as $T_2$ has already been evicted from the KV cache by the time $M_3$ is computed. Any signal recovered here must have propagated through the memento chain.
  \item \textbf{Causal control:} Probe the KV states of memento $M_1$, which \emph{precedes} the target block $T_2$ in the sequence. Since $M_1$ is computed before $T_2$ is even generated, it cannot contain any information about the passcode. This serves as a sanity check; we expect chance-level accuracy (10\%).
\end{itemize}

We run this experiment at two scales:
\begin{itemize}[leftmargin=*]
  \item \textbf{Qwen3-8B}: 15K samples from AIME'25 traces generated by the 8B model, with injected passcodes, probing at layers 3 and 35.
  \item \textbf{Qwen3-32B}: 15K samples from AIME'25 traces generated by the 32B model, with injected passcodes, probing at layers 3 and 63.
\end{itemize}

\begin{figure}[H]
    \centering
    \begin{minipage}[t]{0.48\textwidth}
        \centering
        \vspace{-.1in}
        \includegraphics[width=\textwidth]{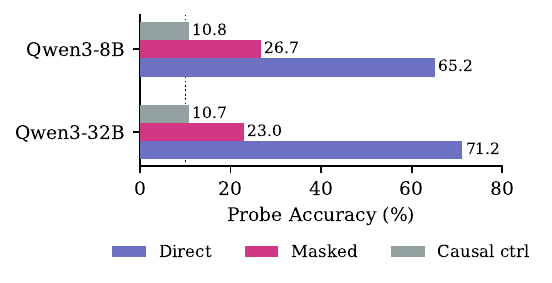}
        \captionof{figure}{\textbf{Probing the implicit KV channel.}
        Both Qwen3-8B and Qwen3-32B recover the passcode well above 10\% chance from \emph{masked} memento positions (26.7\% and 23.0\%), while causal controls show exactly chance-level accuracy.
        The dotted line marks 10\% chance (random guessing over 10 digits).}
        \label{fig:kv-probing}
    \end{minipage}
    \hfill
    \begin{minipage}[t]{0.48\textwidth}
        \centering
        \vspace{0pt}
        \captionof{table}{\textbf{Deeper layers carry the signal.} Per-layer probe accuracy~(\%) under direct and masked conditions (\texttt{keep0}). The leaked signal concentrates in deeper layers; early layers show near-chance masked accuracy.}
        \label{tab:kv-perlayer}
        \vspace{0.3em}
        \small
        \setlength{\tabcolsep}{4pt}
        \begin{tabular}{l cc cc}
        \toprule
        & \multicolumn{2}{c}{\textbf{Qwen3-8B}} & \multicolumn{2}{c}{\textbf{Qwen3-32B}} \\
        \cmidrule(lr){2-3} \cmidrule(lr){4-5}
        & Direct & Masked & Direct & Masked \\
        \midrule
        4th layer (early)    & 51.6 & 10.8 & 53.8 & 12.8 \\
        Last layer & 64.9 & 26.5 & 68.7 & 22.4 \\
        Both layers        & 65.2 & 26.7 & 71.2 & 23.0 \\
        \midrule
        Chance             & \multicolumn{4}{c}{10.0} \\
        \bottomrule
        \end{tabular}
    \end{minipage}
\end{figure}

\paragraph{Results.}
\Cref{fig:kv-probing} summarizes the findings.
At the direct position, the memento text itself bears no relation to the passcode, yet the KV states recover the injected digits with 60--70\% accuracy, demonstrating that KV representations encode far more information than the corresponding tokens.
At the masked position, where the memento \emph{cannot} attend to the target block, both models still recover the passcode well above chance (26.7\% for Qwen3-8B, 23.0\% for Qwen3-32B vs.\ 10\% chance).
The causal control, probing a memento that \emph{precedes} the target block, shows exactly chance-level accuracy, confirming that the recovered signal is real and directional.
\Cref{tab:kv-perlayer} further shows that leakage concentrates in deeper layers. In Qwen3-8B, an early layer (the 4th) shows near-chance masked accuracy (10.8\%) while the last layer (the 36th) reaches 26.5\%; the pattern repeats in Qwen3-32B (12.8\% at the 4th layer vs.\ 22.4\% at the 64th). The same trend holds for the direct condition, where deeper layers carry substantially more signal (64.9\% vs.\ 51.6\% for 8B; 68.7\% vs.\ 53.8\% for 32B). This is consistent with the residual stream accumulating information across layers.
We further validate these findings with a controlled toy transformer experiment (\Cref{sec:toy-kv-probing}): a 4-layer model trained on synthetic data exhibits the same leakage pattern (24.9\% masked accuracy vs.\ 10\% chance), with signal decaying gradually over distance but persisting up to 7 hops from the target block. Leakage remains constant across training checkpoints even as task accuracy improves, confirming the channel is architectural---not learned.

\begin{tcolorbox}[finding]
\textbf{KV states carry an implicit information channel.} Memento KV representations propagate block information across masked boundaries---an architectural effect that complements the explicit memento text and explains why single-pass \memento{} outperforms restart-based methods.
\end{tcolorbox}

\section{Conclusion}
\label{sec:conclusion}

We introduced \memento{}, a method that teaches language models to manage their own context by segmenting reasoning into blocks, compressing each into a dense memento, and masking completed blocks via sparse attention.
Across three model families (Qwen3, Phi-4-reasoning, Olmo-3-7B-Think), \memento{} reduces peak KV cache by $2$--$3\times$ and KV AUC by up to $3.5\times$, translating to $1.75\times$ higher serving throughput, while preserving strong reasoning accuracy: Qwen3-32B loses just 2.6\,pp on AIME'26 and 3.5\,pp averaged across five benchmark groups.
The gap shrinks with scale (6.3\,pp at 8B $\to$ 3.5\,pp at 32B), and our initial CISPO RL result on Qwen3-8B recovers much of the remaining single-sample gap while retaining the KV savings.

A key finding is that mementos carry information from masked blocks through two complementary channels: the explicit summary text and the implicit KV representations computed while the block was still visible. Our KV ablation shows that removing this implicit channel degrades accuracy by 15\,pp, distinguishing \memento{} from methods that simply discard context after summarization.

Looking forward, we see two natural extensions: scaling the RL recipe to larger models, and applying \memento{} to long-horizon agent tasks where agent steps form natural blocks and context windows are the primary bottleneck. We release \openmementos{} (228K annotated reasoning traces) and our vLLM fork with native block masking support to facilitate further research.

\newpage
\appendix
\section{Supplementary Material}
\label{sec:appendix}

\subsection{Dataset Construction Details}
\label{appendix:sec2-details}

\subsubsection{Full Prompts}
\label{appendix:prompts}

This section contains the complete prompts used in the \memento{} data labeling pipeline.

\paragraph{Boundary Scoring Prompt.}
\label{appendix:boundary-prompt}

\begin{tcolorbox}[
  enhanced, breakable, colback=white, colframe=black!70,
  fonttitle=\bfseries, title={Boundary Scoring System Prompt}
]
\small\ttfamily
You are an expert at analyzing chain-of-thought reasoning. Your task is to score potential breakpoints in a reasoning trace.

For each boundary between sentences, assign a score from 0-3:\\
- 0: Poor break (mid-thought, would disrupt flow)\\
- 1: Weak break (minor transition)\\
- 2: Good break (clear transition, coherent endpoint)\\
- 3: Strong break (major transition, natural chapter boundary)

Consider:\\
- Semantic coherence (does previous sentence complete a thought?)\\
- Topic shifts (does next sentence start new topic?)\\
- Logical flow (would breaking here preserve reasoning structure?)

CRITICAL RULES FOR MATHEMATICAL DERIVATIONS:\\
- NEVER give high scores (2-3) to boundaries in the middle of a calculation\\
- If previous sentence ends with ":" (colon), ALWAYS score 0\\
- If previous sentence ends with "=", "=>", or introduces a calculation, score 0\\
- If next sentence starts with "Therefore", "Thus", "Hence" continuing a derivation, score 0\\
- Multi-step calculations must stay together (score 0-1)\\
- Only score 2-3 when the derivation COMPLETES and topic shifts

Output JSON: \{"scores": [score1, score2, ...]\}
\end{tcolorbox}

\paragraph{Summarizer Prompt.}
\label{appendix:summarizer-prompt}

\begin{tcolorbox}[
  enhanced, breakable, colback=white, colframe=black!70,
  fonttitle=\bfseries, title={Summarizer System Prompt}
]
\small\ttfamily
You are a STATE-COMPRESSOR for long reasoning traces produced by advanced models.

You receive the FULL reasoning trace partitioned into blocks. Your ONLY job is to produce an extremely information-dense, lemma-like STATE SUMMARY for each block.

== CORE OBJECTIVE ==

Minimize the number of tokens in each summary, subject to fully capturing all logically relevant information:
- Definitions, variables, functions, data structures
- Assumptions, constraints, case splits
- Key intermediate results: equations, inequalities, derived identities
- Chosen strategies, algorithmic ideas, invariants
- Important rejected attempts

You MUST NOT: omit facts needed later, invent new facts, or paraphrase so aggressively that conclusions become ambiguous.

== NO NEW REASONING ==

Behave as a purely extractive compressor. Do NOT derive new values, repair mistakes, or re-solve the problem.

== COMPRESSION STYLE ==

Target: $\sim$10\% of block tokens ($\leq$20\%). Terse, lemma-like, not literary. Prefer compact symbolic notation. Example: "Let f(n)=...; assume n>=3; derived f(n)=3n-2>0 for n>=3."
\end{tcolorbox}

\paragraph{Judge Prompt.}
\label{appendix:judge-prompt}

\begin{tcolorbox}[
  enhanced, breakable, colback=white, colframe=black!70,
  fonttitle=\bfseries, title={Judge System Prompt}
]
\small\ttfamily
You are a SUMMARY QUALITY JUDGE for compressed reasoning traces.

Evaluate whether a summary successfully extracts all critical information such that it can REPLACE the original block entirely.

== SCORING RUBRIC (0-10) ==

FORMULAS \& EQUATIONS (0-3): All key formulas extracted verbatim with complete notation.\\
NUMERICAL VALUES (0-2): All critical intermediate and final numerical values included.\\
METHODS \& TECHNIQUES (0-2): Explicitly names approach used.\\
VALIDATION (0-1): If block contains verification, summary includes outcomes.\\
CORRECTNESS (0-1): Only confirmed findings; excludes wrong intermediate steps.\\
STRUCTURE (0-1): Leads with findings before process.

DEDUCTIONS: -1 for hallucination, -1 for process-focused narrative, -2 for missing critical formula.

FEEDBACK must be SPECIFIC and ACTIONABLE:\\
- BAD: "More details needed"\\
- GOOD: "Missing formula: $K^2-3K+3$ from line 45"
\end{tcolorbox}

\subsubsection{Worked Examples}

\begin{figure}[H]
\centering
\begin{tcolorbox}[colback=findingbg, colframe=findingborder, boxrule=0.4pt, arc=2pt, left=5pt, right=5pt, top=3pt, bottom=3pt, fontupper=\footnotesize, title={\footnotesize\textbf{Memento refinement example} --- NBA playoff probability trace, Block~2 (876~tokens $\to$ memento)}, coltitle=black, colbacktitle=findingbg]
\textbf{Block excerpt:} \textit{``\ldots defines f(n,a,b) as the probability that after n games, Team A has a wins and Team B has b wins. Starting point: f(0,0,0)=1. Need f(6,3,3). For Game n: if n is odd, home=A, P(A wins)=0.6; if even, home=B, P(B wins)=0.6 \ldots''}

\vspace{4pt}
\textbf{Initial memento} (pass~1, score 5/10):\quad
{\small\texttt{Defines target probability P[(A,B)=(3,3) after 6]. Home team alternates each game; home win prob=0.6. Proposes DP approach and considers enumeration of all sequences.}}

\vspace{4pt}
\textbf{Judge feedback:}\quad
{\small\itshape Missing formula: recurrence f(n,a,b) with starting condition f(0,0,0)=1. Missing: explicit transition probabilities for odd/even games. Replace ``proposes DP approach'' with the named state variables and recurrence.}

\vspace{4pt}
\textbf{Refined memento} (pass~2, score 8/10):\quad
{\small\texttt{Defines f(n,a,b)=P(A has a wins, B has b wins after n games); f(0,0,0)=1; target f(6,3,3). Home pattern: odd games home=A, even home=B; p\_homewin=0.6. Transitions: if home=A, f(n,a,b)+=f(n-1,a-1,b)*0.6+f(n-1,a,b-1)*0.4; if home=B, swap.}}
\end{tcolorbox}
\caption{Iterative memento refinement on a reasoning block about NBA playoff probabilities. The initial memento (pass~1, score~5/10) describes the \emph{approach} but omits critical formulas. After judge feedback requesting the specific recurrence and transition probabilities, the refined memento (pass~2, score~8/10) captures the full computational state: function definition, base case, target, and recurrence relation.}
\label{fig:refinement-example}
\end{figure}

\subsection{Training and Evaluation Details}
\label{appendix:sec3-details}

\subsubsection{Training Details}
\label{appendix:training}

This section provides full details of the SFT training pipeline, complementing the high-level description in \Cref{sec:experiments}.

\paragraph{Training Configuration and Hyperparameters.}
\label{appendix:training-config}
\label{appendix:training-framework}

All SFT experiments use TRL's \texttt{SFTTrainer}~\citep{vonwerra2022trl} with PyTorch 2.8+ and its native SDPA attention backend. All models are trained on 31K samples from \openmementos{} with 32K sequence length on 32 NVIDIA B200 GPUs (4 nodes $\times$ 8 GPUs, 192\,GB HBM per GPU). All runs share the same hyperparameters: AdamW optimizer ($\beta_1{=}0.9$, $\beta_2{=}0.999$, no weight decay), learning rate $8 \times 10^{-5}$ with cosine schedule and 5\% warmup, 5 epochs per stage, gradient clipping at 1.0, global batch size 512, bfloat16 precision, gradient checkpointing, and seed 42. Checkpoints are saved every 50 steps (100 for Qwen3-32B). Qwen3-8B and Olmo-3-7B fit without model sharding; Phi-4 uses DeepSpeed ZeRO-2; Qwen3-32B requires ZeRO-3.

\paragraph{Two-Stage Training Procedure.}
\label{appendix:training-stages}

\paragraph{Stage~1 (Full Attention).} The environment variable \texttt{KEEP\_LAST\_N\_BLOCKS=-1} disables block masking. Loss is computed on all completion tokens (prompt tokens are masked via \texttt{labels=-100}). Checkpoints are saved at regular intervals and evaluated on AIME24 to select the best checkpoint for Stage~2.

\paragraph{Stage~2 (Memento Attention).} The environment variable \texttt{KEEP\_LAST\_N\_BLOCKS} is set to $0$. The model is initialized from the best Stage~1 checkpoint and trained with identical hyperparameters. Loss is computed on all tokens, identical to Stage~1. The only difference from Stage~1 is the attention pattern: the custom block-masked attention implementation (\Cref{appendix:attention-mask}) applies a sparse attention mask during the forward pass, ensuring that tokens after a completed block+summary cannot attend to the masked block content.

\paragraph{Sparse Attention Mask Implementation.}
\label{appendix:attention-mask}

We maintain custom model forks for each architecture family (Qwen3, Phi-4, Olmo 3) that modify the \texttt{forward()} method to support block masking during both training and inference. The implementation works as follows:

\begin{enumerate}[leftmargin=*]
  \item A \textbf{block cache} tracks the position type of every token: \textsc{block}, \textsc{summary}, or \textsc{other}. It detects the learned special tokens (\texttt{<|block\_start|>}, \texttt{<|block\_end|>}, \texttt{<|summary\_start|>}, \texttt{<|summary\_end|>}) in the token stream.
  \item When \texttt{<|summary\_end|>} is generated, the block cache marks the preceding thinking block as \textbf{completed}. All KV-cache entries for that block's reasoning tokens are masked from future queries.
  \item The block cache is \textbf{stateful across autoregressive steps}, persisting through the KV cache. This avoids re-scanning the full sequence at every generation step.
  \item For \textbf{training}, the full sequence is available, so the attention mask is constructed upfront as a dense mask matrix that zeroes out attention from post-summary tokens to their corresponding block content.
\end{enumerate}

\paragraph{Special Token Initialization.}
\label{appendix:special-tokens}

Four special tokens are added to the tokenizer vocabulary: \texttt{<|block\_start|>}, \texttt{<|block\_end|>}, \texttt{<|summary\_start|>}, \texttt{<|summary\_end|>}. Their embeddings are initialized as the mean of semantically related existing tokens plus small Gaussian noise ($\sigma{=}0.01$):
\begin{itemize}[leftmargin=*]
  \item \texttt{<|block\_start|>} $\leftarrow$ mean(\textit{block}, \textit{start}, \textit{begin}, \textit{section}, \textit{step})
  \item \texttt{<|block\_end|>} $\leftarrow$ mean(\textit{block}, \textit{end}, \textit{finish}, \textit{section}, \textit{done})
  \item \texttt{<|summary\_start|>} $\leftarrow$ mean(\textit{summary}, \textit{summarize}, \textit{brief}, \textit{recap}, \textit{overview})
  \item \texttt{<|summary\_end|>} $\leftarrow$ mean(\textit{summary}, \textit{end}, \textit{finish}, \textit{done}, \textit{complete})
\end{itemize}

\paragraph{Data Format.}
\label{appendix:data-format}

Training data is pre-tokenized into HuggingFace Arrow format for efficiency. Each example is formatted in ChatML:
\begin{itemize}[leftmargin=*]
  \item \textbf{User message:} the problem statement.
  \item \textbf{Assistant response:} \texttt{<think>} $+$ [\texttt{<|block\_start|>} reasoning \texttt{<|block\_end|>} \texttt{<|summary\_start|>} summary \texttt{<|summary\_end|>}]$^*$ $+$ \texttt{</think>} $+$ final answer
\end{itemize}
For Qwen3 models, no system prompt is used (matching Qwen3's default behavior). For Phi-4, the native system prompt is preserved. Sequences are tokenized to a maximum length of 32{,}768 tokens with truncation; no padding is applied during tokenization (the data collator handles dynamic padding at batch time).

\subsubsection{Evaluation Details}
\label{appendix:evaluation}

\paragraph{Inference Backend.}
\label{appendix:eval-backend}

We evaluate all models using a standalone evaluation script (\texttt{evaluate\_vllm.py}) built on a custom vLLM fork (branch \texttt{token-span-removal}) that implements native \texttt{BlockMaskingConfig} support for KV-cache-level block masking. The fork is installed as a Python overlay on top of the container's vLLM installation. The script operates in two modes:
\begin{enumerate}[leftmargin=*]
  \item \textbf{Offline mode} (default): Uses vLLM's Python \texttt{LLM} class for fast batched generation. All problems $\times$ repetitions are submitted as a single batch, and vLLM handles scheduling internally.
  \item \textbf{Server mode} (\texttt{--track\_kv}): Launches vLLM as OpenAI-compatible HTTP servers (one per GPU), sends requests sequentially per GPU, and polls the \texttt{/metrics} endpoint to record per-request KV cache usage over time. This mode produces \texttt{kv\_trace.csv} files used for the KV cache simulation validation in \Cref{sec:kv-validation}.
\end{enumerate}

\paragraph{Block masking configuration.} The \texttt{BlockMaskingConfig} specifies:
\begin{itemize}[leftmargin=*]
  \item \texttt{keep\_last\_n\_blocks}: $-1$ (vanilla, no masking), $0$ (memento attention, compact all blocks), or $N$ (keep last $N$ blocks visible).
  \item \texttt{mask\_delimiters}: \texttt{False} for Qwen3 and Olmo 3 (block delimiters remain visible); \texttt{True} for Phi-4 (delimiters are also masked).
  \item \texttt{compact\_on\_summary\_end}: \texttt{True}---block content is evicted from the KV cache when \texttt{<|summary\_end|>} is generated.
  \item \texttt{enable\_prefix\_caching}: \textbf{must be \texttt{False}} when block masking is active, since block eviction modifies the KV cache in ways incompatible with prefix sharing.
\end{itemize}

\paragraph{Generation Parameters.}
\label{appendix:eval-generation}

All evaluations use the same generation parameters unless otherwise noted: temperature~0.6, top-$p$~0.95, top-$k$~20, min-$p$~0.0, max new tokens 32{,}000, and \texttt{skip\_special\_tokens=False} (to preserve block/summary tokens in the output for post-hoc analysis). Competition-math benchmarks (AIME, HMMT, BrUMO, SMT, CMIMC; $\leq$53 problems each) receive 64 independent generations per problem. Larger benchmarks (MATH-500, GPQA-Diamond, LiveCodeBench) receive 2 generations per problem. We report pass@1 accuracy: the per-problem correct fraction averaged across all generations.

\paragraph{Hardware and Parallelism.}
\label{appendix:eval-hardware}

Evaluations run on NVIDIA B200 GPUs. All models use tensor parallelism (TP) = 1 with data parallelism (DP) = 8 across 8 GPUs on a single node. The vLLM engine is configured with \texttt{max\_model\_length=32768}, \texttt{max\_num\_batched\_tokens=32768}, and \texttt{gpu\_memory\_utilization=0.85}.

\paragraph{Benchmarks.}
\label{appendix:eval-benchmarks}

\Cref{tab:benchmark-details} summarizes the 14 benchmarks used in our evaluation. All benchmarks use 0-shot evaluation (no few-shot examples).

\begin{table}[H]
\centering
\caption{\textbf{Benchmark details.} Competition-math benchmarks use 64 generations per problem; MATH-500, GPQA, and LiveCodeBench use 2 generations. All use temperature 0.6 and 32K max output tokens.}
\label{tab:benchmark-details}
\small
\setlength{\tabcolsep}{4pt}
\begin{tabular}{llcl}
\toprule
\textbf{Benchmark} & \textbf{Source} & \textbf{\# Problems} & \textbf{Answer format} \\
\midrule
AIME 2024 & HuggingFaceH4/aime\_2024 & 30 & Integer (0--999), \texttt{\textbackslash boxed\{\}} \\
AIME 2025 & yentinglin/aime\_2025 & 30 & Integer (0--999), \texttt{\textbackslash boxed\{\}} \\
AIME 2026 & MathArena/aime\_2026 & 30 & Integer (0--999), \texttt{\textbackslash boxed\{\}} \\
HMMT Feb 2023 & MathArena/hmmt\_feb\_2023 & 30 & Math expression \\
HMMT Feb 2024 & MathArena/hmmt\_feb\_2024 & 30 & Math expression \\
HMMT Feb 2025 & MathArena/hmmt\_feb\_2025 & 30 & Math expression \\
HMMT Feb 2026 & MathArena/hmmt\_feb\_2026 & 33 & Math expression \\
HMMT Nov 2025 & MathArena/hmmt\_nov\_2025 & 30 & Math expression \\
BrUMO 2025 & MathArena/brumo\_2025 & 30 & Math expression \\
SMT 2025 & MathArena/smt\_2025 & 53 & Math expression \\
CMIMC 2025 & MathArena/cmimc\_2025 & 40 & Math expression \\
MATH-500 & HuggingFaceH4/MATH-500 & 500 & Math expression \\
GPQA-Diamond & Idavidrein/gpqa & 198 & Multiple choice (A--D) \\
LiveCodeBench v6 & lighteval/code\_generation\_lite & 1{,}055 & Python code (execution-based) \\
\bottomrule
\end{tabular}
\end{table}

\paragraph{Answer verification.} For mathematics benchmarks (AIME, HMMT, BrUMO, SMT, CMIMC, MATH-500), answer verification follows a two-stage pipeline adapted from OlmoMathReward~\citep{olmo3}:
\begin{enumerate}[leftmargin=*]
  \item \textbf{Candidate extraction}: Multiple strategies are tried in order---\texttt{\textbackslash boxed\{\ldots\}} content, ``Final Answer: \ldots'' patterns, last \texttt{\$\ldots\$} content, and raw normalized text.
  \item \textbf{Equivalence checking}: Each candidate is compared against the ground truth using two methods: (a)~SymPy-based symbolic equivalence (LaTeX is parsed to symbolic expressions and their difference is simplified with a 5-second timeout), and (b)~Hendrycks-style string normalization (strip \texttt{\textbackslash left/\textbackslash right} delimiters, \texttt{\textbackslash dfrac$\to$\textbackslash frac}, whitespace, and unit strings). A candidate is accepted if either method succeeds.
\end{enumerate}
For GPQA-Diamond, we extract the last letter choice (A/B/C/D) from the response and compare against the gold label. For LiveCodeBench, generated Python code is executed against public and private test cases; a solution passes only if all test cases succeed.

\paragraph{Competition-math benchmarks.}
All eleven competition-math benchmarks are sourced from MathArena~\citep{balunovic2025matharena}, a platform that evaluates LLMs on recently held math competitions to minimize contamination risk.
The individual competitions are:
\begin{itemize}[leftmargin=*]
  \item \textbf{AIME} (American Invitational Mathematics Examination, 2024/2025/2026): A pre-olympiad competition administered by the Mathematical Association of America (MAA). Each year comprises two 15-problem papers (AIME~I and~II, 30 total) with integer answers in the range 0--999.
  \item \textbf{HMMT} (Harvard-MIT Mathematics Tournament, Feb 2023/2024/2025/2026 and Nov 2025): A major university-organized competition with problems in algebra, combinatorics, geometry, and number theory. February and November tournaments are held separately; most editions contribute 30 problems (HMMT Feb 2026 has 33).
  \item \textbf{BrUMO} (Brown University Mathematics Online, 2025): An online math competition organized by Brown University; 30 problems.
  \item \textbf{SMT} (Stanford Mathematics Tournament, 2025): A competition organized by Stanford University students covering algebra, combinatorics, geometry, and number theory; 53 problems.
  \item \textbf{CMIMC} (Carnegie Mellon Informatics and Mathematics Competition, 2025): A competition organized by Carnegie Mellon University; 40 problems spanning algebra, combinatorics, geometry, and number theory.
\end{itemize}
All competition datasets are available on HuggingFace under the \texttt{MathArena} organization (\url{https://huggingface.co/MathArena}).

\paragraph{Metrics.} We report pass@1 accuracy: for each problem, we average the binary correctness indicator across all generations (64 for competition math, 2 for MATH-500/GPQA/LCB), then average across problems. For the main table (\Cref{tab:main-sft}), competition math (``Comp.\ Math'') is the unweighted average of pass@1 across eleven competition-math benchmarks (AIME'24/25/26, HMMT~Feb'23/24/25/26, Nov'25, BrUMO'25, SMT'25, CMIMC'25). Standard errors are computed as the standard error of per-problem means across problems.

\subsubsection{RLVR Training Details}
\label{appendix:rl-details}

\paragraph{Rollout infrastructure.}
During RL rollouts, block masking must be active so that the policy generates under the same conditions as deployment. We use our custom vLLM engine (\Cref{sec:inference}) with \texttt{BlockMaskingConfig(enable=True, keep\_last\_n\_blocks=0)} to physically compact KV cache entries after each summary, exactly as at inference time. Training forward passes use a corresponding sparse block-masked attention implementation so that the gradient computation matches the masked generation pattern.

\paragraph{Training configuration.}
We fine-tune the Qwen3-8B \memento{} SFT checkpoint (the keep-0 model from Stage~2 of our two-stage SFT recipe) using the DolciMath training set (29,670 prompts) with rule-based \texttt{olmo\_math} rewards (sympy-based LaTeX answer verification, no LLM judge). Key hyperparameters:

\begin{itemize}[leftmargin=*, itemsep=2pt]
\item \textbf{Compute:} 24 GPUs (3 nodes $\times$ 8), FSDP with optimizer offload
\item \textbf{Group size:} 8 rollouts per prompt (train), 64 per prompt (eval)
\item \textbf{Batch size:} 240 prompts per step
\item \textbf{Optimizer:} AdamW, $\text{lr}=10^{-6}$, cosine warmup (10 steps), gradient clipping 1.0
\item \textbf{Clipping:} $\epsilon_\text{low} = \epsilon_\text{high} = 0.2$
\item \textbf{KL coefficient:} $\beta = 0.001$
\item \textbf{Normalization:} batch-level advantage normalization, group-total loss normalization
\item \textbf{Sampling:} temperature 1.0 (train), 0.6 (eval); prompt length 1024, response length 31744
\item \textbf{Evaluation:} AIME'24 and AIME'25 every 25 steps
\item \textbf{Selected checkpoint:} step 350 (best AIME'25 validation at 66.2\%)
\end{itemize}

\paragraph{Detailed results.}
After 350 steps, the \memento{}+RL model improves across all benchmarks relative to \memento{} SFT: AIME'26 rises from 57.3\% to 64.9\%, Comp.\ Math from 45.1\% to 49.4\%, GPQA-D from 55.8\% to 62.9\% (above the 61.4\% vanilla baseline), LCB~v6 from 66.5\% to 68.8\%, and MATH-500 from 90.1\% to 91.0\%. The KV footprint increases modestly: peak KV rises from 1.08 to 1.48\,GB (still 45\% below vanilla's 2.71\,GB) and KV AUC from 10.7 to 16.4\,GB$\cdot$ktok (roughly half of vanilla's 30.9). RL trades a small amount of compression for substantially improved single-sample accuracy.

\subsection{Inference and KV Cache Details}
\label{appendix:sec4-details}

\subsubsection{KV Cache Simulation Validation}
\label{sec:kv-validation}

The KV cache metrics in \Cref{tab:main-sft} are computed via offline simulation rather than live profiling, enabling measurement across all 14 benchmarks and 64 repetitions per problem ($>$80{,}000 total completions). We validate this approach against real KV cache tracking data.

\paragraph{Simulation procedure.}
For each generated response, we tokenize the full output and replay generation step-by-step. At each step, we track the number of tokens in the KV cache, detecting \texttt{<|block\_start|>}, \texttt{<|block\_end|>}, and \texttt{<|summary\_end|>} tokens. In memento attention mode, all completed blocks are evicted from the KV cache when the summary ends. This produces a per-token KV trajectory from which we extract peak KV (maximum occupancy) and average KV (area under curve / generation length).

\paragraph{Token-to-GB conversion.}
We convert token counts to memory using the exact model architecture:
\[
\text{GB} = \text{tokens} \times \underbrace{2}_{\text{K+V}} \times n_{\text{layers}} \times n_{\text{kv\_heads}} \times d_{\text{head}} \times \underbrace{2}_{\text{bf16 bytes}} \;/\; 10^9
\]
This yields 144\,KB/token for Qwen3-8B (36 layers, 8 KV heads, $d=128$), 256\,KB/token for Qwen3-32B (64 layers), and 200\,KB/token for Phi4-RP (40 layers, 10 KV heads).

\paragraph{Validation against real measurements.}
We previously ran evaluations with vLLM's KV cache tracking enabled (Prometheus metric polling) on Qwen3-8B/32B across 4 benchmarks in vanilla and memento attention modes (12{,}680 paired observations on 1$\times$B200 GPU, TP=1). For each sample, we compare the simulated peak KV (in tokens) against the real measured peak KV (as a fraction of the GPU's KV pool). The linear correlation yields $R^2 > 0.999$ for all model/benchmark/mode combinations, with mean absolute error below 0.02\,GB. The measured KV pool sizes are consistent with the B200's 192\,GB HBM: after accounting for model weights and vLLM overhead (${\sim}$6--16\,GB), the observed pools match theoretical predictions to within 6--10\%.

\subsubsection{vLLM Block Masking Implementation}
\label{appendix:vllm-implementation}

We extend vLLM (branch \texttt{token-span-removal}) with a \texttt{BlockMaskingConfig} that enables KV-cache-level block compaction during autoregressive generation, requiring no changes to the model weights or attention kernels. Even systems that implement fixed sparse patterns (e.g., DeepSeek-V3's Multi-head Latent Attention~\citep{deepseekv3}) do so by modifying the model architecture rather than providing a general-purpose masking API.
The implementation has three components:

\textit{(i)~Per-request state machine.}
Each request carries a \texttt{BlockMaskingState} that tracks open blocks, completed blocks, and pending compactions.
A lightweight \texttt{BlockMaskingProcessor} inspects each generated token for the four special tokens (\texttt{<|block\_start|>}, \texttt{<|block\_end|>}, \texttt{<|summary\_start|>}, \texttt{<|summary\_end|>}) and drives state transitions.
When \texttt{<|summary\_end|>} is produced, the corresponding reasoning block is marked for compaction.

\textit{(ii)~Physical KV cache compaction.}
Unlike logical masking (which would require custom attention kernels), we \emph{physically} remove masked tokens from the KV cache.
When a block is compacted, the scheduler computes the set of active (non-masked) logical positions, translates them to physical KV cache locations, and issues a \texttt{compact\_kv\_cache} operation that copies the active entries into contiguous slots and frees trailing KV blocks.
This means standard FlashAttention and paged-attention kernels work unmodified---they simply never see the evicted tokens.
A logical-to-physical translation layer (via sorted spans and binary search) maintains $O(\log n)$ position lookups across multiple compactions per request.

\textit{(iii)~Scheduler integration.}
The scheduler orchestrates compaction between forward passes: it collects pending KV copy operations and block table truncations, dispatches them to the GPU worker, and updates the request's \texttt{num\_computed\_tokens}.
In restart mode (\Cref{sec:kv-ablation}), the scheduler additionally rewinds the request to the summary start position, triggering a re-prefill of memento tokens with a clean KV cache from which block content has been removed.

The \texttt{keep\_last\_n\_blocks} parameter controls compaction aggressiveness: $-1$ disables masking (vanilla), $0$ compacts all completed blocks (memento attention), and $N{>}0$ keeps the last $N$ blocks visible while compacting older ones. This provides a knob for trading off KV savings against the informational value of recent block context.

\subsubsection{Throughput Details}

See \Cref{fig:throughput-full} in the main text for the throughput figure.

\subsection{Additional Experiments and Ablations}
\label{sec:ablations}

\subsubsection{Multi-Stage SFT Ablation}
\label{sec:multistage}

Our two-stage SFT recipe is a key design choice. What happens if we skip stages? We compare five training strategies on Qwen2.5-7B, varying the curriculum:
\begin{itemize}[leftmargin=*]
    \item \textbf{OT}: trained on OpenThoughts-v3 only (standard reasoning SFT).
    \item \textbf{OM/Full}: trained directly from the base model on \openmementos{} with full causal loss.
    \item \textbf{OM/Mem}: trained directly from the base model on \openmementos{} while masking all thinking-block content.
    \item \textbf{OT $\to$ OM/Full}: first trained on OpenThoughts-v3, then fine-tuned on \openmementos{} with full causal loss.
    \item \textbf{OT $\to$ OM/Full $\to$ OM/Mem (Ours)}: first train a reasoning model, then apply two-stage memento SFT.
\end{itemize}

\begin{figure}[H]
    \centering
    \pgfplotsset{
      mstage/.style={
        ybar, bar width=12pt, bar shift=0pt,
        enlarge x limits=0.18,
        width=0.35\textwidth, height=5.5cm,
        ylabel style={font=\small},
        tick label style={font=\footnotesize},
        ymajorgrids=true, grid style={cGrid, thin},
        ymin=0,
        symbolic x coords={OT,{OM/Full},{OM/Mem},{2-Stg},{3-Stg}},
        xtick={OT,{OM/Full},{OM/Mem},{2-Stg},{3-Stg}},
        x tick label style={rotate=35, anchor=east, font=\scriptsize},
        error bars/y dir=both,
        error bars/y explicit,
        error bars/error bar style={black, line width=0.8pt},
        error bars/error mark options={rotate=90, mark size=3pt, line width=0.8pt, black},
      },
    }
    \begin{tikzpicture}
      \begin{groupplot}[
        group style={group size=3 by 1, horizontal sep=0.5cm},
        mstage,
      ]
        \nextgroupplot[
          title={\small AIME24}, title style={at={(0.5,1.02)}},
          ylabel={Pass@1 (\%)}, ymax=56,
        ]
        \addplot[fill=cVanilla,  draw=cVanilla!70!black]  coordinates {(OT,38.0)       +- (0,7.32)};
        \addplot[fill=solCyan,   draw=solCyan!70!black]   coordinates {({OM/Full},24.0)  +- (0,6.67)};
        \addplot[fill=solOrange, draw=solOrange!70!black]  coordinates {({OM/Mem},20.7)  +- (0,5.29)};
        \addplot[fill=solViolet, draw=solViolet!70!black]  coordinates {({2-Stg},44.7)  +- (0,7.35)};
        \addplot[fill=cMemento,  draw=cMemento!70!black]  coordinates {({3-Stg},32.7)  +- (0,6.61)};

        \nextgroupplot[
          title={\small AIME25}, title style={at={(0.5,1.02)}},
          ymax=42,
        ]
        \addplot[fill=cVanilla,  draw=cVanilla!70!black]  coordinates {(OT,32.0)       +- (0,7.65)};
        \addplot[fill=solCyan,   draw=solCyan!70!black]   coordinates {({OM/Full},24.0)  +- (0,6.74)};
        \addplot[fill=solOrange, draw=solOrange!70!black]  coordinates {({OM/Mem},21.3)  +- (0,6.36)};
        \addplot[fill=solViolet, draw=solViolet!70!black]  coordinates {({2-Stg},28.0)  +- (0,7.21)};
        \addplot[fill=cMemento,  draw=cMemento!70!black]  coordinates {({3-Stg},28.7)  +- (0,6.83)};

        \nextgroupplot[
          title={\small GPQA-D}, title style={at={(0.5,1.02)}},
          ymax=56,
        ]
        \addplot[fill=cVanilla,  draw=cVanilla!70!black]  coordinates {(OT,42.4)       +- (0,3.52)};
        \addplot[fill=solCyan,   draw=solCyan!70!black]   coordinates {({OM/Full},38.9)  +- (0,3.47)};
        \addplot[fill=solOrange, draw=solOrange!70!black]  coordinates {({OM/Mem},34.3)  +- (0,3.38)};
        \addplot[fill=solViolet, draw=solViolet!70!black]  coordinates {({2-Stg},42.4)  +- (0,3.52)};
        \addplot[fill=cMemento,  draw=cMemento!70!black]  coordinates {({3-Stg},45.5)  +- (0,3.55)};

      \end{groupplot}
    \end{tikzpicture}
    \caption{\textbf{Multi-stage SFT ablation} on AIME~2024, AIME~2025, and GPQA-Diamond (Pass@1, $n{=}8$, Qwen2.5-7B). OT = OpenThoughts only; OM/Full = \openmementos{} Full Attention; OM/Mem = \openmementos{} Memento Attention; 2-Stg = OT~$\to$~OM/Full; 3-Stg = OT~$\to$~OM/Full~$\to$~OM/Mem (Ours). Training directly on \openmementos{} from the base model (OM variants) substantially underperforms vanilla reasoning SFT~(OT). Our three-stage pipeline enables block masking while retaining strong performance.}
    \label{fig:2stage}
\end{figure}
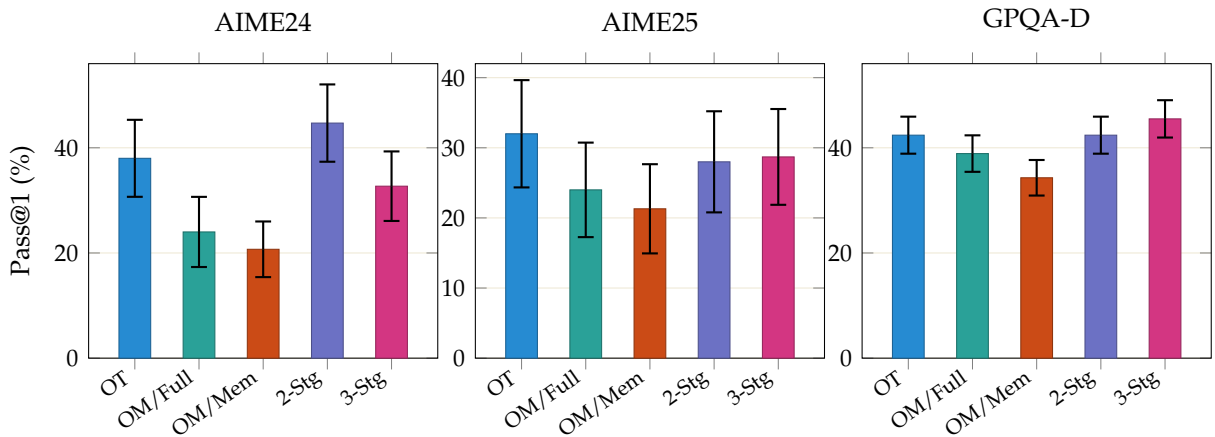

Training directly on \openmementos{} from the base model (OM/Full and OM/Mem) substantially underperforms the vanilla reasoning model~(OT) across all three benchmarks, indicating that the block-memento format is difficult to learn without first acquiring strong reasoning abilities.
Fine-tuning the reasoning model in a single step (OT~$\to$~OM/Full) recovers and sometimes exceeds OT---most notably on AIME~2024, where it improves from 38.0\% to 44.7\%---but this model cannot benefit from block masking at inference.
Our full pipeline (OT~$\to$~OM/Full~$\to$~OM/Mem) enables block masking while retaining strong performance: 32.7\% on AIME~2024, 28.7\% on AIME~2025, and 45.5\% on GPQA-Diamond---the highest GPQA-Diamond score among all configurations.

For models that already have reasoning skills (Qwen3-8B/32B), the OpenThoughts stage is unnecessary and we use two-stage SFT directly.

\begin{tcolorbox}[finding]
\textbf{Staged training helps.} For Qwen2.5-7B, acquiring reasoning ability first (OT), then learning the block-memento format (Full Attention), and finally adapting to block masking (Memento Attention) yields the best results. The full pipeline achieves the highest GPQA-Diamond score among all configurations.
\end{tcolorbox}

\subsubsection{Toy Transformer: KV Channel is Architectural}
\label{sec:toy-kv-probing}

\paragraph{Experiment setup.}
To verify that the implicit KV channel is an architectural property rather than an artifact of large-scale training, we replicate the probing experiment on a controlled toy transformer (4 layers, $d{=}128$, 810K parameters) trained on synthetic sequences of 10 blocks with \texttt{keep\_last\_n\_blocks=0}. Each block contains 5 random digits followed by a 3-digit cumulative sum. We train probes on 50K samples to predict the injected digits from memento KV states.

\begin{figure}[H]
    \centering
    \includegraphics[width=\textwidth]{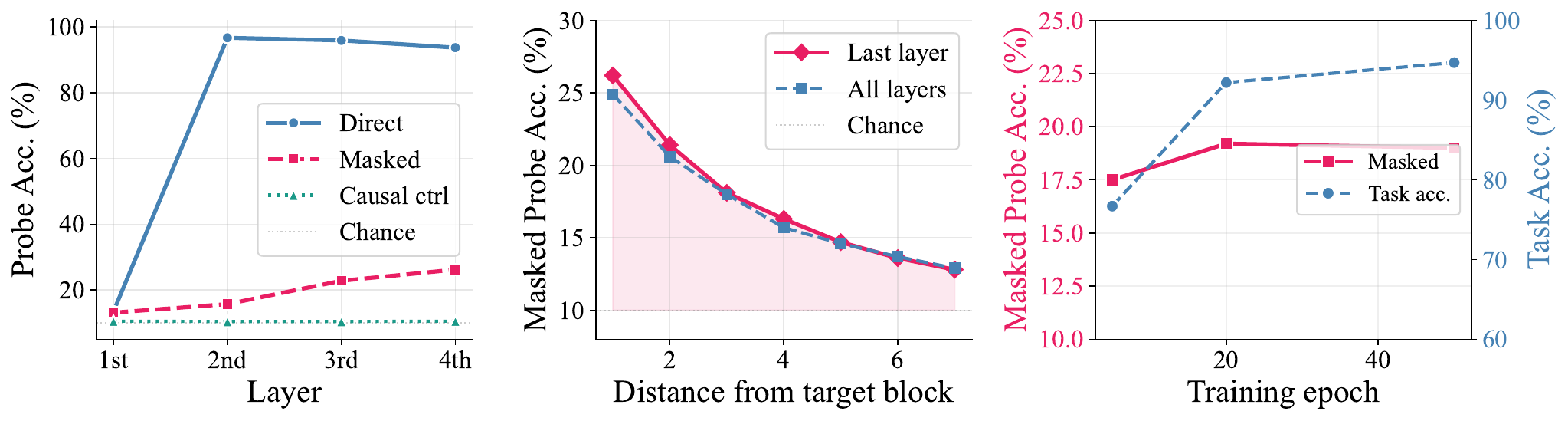}
    \caption{\textbf{Toy transformer KV probing (\texttt{keep0}).} \textbf{Left:} Probe accuracy by layer depth---deeper layers carry more signal, especially for the masked condition. \textbf{Center:} Masked probe accuracy decays with distance from the target block but remains above chance even 7 hops away. \textbf{Right:} Leakage stays constant (17--19\%) across training checkpoints even as task accuracy climbs from 77\% to 95\%, confirming the channel is architectural.}
    \label{fig:toy-kv-probing}
\end{figure}

\paragraph{Results.}
The toy model exhibits the same leakage patterns as the production models (\Cref{fig:toy-kv-probing}). 
The left panel shows how leakage varies with layer depth. 
The first layer carries almost no signal for any condition, but deeper layers progressively encode more---the last layer reaches 26.2\% masked accuracy compared to 13.1\% at the first layer. 
The right panel shows how masked accuracy decays with distance from the target block: a sharp drop from \texttt{block\_index}=+1 (26.2\%) to \texttt{block\_index}=+2 (21.4\%), followed by gradual attenuation, with signal still $1.3\times$ chance at \texttt{block\_index}=+7. 
The right panel confirms that the channel is architectural rather than learned: masked leakage stays roughly constant (17--19\%) across training checkpoints even as task accuracy climbs from 77\% to 95\%.

\subsection{Example Memento Traces}\label{app:example}

We present representative \memento{} traces from AIME'26 (\cref{example:aime26}), LiveCodeBench (\cref{example:lcb}), and GPQA Diamond (\cref{example:gpqa}), showing how the model segments reasoning into blocks and generates compressed mementos.

\subsubsection{AIME'26}\label{example:aime26}

\begin{usermsg}
Find the sum of the $10$th terms of all arithmetic sequences of integers that have first term equal to $4$ and include both $24$ and $34$ as terms.
\end{usermsg}

\begin{botmsg}
\tagtext{<think>}

\begingroup\footnotesize\color{solBase01}
Okay, so I need to find the sum of the 10th terms of all arithmetic sequences of integers that have the first term equal to 4 and include both 24 and 34 as terms. Hmm, let's break this down step by step. First, an arithmetic sequence has a common difference between consecutive terms. Since the first term is 4, the sequence can be written as:
$a_{1}$ = 4,\;
$a_n = a_{n-1} + d$, where d is the common difference (an integer because the terms are integers).
So, the nth term formula is $a_n$ = 4 + (n-1)d. The problem states that both 24 and 34 must be terms in the sequence. [\ldots]
\endgroup

\begin{summaryblock}
Defines arithmetic sequence with $a_1=4$, integer $d>0$; terms $a_n=4+(n-1)d$. Requires 24 and 34 as terms: $(k-1)d=20$, $(m-1)d=30$ with integers $k,m\geq 1$. Concludes $d$ must be a positive common divisor of 20 and 30; candidates $d \in \{1,2,5,10\}$.
\end{summaryblock}

\vspace{4pt}\noindent\textit{[\ldots two more reasoning blocks and mementos omitted for brevity \ldots]}\vspace{4pt}

\begin{summaryblock}
Re-derives via $p,q$ term indices: $(p-1)d=20$, $(q-1)d=30$, $(q-p)d=10$; concludes $d$ divides 20 and 30, so $d \in \{1,2,5,10\}$. Recomputes 10th terms and confirms sum $=178$. Final answer: $\boxed{178}$.
\end{summaryblock}

\tagtext{</think>}

\begin{finalanswer}
Given an arithmetic sequence with the first term \( a_1 = 4 \) and including both 24 and 34 as terms, we need to find the sum of the 10th terms of all such sequences.

1. \textbf{General Form of Arithmetic Sequence}:
   The \( n \)-th term of an arithmetic sequence is given by \( a_n = 4 + (n-1)d \), where \( d \) is the common difference.

2. \textbf{Conditions}:
   - \( 24 = 4 + (k-1)d \) implies \( (k-1)d = 20 \)
   - \( 34 = 4 + (m-1)d \) implies \( (m-1)d = 30 \)

3. \textbf{Common Divisors}:
   - \( d \) must be a common divisor of 20 and 30. The greatest common divisor (GCD) of 20 and 30 is 10, so the common divisors are 1, 2, 5, and 10.

4. \textbf{Possible Values of \( d \)}:
   - \( d = 1 \)
   - \( d = 2 \)
   - \( d = 5 \)
   - \( d = 10 \)

5. \textbf{Calculating the 10th Term for Each \( d \)}:
   - For \( d = 1 \): \( a_{10} = 4 + 9 \times 1 = 13 \)
   - For \( d = 2 \): \( a_{10} = 4 + 9 \times 2 = 22 \)
   - For \( d = 5 \): \( a_{10} = 4 + 9 \times 5 = 49 \)
   - For \( d = 10 \): \( a_{10} = 4 + 9 \times 10 = 94 \)

6. \textbf{Summing the 10th Terms}:
   - Sum = \( 13 + 22 + 49 + 94 = 178 \)

Thus, the sum of the 10th terms of all such arithmetic sequences is \(\boxed{178}\).
\end{finalanswer}
\end{botmsg}

\subsubsection{LiveCodeBench v6}\label{example:lcb}

\begin{usermsg}
There are $N$ people labeled $1$ to $N$, who have played several one-on-one games without draws. Initially, each person started with $0$ points. In each game, the winner's score increased by $1$ and the loser's score decreased by $1$ (scores can become negative). Determine the final score of person $N$ if the final score of person $i$ \,($1 \leq i \leq N{-}1$) is $A_i$. It can be shown that the final score of person $N$ is uniquely determined regardless of the sequence of games.

\medskip
\textbf{Input}\quad The input is given from Standard Input in the following format:
\begin{codeio}
N\\
$A_1\;A_2\;\ldots\;A_{N-1}$
\end{codeio}

\textbf{Output}\quad Print the answer.

\medskip
\textbf{Constraints}
\begin{itemize}[nosep,leftmargin=1.5em]
\item $2 \leq N \leq 100$
\item $-100 \leq A_i \leq 100$
\item All input values are integers.
\end{itemize}

\medskip\textbf{Sample Input 1}
\begin{codeio}4\\ 1 -2 -1\end{codeio}
\textbf{Sample Output 1}
\begin{codeio}2\end{codeio}
Here is one possible sequence of games where the final scores of persons 1, 2, 3 are $1, -2, -1$, respectively.
\begin{itemize}[nosep,leftmargin=1.5em]
\item Initially, persons 1, 2, 3, 4 have $0, 0, 0, 0$ points, respectively.
\item Persons 1 and 2 play, and person 1 wins. Scores: $1, -1, 0, 0$.
\item Persons 1 and 4 play, and person 4 wins. Scores: $0, -1, 0, 1$.
\item Persons 1 and 2 play, and person 1 wins. Scores: $1, -2, 0, 1$.
\item Persons 2 and 3 play, and person 2 wins. Scores: $1, -1, -1, 1$.
\item Persons 2 and 4 play, and person 4 wins. Scores: $1, -2, -1, 2$.
\end{itemize}
In this case, the final score of person 4 is 2. Other possible sequences exist, but the score of person 4 will always be 2.

\medskip\textbf{Sample Input 2}
\begin{codeio}3\\ 0~0\end{codeio}
\textbf{Sample Output 2}
\begin{codeio}0\end{codeio}

\medskip\textbf{Sample Input 3}
\begin{codeio}6\\ 10~20~30~40~50\end{codeio}
\textbf{Sample Output 3}
\begin{codeio}$-$150\end{codeio}
\end{usermsg}

\begin{botmsg}
\tagtext{<think>}

\begingroup\footnotesize\color{solBase01}
Okay, let's see. I have to find the final score of person N based on the given scores of the first N-1 people. Hmm. The problem says that the sum is uniquely determined regardless of the games, so there's a formula here. Let me think about the constraints. Each game involves two people, and each time, the winner gains 1, loser loses 1. So for each game, the total sum of all scores changes by (1 + -1) = 0. So the total sum of all people's scores must remain the same as the initial total. Wait, initially, everyone has 0, so the total sum is 0. Therefore, after all games, the sum of all N people's scores must still be zero. Oh right! Because each game's change adds up to zero. So the total sum is always zero. That's the key here. So, the sum of all the given $A_i$ (for i from 1 to N-1) plus the score of person N must equal zero. Therefore, the score of person N is equal to negative the sum of the given $A_i$'s.
\endgroup

\begin{summaryblock}
Let $s = \sum_{i=1}^{N-1} A_i$; invariant: each 1-on-1 game changes total score by 0, so total score remains initial 0; hence $s + x = 0$ where $x$ is person $N$'s score $\rightarrow$ $x = -s$.
\end{summaryblock}

\begingroup\footnotesize\color{solBase01}
Let me check the sample input 1:
Sample 1 has N=4, and A values 1, -2, -1. Their sum is 1 + (-2) + (-1) = -2. So the Nth person's score is -(-2) = 2, which matches the sample output.
Sample 2: N=3, A is [0,0], sum is 0. So the answer is 0, which matches.
Sample 3: N=6, the A's are 10,20,30,40,50. Sum is 10+20 is 30, plus 30 is 60, plus 40 is 100, plus 50 is 150. So the answer would be -150. Which matches the sample output. Okay, so the formula is correct. Therefore, the solution is to compute the sum of all the $A_i$'s, then take the negative of that sum.
\endgroup

\begin{summaryblock}
Verifies formula on samples: sample\,1 sum $=1{-}2{-}1={-}2$ $\Rightarrow$ output 2; sample\,2 sum $=0$ $\Rightarrow$ output 0; sample\,3 sum $=10{+}20{+}30{+}40{+}50=150$ $\Rightarrow$ output $-150$; concludes solution is compute $\sum A_i$ and output its negation.
\end{summaryblock}

\begingroup\footnotesize\color{solBase01}
The steps for the code would be:
1. Read N from input. 2. Read the next line which has N-1 integers. 3. Sum those integers. 4. The answer is negative of that sum. So the code in Python would be straightforward. Let's see. Read N, then the list of $A_i$, compute \texttt{sum(A\_list)}, then print(-sum). Wait, let me check possible edge cases. Like N=2. Let's say N=2, and the input is, say, [-5].
Then the sum is -5, so the answer is 5. Another test case: if all $A_i$ are 0, then the answer is 0. That's covered in sample 2. Another test case where the sum is positive. Like N=3 and A is [3, -1].
Sum is 2, so the answer is -2. So the code should be simple. Let me code that. Wait, code structure:
Read first line as N. Second line is list of N-1 numbers. So in Python: Read input: Probably: import sys n = int(sys.stdin.readline()) \texttt{a\_list} = list(map(int, sys.stdin.readline().split())) \texttt{sum\_a} = sum(\texttt{a\_list}) print(-\texttt{sum\_a})
Wait, that's all. That's the code. Let me check the sample input 1: Sample1 input: 4 1 -2 -1 $\rightarrow$ sum is 1-2-1 =-2 $\rightarrow$ output 2. Correct.
Sample2 input:3 0 0 $\rightarrow$ sum is 0 $\rightarrow$ output 0. Sample3:6 10 20 30 40 50 $\rightarrow$ sum is 150 $\rightarrow$ output -150. Correct. Looks like this works. I think that's the solution. So the code is really simple.
\endgroup

\begin{summaryblock}
Algorithm plan: read $N$; read $N{-}1$ ints; print negative of their sum. Writes Python sketch using \texttt{sum()} and \texttt{print(-sum)}; checks edge case $N=2$ (e.g., input $[-5]$ $\rightarrow$ $5$) and other example $[3,-1]$ $\rightarrow$ $-2$; reiterates code simplicity.
\end{summaryblock}

\tagtext{</think>}

\begin{finalanswer}
\begin{verbatim}
n = int(input())
a_list = list(map(int, input().split()))
total = sum(a_list)
print(-total)
\end{verbatim}
\end{finalanswer}
\end{botmsg}

\subsubsection{GPQA Diamond}\label{example:gpqa}
\begin{usermsg}
A eukaryotic cell evolved a mechanism to turn macromolecular building blocks into energy. The process occurs in mitochondria, which are cellular energy factories. In the series of redox reactions, the energy from food is stored between the phosphate groups and used as a universal cellular currency. The energy-laden molecules are shuttled out of the mitochondrion to serve in all cellular processes. You discovered a new anti-diabetes drug and want to investigate whether it has an effect on the mitochondria. You set up a range of experiments with your HEK293 cell line. Which of the experiments listed below will not help you discover the mitochondrial role of your drug:

\begin{itemize}[leftmargin=*]
    \item A) Confocal fluorescence microscopy with Mito-RTP staining
    \item B) Differential centrifugation extraction of mitochondria + Glucose Uptake Colorimetric Assay
    \item C) Flow cytometry with JC-1 dye
    \item D) Luciferase/luciferin ATP bioluminescence assay
\end{itemize}
\end{usermsg}

\begin{botmsg}
\tagtext{<think>}

\begingroup\footnotesize\color{solBase01}
Okay, so I need to figure out which of these experiments won't help in discovering the mitochondrial role of the new anti-diabetes drug. The question is about which experiment isn't related to mitochondria's function, right? Let me start by recalling what each part involves. First, let me go through each option one by one. Option A: Confocal fluorescence microscopy after Mito-RTP staining. Mito-RTP must be a mitochondrial stain. Confocal microscopy would let them visualize the mitochondria's structure or distribution in the cells. If the drug affects mitochondria, maybe it would change their morphology, so this would be a relevant experiment. So A is probably not the answer. Option B: Differential centrifugation extraction of mitochondria followed by a Glucose Uptake Colorimetric Assay. Hmm, differential centrifugation is a method to isolate mitochondria. But the glucose uptake assay... Wait, glucose uptake is typically a process that happens before it's used in the mitochondria. Glucose is taken up by the cell via transporters like GLUT4, and then it's processed in glycolysis in the cytoplasm. The mitochondria are involved in later stages like the Krebs cycle, but glucose uptake itself is not a mitochondrial process. So if they're measuring glucose uptake after isolating mitochondria, that might not make sense. Maybe the experiment is flawed here. Wait, or maybe the glucose uptake assay is testing mitochondrial function indirectly? Or maybe the mitochondria aren't the right organelle for that. Since glucose isn't taken up by mitochondria but by the whole cell, using isolated mitochondria in this assay might not be effective. Therefore, this experiment might not help in assessing mitochondrial role. So B could be the answer.
\endgroup

\begin{summaryblock}
Evaluates options A--D; identifies goal of finding experiment unrelated to mitochondrial function; notes A uses mitochondrial stain $\rightarrow$ relevant; analyzes B as glucose uptake assay on isolated mitochondria $\rightarrow$ likely irrelevant since mitochondria do not take up glucose; tentatively marks B as candidate.
\end{summaryblock}

\begingroup\footnotesize\color{solBase01}
Option C: Flow cytometry with Tetrachloro... some dye. I think that's a mitochondrial membrane potential dye. JC-1? Yes, that's a common dye for mitochondrial membrane potential. So labeling with that would measure if the drug affects mitochondrial function (since changes in membrane potential indicate issues). So C is a valid experiment related to mitochondria. Option D: Transformation with luciferase and adding luciferin. Wait, firefly luciferase requires luciferin and uses ATP to produce light. So if they put the luciferase in the cells, then adding luciferin, the light output would depend on ATP levels. Because the ATP in the supernatant (or in the cells?) would be used by the luciferase. Wait, the question says "transformation of cells with recombinant luciferase" – maybe the luciferase is expressed in the cells, so when you add luciferin, the cells' ATP would power the reaction, and the light measured would reflect ATP levels. Since mitochondrial function affects ATP production, this would be a way to test that. Therefore, D is measuring ATP levels via luciferase, so it's related to mitochondria. Therefore, the one that doesn't help is B. Because the glucose uptake assay on isolated mitochondria isn't the right approach, since glucose uptake is done by the cell membrane, not mitochondria. The mitochondria process pyruvate, but not glucose directly. So the glucose uptake assay would be measuring something that's not directly a mitochondrial function. Hence, B is the one that won't help.
\endgroup

\begin{summaryblock}
Identifies C as JC-1 mitochondrial potential dye $\rightarrow$ valid; D as luciferase--ATP bioluminescence assay $\rightarrow$ valid; concludes only B (glucose uptake on mitochondria) irrelevant to mitochondrial function; final answer: B.
\end{summaryblock}

\tagtext{</think>}

\begin{finalanswer}
The experiment that will not help discover the mitochondrial role of the drug is \textbf{B) Differential centrifugation extraction of mitochondria followed by the Glucose Uptake Colorimetric Assay Kit}.

\#\#\# Reasoning: 
\begin{itemize}[leftmargin=*]
    \item \textbf{Option A}: Uses \textbf{Mito-RTP staining} with confocal microscopy, directly assessing mitochondrial structure or localization. This is relevant for determining if the drug affects mitochondrial integrity or distribution.  
    \item \textbf{Option B}: The \textbf{Glucose Uptake Assay} measures glucose transport into cells or specific fractions. Mitochondria do not directly take up glucose; glucose uptake is mediated by plasma membrane glucose transporters (e.g., GLUT4). Performing this assay on isolated mitochondria is irrelevant because mitochondria are not involved in glucose uptake. This experiment would not provide meaningful data about mitochondrial function or drug effects on mitochondria.  
    \item \textbf{Option C}: Uses \textbf{flow cytometry with JC-1 dye}, a mitochondrial membrane potential indicator. This directly evaluates mitochondrial health and redox activity, making it highly relevant.  
    \item \textbf{Option D}: Measures \textbf{ATP levels} via luciferase/luciferin bioluminescence. Since mitochondria are the primary ATP producers, this assay indirectly assesses mitochondrial energy production and drug effects on it.  
\end{itemize}

\#\#\# Conclusion: \\
\textbf{B} is the only experiment unrelated to mitochondrial function, as glucose uptake is not a mitochondrial process. The others directly or indirectly probe mitochondrial activity or structure. 

\textbf{Answer: B}
\end{finalanswer}
\end{botmsg}


\begin{thebibliography}{41}
\providecommand{\natexlab}[1]{#1}
\providecommand{\url}[1]{\texttt{#1}}
\expandafter\ifx\csname urlstyle\endcsname\relax
  \providecommand{\doi}[1]{doi: #1}\else
  \providecommand{\doi}{doi: \begingroup \urlstyle{rm}\Url}\fi

\bibitem[Aghajohari et~al.(2025)Aghajohari, Chitsaz, Kazemnejad, Chandar,
  Sordoni, Courville, and Reddy]{markovianThinker2025}
Milad Aghajohari, Kamran Chitsaz, Amirhossein Kazemnejad, Sarath Chandar,
  Alessandro Sordoni, Aaron Courville, and Siva Reddy.
\newblock {The Markovian Thinker}: Architecture-agnostic linear scaling of
  reasoning.
\newblock \emph{arXiv preprint arXiv:2510.06557}, 2025.

\bibitem[Balunovi{\'c} et~al.(2025)Balunovi{\'c}, Dekoninck, Petrov,
  Jovanovi{\'c}, and Vechev]{balunovic2025matharena}
Mislav Balunovi{\'c}, Jasper Dekoninck, Ivo Petrov, Nikola Jovanovi{\'c}, and
  Martin Vechev.
\newblock {MathArena}: Evaluating {LLMs} on uncontaminated math competitions.
\newblock \emph{Advances in Neural Information Processing Systems, Datasets and
  Benchmarks Track}, 2025.

\bibitem[Beltagy et~al.(2020)Beltagy, Peters, and Cohan]{beltagy2020longformer}
Iz~Beltagy, Matthew~E Peters, and Arman Cohan.
\newblock Longformer: The long-document transformer.
\newblock \emph{arXiv preprint arXiv:2004.05150}, 2020.

\bibitem[Cai et~al.(2025)Cai, Xiao, Sun, Luo, Zhang, Wan, Li, Zhou, Chang, Gu,
  Dong, Anandkumar, Asi, and Hu]{cai2025r}
Zefan Cai, Wen Xiao, Hanshi Sun, Cheng Luo, Yikai Zhang, Ke~Wan, Yucheng Li,
  Yeyang Zhou, Li-Wen Chang, Jiuxiang Gu, Zhen Dong, Anima Anandkumar,
  Abedelkadir Asi, and Junjie Hu.
\newblock {R-KV}: Redundancy-aware {KV} cache compression for reasoning models.
\newblock \emph{arXiv preprint arXiv:2505.24133}, 2025.

\bibitem[{DeepSeek-AI}(2024)]{deepseekv3}
{DeepSeek-AI}.
\newblock {DeepSeek-V3} technical report.
\newblock \emph{arXiv preprint arXiv:2412.19437}, 2024.

\bibitem[{DeepSeek-AI} et~al.(2025){DeepSeek-AI}, Guo, Yang, Zhang, Song,
  Zhang, Xu, Zhu, Ma, Wang, et~al.]{deepseekr1}
{DeepSeek-AI}, Daya Guo, Dejian Yang, Haowei Zhang, Junxiao Song, Ruoyu Zhang,
  Runxin Xu, Qihao Zhu, Shirong Ma, Peiyi Wang, et~al.
\newblock {DeepSeek-R1}: Incentivizing reasoning capability in {LLMs} via
  reinforcement learning.
\newblock \emph{arXiv preprint arXiv:2501.12948}, 2025.

\bibitem[Guha et~al.(2025)Guha, Marten, Keh, Raoof, Smyrnis, Bansal, Nezhurina,
  Mercat, Vu, Sprague, Suvarna, Feuer, Chen, Khan, Frankel, Grover, Choi,
  Muennighoff, Su, Zhao, Yang, Pimpalgaonkar, Sharma, Ji, Deng, Pratt,
  Ramanujan, Saad-Falcon, Li, Dave, Albalak, Arora, Wulfe, Hegde, Durrett, Oh,
  Bansal, Gabriel, Grover, Chang, Shankar, Gokaslan, Merrill, Hashimoto, Choi,
  Jitsev, Heckel, Sathiamoorthy, Dimakis, and Schmidt]{openthoughts2025}
Etash Guha, Ryan Marten, Sedrick Keh, Negin Raoof, Georgios Smyrnis, Hritik
  Bansal, Marianna Nezhurina, Jean Mercat, Trung Vu, Zayne Sprague, Ashima
  Suvarna, Benjamin Feuer, Liangyu Chen, Zaid Khan, Eric Frankel, Sachin
  Grover, Caroline Choi, Niklas Muennighoff, Shiye Su, Wanjia Zhao, John Yang,
  Shreyas Pimpalgaonkar, Kartik Sharma, Charlie Cheng-Jie Ji, Yichuan Deng,
  Sarah Pratt, Vivek Ramanujan, Jon Saad-Falcon, Jeffrey Li, Achal Dave, Alon
  Albalak, Kushal Arora, Blake Wulfe, Chinmay Hegde, Greg Durrett, Sewoong Oh,
  Mohit Bansal, Saadia Gabriel, Aditya Grover, Kai-Wei Chang, Vaishaal Shankar,
  Aaron Gokaslan, Mike~A. Merrill, Tatsunori Hashimoto, Yejin Choi, Jenia
  Jitsev, Reinhard Heckel, Maheswaran Sathiamoorthy, Alexandros~G. Dimakis, and
  Ludwig Schmidt.
\newblock Open{T}houghts: Data recipes for reasoning models, 2025.
\newblock URL \url{https://arxiv.org/abs/2506.04178}.

\bibitem[Hao et~al.(2024)Hao, Sukhbaatar, Su, Li, Hu, Weston, and
  Tian]{coconut2024}
Shibo Hao, Sainbayar Sukhbaatar, DiJia Su, Xian Li, Zhiting Hu, Jason Weston,
  and Yuandong Tian.
\newblock Training large language models to reason in a continuous latent
  space.
\newblock \emph{arXiv preprint arXiv:2412.06769}, 2024.

\bibitem[Hou et~al.(2025)Hou, Zhang, Ji, Liu, Qian, Andreas, and
  Chang]{thinkPrune2025}
Bairu Hou, Yang Zhang, Jiabao Ji, Yujian Liu, Kaizhi Qian, Jacob Andreas, and
  Shiyu Chang.
\newblock {ThinkPrune}: Pruning long chain-of-thought of {LLMs} via
  reinforcement learning.
\newblock \emph{arXiv preprint arXiv:2504.01296}, 2025.

\bibitem[Kang et~al.(2025)Kang, Sun, Chen, and Zou]{c3ot2024}
Yu~Kang, Xianghui Sun, Liangyu Chen, and Wei Zou.
\newblock {C3oT}: Generating shorter chain-of-thought without compromising
  effectiveness.
\newblock In \emph{Proceedings of the AAAI Conference on Artificial
  Intelligence}, volume~39, pages 24312--24320, 2025.

\bibitem[Kwon et~al.(2023)Kwon, Li, Zhuang, Sheng, Zheng, Yu, Gonzalez, Zhang,
  and Stoica]{kwon2023vllm}
Woosuk Kwon, Zhuohan Li, Siyuan Zhuang, Ying Sheng, Lianmin Zheng, Cody~Hao Yu,
  Joseph~E. Gonzalez, Hao Zhang, and Ion Stoica.
\newblock Efficient memory management for large language model serving with
  {PagedAttention}.
\newblock In \emph{Proceedings of the 29th Symposium on Operating Systems
  Principles (SOSP)}, 2023.

\bibitem[Li et~al.(2025)Li, Zhong, Zheng, Wen, Xu, Cheng, Zhang, and
  Xu]{stepEntropy2025}
Zeju Li, Jianyuan Zhong, Ziyang Zheng, Xiangyu Wen, Zhijian Xu, Yingying Cheng,
  Fan Zhang, and Qiang Xu.
\newblock Making slow thinking faster: Compressing {LLM} chain-of-thought via
  step entropy.
\newblock \emph{arXiv preprint arXiv:2508.03346}, 2025.

\bibitem[{MiniMax} et~al.(2025){MiniMax}, Li, Gong, Yang, Shan, Liu, Zhu,
  Zhang, Guo, Chen, et~al.]{minimax_m1}
{MiniMax}, Aonian Li, Bangwei Gong, Bo~Yang, Boji Shan, Chang Liu, Cheng Zhu,
  Chunhao Zhang, Congchao Guo, Da~Chen, et~al.
\newblock {MiniMax-M1}: Scaling test-time compute efficiently with lightning
  attention.
\newblock \emph{arXiv preprint arXiv:2506.13585}, 2025.

\bibitem[Monea et~al.(2025)Monea, Feldman, Padmanabhan, Brantley, and
  Artzi]{monea2025breadcrumbs}
Giovanni Monea, Yair Feldman, Shankar Padmanabhan, Kiant{\'e} Brantley, and
  Yoav Artzi.
\newblock Breadcrumbs reasoning: Memory-efficient reasoning with compression
  beacons.
\newblock \emph{arXiv preprint arXiv:2510.13797}, 2025.

\bibitem[{OpenAI}(2024)]{openai_o1}
{OpenAI}.
\newblock Learning to reason with {LLMs}.
\newblock \url{https://openai.com/index/learning-to-reason-with-llms/}, 2024.
\newblock Blog post, September 2024.

\bibitem[{Qwen Team}(2024)]{qwq2024}
{Qwen Team}.
\newblock {QwQ}: Reflect deeply on the boundaries of the unknown.
\newblock \url{https://qwenlm.github.io/blog/qwq-32b-preview/}, 2024.
\newblock Blog post, November 2024.

\bibitem[Ramachandran et~al.(2025)Ramachandran, Neseem, Sakr, Venkatesan,
  Khailany, and Krishna]{ramachandran2025thinkv}
Akshat Ramachandran, Marina Neseem, Charbel Sakr, Rangharajan Venkatesan,
  Brucek Khailany, and Tushar Krishna.
\newblock Thinkv: Thought-adaptive kv cache compression for efficient reasoning
  models.
\newblock \emph{arXiv preprint arXiv:2510.01290}, 2025.

\bibitem[Schulman et~al.(2017)Schulman, Wolski, Dhariwal, Radford, and
  Klimov]{ppo}
John Schulman, Filip Wolski, Prafulla Dhariwal, Alec Radford, and Oleg Klimov.
\newblock Proximal policy optimization algorithms.
\newblock \emph{arXiv preprint arXiv:1707.06347}, 2017.

\bibitem[Shao et~al.(2024)Shao, Wang, Zhu, Xu, Song, Bi, Zhang, Zhang, Li, Wu,
  and Guo]{grpo}
Zhihong Shao, Peiyi Wang, Qihao Zhu, Runxin Xu, Junxiao Song, Xiao Bi, Haowei
  Zhang, Mingchuan Zhang, Y.K. Li, Y.~Wu, and Daya Guo.
\newblock {DeepSeekMath}: Pushing the limits of mathematical reasoning in open
  language models.
\newblock \emph{arXiv preprint arXiv:2402.03300}, 2024.

\bibitem[Shen et~al.(2025)Shen, Yan, Zhang, Hu, Du, and He]{codi2025}
Zhenyi Shen, Hanqi Yan, Linhai Zhang, Zhanghao Hu, Yali Du, and Yulan He.
\newblock Codi: Compressing chain-of-thought into continuous space via
  self-distillation.
\newblock In \emph{Proceedings of the 2025 Conference on Empirical Methods in
  Natural Language Processing}, pages 677--693, 2025.

\bibitem[Shrivastava et~al.(2025)Shrivastava, Awadallah, Balachandran, Garg,
  Behl, and Papailiopoulos]{gfpo2025}
Vaishnavi Shrivastava, Ahmed Awadallah, Vidhisha Balachandran, Shivam Garg,
  Harkirat Behl, and Dimitris Papailiopoulos.
\newblock Sample more to think less: Group filtered policy optimization for
  concise reasoning.
\newblock \emph{arXiv preprint arXiv:2508.09726}, 2025.

\bibitem[Song et~al.(2025)Song, Jo, Kim, and Kim]{song2025reasoning}
Jiwon Song, Dongwon Jo, Yulhwa Kim, and Jae-Joon Kim.
\newblock Reasoning path compression: Compressing generation trajectories for
  efficient {LLM} reasoning.
\newblock \emph{arXiv preprint arXiv:2505.13866}, 2025.

\bibitem[Tan et~al.(2025)Tan, Li, Ju, Luo, Song, and Luan]{colar2025}
Wenhui Tan, Jiaze Li, Jianzhong Ju, Zhenbo Luo, Ruihua Song, and Jian Luan.
\newblock Think silently, think fast: Dynamic latent compression of {LLM}
  reasoning chains.
\newblock \emph{arXiv preprint arXiv:2505.16552}, 2025.
\newblock URL \url{https://arxiv.org/abs/2505.16552}.

\bibitem[{Team Olmo} et~al.(2025){Team Olmo}, Ettinger, Bertsch, Kuehl, Graham,
  Heineman, Groeneveld, Brahman, Timbers, Ivison, et~al.]{olmo3}
{Team Olmo}, Allyson Ettinger, Amanda Bertsch, Bailey Kuehl, David Graham,
  David Heineman, Dirk Groeneveld, Faeze Brahman, Finbarr Timbers, Hamish
  Ivison, et~al.
\newblock Olmo 3.
\newblock \emph{arXiv preprint arXiv:2512.13961}, 2025.

\bibitem[von Werra et~al.(2020)von Werra, Belkada, Tunstall, Beeching, Thrush,
  Lambert, and Huang]{vonwerra2022trl}
Leandro von Werra, Younes Belkada, Lewis Tunstall, Edward Beeching, Tristan
  Thrush, Nathan Lambert, and Shengyi Huang.
\newblock Trl: Transformer reinforcement learning.
\newblock \url{https://github.com/huggingface/trl}, 2020.

\bibitem[Wang(2026)]{wang2026memento2}
Jun Wang.
\newblock Memento 2: Learning by stateful reflective memory, 2026.
\newblock URL \url{https://arxiv.org/abs/2512.22716}.

\bibitem[Wang et~al.(2025)Wang, Luo, Yao, Huang, He, Liu, Tan, Huang, Cao, Tao,
  and Shen]{r1compress2025}
Yibo Wang, Haotian Luo, Huanjin Yao, Tiansheng Huang, Haiying He, Rui Liu,
  Naiqiang Tan, Jiaxing Huang, Xiaochun Cao, Dacheng Tao, and Li~Shen.
\newblock R1-compress: Long chain-of-thought compression via chunk compression
  and search, 2025.
\newblock URL \url{https://arxiv.org/abs/2505.16838}.

\bibitem[Wu et~al.(2025)Wu, Li, Zhao, Zhang, Ou, Yin, Zhang, Yu, Zhang, Jiang,
  Xie, Huang, Cheng, Wang, Cheng, and Zhou]{resum2025}
Xixi Wu, Kuan Li, Yida Zhao, Liwen Zhang, Litu Ou, Huifeng Yin, Zhongwang
  Zhang, Xinmiao Yu, Dingchu Zhang, Yong Jiang, Pengjun Xie, Fei Huang, Minhao
  Cheng, Shuai Wang, Hong Cheng, and Jingren Zhou.
\newblock {ReSum}: Unlocking long-horizon search intelligence via context
  summarization.
\newblock \emph{arXiv preprint arXiv:2509.13313}, 2025.

\bibitem[Xia et~al.(2025)Xia, Leong, Wang, Li, and Li]{tokenSkip2025}
Heming Xia, Chak~Tou Leong, Wenjie Wang, Yongqi Li, and Wenjie Li.
\newblock {TokenSkip}: Controllable chain-of-thought compression in {LLMs}.
\newblock In \emph{Proceedings of the 2025 Conference on Empirical Methods in
  Natural Language Processing}, pages 3351--3363, 2025.

\bibitem[Xu et~al.(2025)Xu, Liang, Mei, Gao, Tan, and Zhang]{amem2025}
Wujiang Xu, Zujie Liang, Kai Mei, Hang Gao, Juntao Tan, and Yongfeng Zhang.
\newblock {A-MEM}: Agentic memory for {LLM} agents.
\newblock \emph{arXiv preprint arXiv:2502.12110}, 2025.

\bibitem[Yan et~al.(2025)Yan, Shen, Liu, Jiang, Zhang, Shao, and
  Zhuang]{inftyThink2025}
Yuchen Yan, Yongliang Shen, Yang Liu, Jin Jiang, Mengdi Zhang, Jian Shao, and
  Yueting Zhuang.
\newblock {InftyThink}: Breaking the length limits of long-context reasoning in
  large language models.
\newblock \emph{arXiv preprint arXiv:2503.06692}, 2025.

\bibitem[Yan et~al.(2026)Yan, Jiang, Jiang, Li, Wen, Zhang, Zhou, Shao, Zhuang,
  and Shen]{inftyThinkPlus2026}
Yuchen Yan, Liang Jiang, Jin Jiang, Shuaicheng Li, Zujie Wen, Zhiqiang Zhang,
  Jun Zhou, Jian Shao, Yueting Zhuang, and Yongliang Shen.
\newblock {InftyThink+}: Effective and efficient infinite-horizon reasoning via
  reinforcement learning.
\newblock \emph{arXiv preprint arXiv:2602.06960}, 2026.

\bibitem[Yang et~al.(2025)Yang, Srebro, McAllester, and Li]{pencil2025}
Chenxiao Yang, Nathan Srebro, David McAllester, and Zhiyuan Li.
\newblock {PENCIL}: Long thoughts with short memory.
\newblock \emph{arXiv preprint arXiv:2503.14337}, 2025.

\bibitem[Yang et~al.(2026)Yang, Guo, Huang, Wang, Shi, Wang, Liang, and
  Tang]{accordionThinking2026}
Zhicheng Yang, Zhijiang Guo, Yinya Huang, Yongxin Wang, Wenlei Shi, Yiwei Wang,
  Xiaodan Liang, and Jing Tang.
\newblock {Accordion-Thinking}: Self-regulated step summaries for efficient and
  readable {LLM} reasoning.
\newblock \emph{arXiv preprint arXiv:2602.03249}, 2026.

\bibitem[Yu et~al.(2025)Yu, Chen, Feng, Chen, Dai, Yu, Zhang, Ma, Liu, Wang,
  and Zhou]{memagent2025}
Hongli Yu, Tinghong Chen, Jiangtao Feng, Jiangjie Chen, Weinan Dai, Qiying Yu,
  Ya-Qin Zhang, Wei-Ying Ma, Jingjing Liu, Mingxuan Wang, and Hao Zhou.
\newblock {MemAgent}: Reshaping long-context {LLM} with multi-conv {RL}-based
  memory agent.
\newblock \emph{arXiv preprint arXiv:2507.02259}, 2025.

\bibitem[Zhang et~al.(2025{\natexlab{a}})Zhang, Zhang, Ma, Zhang, and
  Guo]{lazyEviction2025}
Haoyue Zhang, Hualei Zhang, Xiaosong Ma, Jie Zhang, and Song Guo.
\newblock {LazyEviction}: Lagged {KV} eviction with attention pattern
  observation for efficient long reasoning.
\newblock \emph{arXiv preprint arXiv:2506.15969}, 2025{\natexlab{a}}.

\bibitem[Zhang et~al.(2025{\natexlab{b}})Zhang, Zhu, Sun, Luo, Qiao, Du, Zheng,
  Chen, and Zhang]{lightThinker2025}
Jintian Zhang, Yuqi Zhu, Mengshu Sun, Yujie Luo, Shuofei Qiao, Lun Du,
  Da~Zheng, Huajun Chen, and Ningyu Zhang.
\newblock {LightThinker}: Thinking step-by-step compression.
\newblock In \emph{Proceedings of the 2025 Conference on Empirical Methods in
  Natural Language Processing}, pages 13318--13339, 2025{\natexlab{b}}.

\bibitem[Zhang et~al.(2025{\natexlab{c}})Zhang, Yu, Pan, Jin, Fu, Cai, Lin, and
  Ye]{tokenSqueeze2025}
Yuxiang Zhang, Zhengxu Yu, Weihang Pan, Zhongming Jin, Qiang Fu, Deng Cai,
  Binbin Lin, and Jieping Ye.
\newblock Tokensqueeze: Performance-preserving compression for reasoning llms.
\newblock \emph{arXiv preprint arXiv:2511.13223}, 2025{\natexlab{c}}.

\bibitem[Zheng et~al.(2024)Zheng, Yin, Xie, Sun, Huang, Yu, Cao, Kozyrakis,
  Stoica, Gonzalez, Barrett, and Sheng]{sglang2024}
Lianmin Zheng, Liangsheng Yin, Zhiqiang Xie, Chuyue Sun, Jeff Huang, Cody~Hao
  Yu, Shiyi Cao, Christos Kozyrakis, Ion Stoica, Joseph~E. Gonzalez, Clark
  Barrett, and Ying Sheng.
\newblock {SGLang}: Efficient execution of structured language model programs.
\newblock \emph{arXiv preprint arXiv:2312.07104}, 2024.

\bibitem[Zhou et~al.(2025{\natexlab{a}})Zhou, Chen, Guo, Yan, Lee, Wang, Lee,
  Zhang, Shao, Yang, and Wang]{zhou2025mementoagents}
Huichi Zhou, Yihang Chen, Siyuan Guo, Xue Yan, Kin~Hei Lee, Zihan Wang, Ka~Yiu
  Lee, Guchun Zhang, Kun Shao, Linyi Yang, and Jun Wang.
\newblock Memento: Fine-tuning {LLM} agents without fine-tuning {LLMs},
  2025{\natexlab{a}}.
\newblock URL \url{https://arxiv.org/abs/2508.16153}.

\bibitem[Zhou et~al.(2025{\natexlab{b}})Zhou, Qu, Wu, Kim, Prakash, Rus, Zhao,
  Low, and Liang]{mem1_2025}
Zijian Zhou, Ao~Qu, Zhaoxuan Wu, Sunghwan Kim, Alok Prakash, Daniela Rus,
  Jinhua Zhao, Bryan Kian~Hsiang Low, and Paul~Pu Liang.
\newblock {MEM1}: Learning to synergize memory and reasoning for efficient
  long-horizon agents.
\newblock \emph{arXiv preprint arXiv:2506.15841}, 2025{\natexlab{b}}.

\end{thebibliography}
\end{document}